\documentclass[a4paper,fleqn]{cas-sc}

\usepackage[authoryear]{natbib}

\usepackage{url}
\usepackage{hyperref}
\usepackage{amsmath, bm}
\usepackage{algorithm}
\usepackage{comment}
\usepackage{lineno}
\usepackage{setspace}
\usepackage{subcaption}
\usepackage{siunitx}
\usepackage{graphicx}
\graphicspath{images/}
\usepackage{multirow}
\usepackage{scalerel}

\def\tsc#1{\csdef{#1}{\textsc{\lowercase{#1}}\xspace}}
\tsc{WGM}
\tsc{QE}

\newcommand{\etal}{\emph{et~al.}}
\newcommand{\secref}[1]{Section~\ref{#1}}
\newcommand{\figref}[1]{Figure~\ref{#1}}
\renewcommand{\tabref}[1]{Table~\ref{#1}}

\begin{document}
\let\WriteBookmarks\relax
\def\floatpagepagefraction{1}
\def\textpagefraction{.001}

\shorttitle{Hyperspectral Imaging for Identifying Foreign Objects on Pork Belly}    

\shortauthors{G. Ghimpeteanu \etal}  

\title [mode = title]{Hyperspectral Imaging for Identifying Foreign Objects on Pork Belly} 



%

\author[1]{Gabriela Ghimpeteanu$^{\dagger,}$}[orcid=0000-0001-6005-4703]


\ead{gabriela.ghimpeteanu@coronis.es}


\credit{Methodology, Software, Data Curation, Writing - Original Draft}

\affiliation[1]{organization={Coronis Computing SL},
            addressline={Science and Technology Park of the University of Girona, Carrer Pic de Peguera 11, Edifici Giroemprèn}, 
            city={Girona},
            citysep={}, 
            postcode={17003}, 
            country={Spain}}

\author[2]{Hayat Rajani$^{\dagger,}$}[orcid=0000-0002-2541-2787]
\ead{hayat.rajani@udg.edu}
\credit{Methodology, Software, Data Curation, Writing - Original Draft}

\cormark[1]

\affiliation[2]{organization={Computer Vision and Robotics Research Institute, University of Girona},
            addressline={Campus Montilivi, Edifici P4}, 
            city={Girona},
            citysep={}, 
            postcode={17003}, 
            country={Spain}}

\author[1]{Josep Quintana}[orcid=0000-0003-3577-2011]
\ead{josep.quintana@coronis.es}
\credit{Conceptualization of this study, Software, Supervision, Validation, Writing - Review and Editing}

\author[2]{Rafael Garcia}[orcid=0000-0002-1681-6229]
\ead{rafael.garcia@udg.edu}
\credit{Conceptualization of this study, Project Administration, Supervision, Writing - Review and Editing}

\cortext[1]{Corresponding author}


\nonumnote{$^\dagger$~These authors contributed equally to this work.}


\begin{abstract}
Ensuring food safety and quality is critical in the food processing industry, where the detection of contaminants remains a persistent challenge. This study presents an automated solution for detecting foreign objects on pork belly meat using hyperspectral imaging (HSI). A hyperspectral camera was used to capture data across various bands in the near-infrared (NIR) spectrum (900-1700 nm), enabling accurate identification of contaminants that are often undetectable through traditional visual inspection methods. The proposed solution combines pre-processing techniques with a segmentation approach based on a lightweight Vision Transformer (ViT) to distinguish contaminants from meat, fat, and conveyor belt materials. The adopted strategy demonstrates high detection accuracy and training efficiency, while also addressing key industrial challenges such as inherent noise, temperature variations, and spectral similarity between contaminants and pork belly. Experimental results validate the effectiveness of hyperspectral imaging in enhancing food safety, highlighting its potential for broad real-time applications in automated quality control processes.
\end{abstract}



\maketitle


\section{Introduction}
\label{sec:introduction}

Computer vision solutions have been widely and successfully used in the food processing industry to address numerous challenges arising from quality assessment. These solutions offer economic and safety advantages over manual inspection and, in many cases, yield better results. Over the past decade, the number of industrial computer vision systems has largely increased due to improvements in artificial intelligence, particularly in machine learning methods. These systems have been applied to tasks such as automatic sorting for unwashed eggs and detecting cracks, automatic control of food and drink packaging, and quality analysis of meat, fish, bread, pizza or cheese, as well as the inspection of fruits and vegetables for quality and size \citep{Liu2023}. Human quality inspection is not only expensive but can also be inconsistent and subjective; hence its widespread replacement by computer vision systems. Food safety is a critical aspect of quality control in food factories, leading to the development of several non-destructive, automated techniques. These methods employ hyperspectral and multispectral imaging, metal detection, X-rays, acoustic emission, ultrasound, thermal imaging,  fluorescence spectroscopy, radar and terahertz (THz) imaging or near-infrared (NIR) spectroscopy \citep{Jelali2024}. However, each technology poses certain challenges. Metal detectors are limited to detecting metallic hazards, ignoring other contaminants like plastics or nitrile. Moreover, small metallic contaminants are difficult to detect when embedded in meat, due to the conductive and magnetic properties of the product, which often mask or interfere with the metal detector signal. On the other hand, X-ray systems struggle to detect low-density materials like wood or plastics and carry the risk of ionizing radiation that can potentially alter food products \citep{Zappia2021}, posing health risks to the operators. While safe and fast, infrared systems suffer from high water absorption and limited penetration depth. As for NIR technologies, though non-ionizing and capable of passing through air gaps to detect small anomalies within food structures, only output averaged spectral data, which can be insufficient and challenging for certain applications. On the other hand, thermal imaging is susceptible to temperature interference from neighboring surfaces, while fluorescence spectroscopy is restricted to detecting elements that have a fluorescent compound. THz imaging is the most recent development in non-invasive technology. THz waves, positioned between infrared and microwaves in the electromagnetic spectrum, can penetrate more materials than infrared and provide higher resolution than microwaves. Despite these advantages, THz imaging suffers from slow speeds, susceptibility to noise, and high costs. Traditional optical and electronic devices are also less efficient in detecting THz waves, creating the "terahertz gap" \citep{Zappia2021}.

\begin{figure}[b]
\centering
\begin{tabular}{cc}
 \includegraphics[width=0.8in]{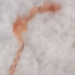} &
 \includegraphics[width=0.8in]{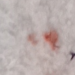} \\
 \fontsize{8.5}{8.5}\textbf{PEHD} &
 \fontsize{8.5}{8.5}\textbf{Blood} \\
\end{tabular}
\caption{RGB images highlighting the similarities between a piece of PEHD and a blood stain on pork belly.}
\label{fig:PEHD_blood}
\end{figure}

Among these non-destructive methods, hyperspectral imaging (HSI) has emerged as a particularly promising technology, capable of overcoming many limitations associated with conventional techniques. In pork processing lines, products can be contaminated with foreign bodies during transportation between the slaughterhouse and packing area, often due to equipment malfunctions or human negligence. These contaminants, which may be as small as a couple of millimeters thin, pose a significant food safety risk as they frequently go unnoticed during human inspection. While RGB cameras are commonly employed in industrial computer vision systems, they struggle to detect such contaminants because of the similar appearance of certain materials. For instance, PEHD plastics and blood stains on fat can appear nearly identical, as depicted in \figref{fig:PEHD_blood}, and materials like Teflon, paper, and sections of the white conveyor belt closely resemble fat. 

In contrast with the above, HSI has proven its versatility in solving diverse problems across various domains such as food safety, agriculture, environmental monitoring and medical diagnosis \citep{Zhu2021}. It has been used to detect the presence of fungi on peaches \citep{Sun2018} and for the quality assessment of fruits such as blueberries \citep{Hu2016}, apples \citep{Tian2018}, pears \citep{Hu2017}, strawberries \citep{Liu2014}, apricots \citep{Buyukcan2017}, and pomegranates \citep{Khodabakhshian2017}. Additionally, hyperspectral imaging has been employed to estimate egg freshness, identify scattered yolks, and detect bubbles inside eggs \citep{Zhang2015}. It has also been applied to detect the adulteration of minced beef with horse meat \citep{Ropodi2017}. Furthermore, hyperspectral imaging has demonstrated impressive results for detection tasks, assessing safety and quality, and grading of nutritional, sensory, and physicochemical qualities in pork, beef, lamb, poultry, and fish \citep{Shi2021, Cheng2017}. These capabilities make HSI particularly well-suited for addressing the challenges of foreign body detection on pork belly meat in industrial settings.


However, despite its inherent advantages, hyperspectral imaging generates abundant data, creating certain challenges for computer vision tasks. The large number of internal parameters and features associated with the hyperspectral cube lead to the ``curse of dimensionality'' problem \citep{Paoletti2019}. Moreover, in the context of machine learning, insufficient training samples can easily result in overfitting. Another critical challenge lies in the high interclass similarity and high intraclass variability in hyperspectral, necessitating sophisticated methods to achieve robust performance.

Early work on hyperspectral classification focused on traditional machine learning techniques such as support vector machines (SVM) \citep{Gualtieri1999}, $k$-nearest neighbor ($k$-NN) \citep{Ma2010}, and multilayer perceptrons (MLPs) \citep{Kalaiarasi2012}. However, the latest advancements in deep learning have introduced more powerful methods for hyperspectral image classification. Deep learning models such as Stacked Autoencoders (SAEs), Recurrent Neural Networks (RNNs) and Deep Belief Networks (DBNs) have demonstrated notable performance. \cite{Hang2017} developed a model with two RNN layers that extract complementary information from non-adjacent spectral bands. While RNN-based models can effectively extract spectral features by treating the spectral signature as a sequence, they often encounter issues with vanishing gradients and struggle to learn long-range dependencies.

Due to their robust ability to capture local information, Convolutional Neural Networks (CNNs) are extremely efficient in extracting informative features from hyperspectral data. Numerous CNN-based models have thus emerged. \cite{Hu2015} proposed a model for HSI classification using several 1D-CNNs that only consider spectral features. However, combining spectral and spatial features yields better performance, as demonstrated by the 3D-CNN model from \cite{Zhong2018} and the 2D-CNN model of \cite{Paoletti}. \cite{Roy2019} introduced HybridSN, which employs a spectral-spatial 3D-CNN followed by a spatial 2D-CNN to model more abstract spatial representations while having a lower complexity compared to a standalone 3D-CNN. Another noteworthy model, SpectralNET \citep{Chakraborty2021}, employs a CNN wavelet to explore both spatial-spectral domains while being computationally less expensive than a 3D-CNN. However, CNNs typically use small convolution kernels, which are effective for capturing local features but struggle to model global dependencies.

Recent works have sought to address this limitation in the context of hyperspectral image classification. \cite{Dang2023} introduced a Double-Branch Feature Fusion Transformer (DBFFT) model, which uses two branches to independently extract spectral and spatial features that are then integrated by a feature fusion layer. Similarly, \cite{Roy2021} proposed an attention-based adaptive spectral-spatial kernel with ResNet backbone. This approach employs spectral attention and adaptive kernel sizes, overcoming the trade-off between small kernels, which capture fine structures but miss coarse details, and large kernels, which capture coarse structures but miss finer details. \cite{ZhuQ2021}, on the other hand, proposed the spectral-spatial dependent global learning (SSDGL) framework to address challenges arising from insufficient and imbalanced HSI data. The framework incorporates a global convolutional long short-term memory module (GCL) and a global joint attention mechanism (GJAM) to improve classification by leveraging both spectral and spatial dependencies.

Despite these advancements, classification models still face challenges, as hyperspectral data often comprise hundreds of bands, with many materials exhibiting very similar spectral signatures. Moreover, existing methods predominantly adopt a patch-based approach, dividing hyperspectral images into small overlapping patches and classifying only the center pixel of each patch. While this increases the number of training samples, it significantly prolongs training time due to the extensive overlap and computational overhead. In contrast, in our method we classify all pixels within each patch while leveraging neighboring context, treating the task as segmentation rather than center-pixel classification. To achieve this, we employ a lightweight Vision Transformer (ViT)-based architecture that progressively refines spectral representations using spatial context by computing self-attention across all pixels within a patch. This design reduces training time, improves efficiency and effectively addresses spectral similarity issues. The following sections describe further details of the proposed approach and the pre-and post-processing strategies adopted for real-time HSI classification, specifically for detecting contaminants on pork belly.


\section{Image acquisition setup}
\label{sec:image_acquisition_setup}

The pork belly pieces are transported between the slaughterhouse and the packing area on a conveyor belt. To identify potential contaminants that may have entered during this process, we set up a Specim FX17 hyperspectral line-scan camera, together with a halogen lighting system, above the conveyor belt to capture hyperspectral images of the pork belly pieces as they move along the belt.

The Specim FX17 \footnote{\url{http://www.specim.fi/products/specim-fx17/}} is a line-scan camera that collects hyperspectral data in the NIR range\textendash a part of the spectrum not visible to the human eye and inaccessible to standard RGB cameras\textendash commonly used in computer vision applications. Imaging in NIR wavelengths allows for the physical and chemical properties of materials to be determined by measuring the proportion of reflected light. Much like a person discerning the strength of tea based on its color intensity, NIR sensors can help differentiate between materials based on the concentration of their spectral properties. Hyperspectral imaging combines spectroscopy with conventional imaging to collect both spatial and spectral information, producing a multidimensional hypercube. The camera captures 224 spectral bands within the NIR range of \SI{1723}[942-]{\nano\meter} at a frame rate of 527 lines per second, with a spectral sensitivity of approximately \SI{3.5}{\nano\meter} per band. This fine granularity allows the system to register subtle material differences. However, the first and the last 20 bands are less informative and noisier due to weaker signal strength. These bands were excluded in our procedure so as to improve data quality, leaving 184 spectral bands. The Specim FX17 camera has a spatial resolution of 640 pixels. In our setup, the camera was mounted orthogonally to the conveyor belt at a distance of \SI{40}{\centi\meter}, measured from the lens to the belt. Positioned this way, the camera's field of view covered a width of \SI{300}{\milli\meter}, resulting in a pixel resolution of \SI{0.47}{\milli\meter}. This resolution ensured that foreign objects with a minimum size of \SI{1}{\square\milli\meter} would be captured by at least two pixels along the line scan, satisfying the detection requirements. Each acquired image comprises a hypercube of dimensions $640 \times 1000 \times 184$.

The lighting system consisted of four commercial halogen bulbs, providing an intensity of 45,000 lux at the belt level. The lights were positioned on either side of the camera along the conveyor belt to ensure consistent illumination and avoid shadow formation on the imaging line. 


\section{Computer vision software}
\label{sec:computer_vision_software}

\begin{figure}[t]
\centering
\begin{tabular}{ccccc}
 \includegraphics[width=0.8in]{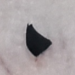} &
 \includegraphics[width=0.8in]{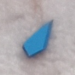} &
 \includegraphics[width=0.8in]{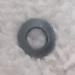} &
 \includegraphics[width=0.8in]{material_PEHD.png} &
 \includegraphics[width=0.8in]{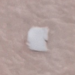} 
 \\
 \fontsize{8.5}{8.5}\textbf{PA-PP} &
 \fontsize{8.5}{8.5}\textbf{PU} &
 \fontsize{8.5}{8.5}\textbf{Metal} &
 \fontsize{8.5}{8.5}\textbf{PEHD} &
 \fontsize{8.5}{8.5}\textbf{Teflon} 
 \\
 \includegraphics[width=0.8in]{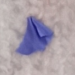} &
 \includegraphics[width=0.8in]{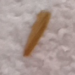} &
 \includegraphics[width=0.8in]{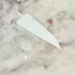} &
 \includegraphics[width=0.8in]{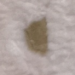} &
 \includegraphics[width=0.8in]{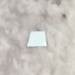} 
 \\
 \fontsize{8.5}{8.5}\textbf{Nitrile} &
 \fontsize{8.5}{8.5}\textbf{Wood} &
 \fontsize{8.5}{8.5}\textbf{Paper} &
 \fontsize{8.5}{8.5}\textbf{Cardboard} &
 \fontsize{8.5}{8.5}\textbf{White Conveyor}
 \\
\end{tabular}
\caption{RGB images depicting the ten classes of contaminants used for this study: PA-PP (PA and PP were grouped into a single class due to the similarity in their spectral signatures), PU, metal, PEHD, Teflon, nitrile, wood, paper, cardboard, and white conveyor belt.}
\label{fig:Materials}
\end{figure}

We framed the problem of foreign body detection as involving the semantic segmentation (or pixel-wise classification) of hyperspectral images captured as the pork belly pieces move along the conveyor belt. The system classifies three materials \textendash meat, fat, and conveyor belt \textendash as negative classes, while all other materials are considered positive detections. For this study, potential contaminants were characterized into ten distinct classes, as illustrated in Figure \ref{fig:Materials}.

\subsection{Data acquisition}
\label{sec:data_acquisition}

\begin{figure}[b]
\begin{tabular}{ccccccc}
\hspace*{-5mm}
 \fontsize{6.5}{6.5}\textbf{Wavelength 1012} &
 \hspace*{-5mm}
 \fontsize{6.5}{6.5}\textbf{Wavelength 1113.5} &
 \hspace*{-5mm}
 \fontsize{6.5}{6.5}\textbf{Wavelength 1218.5} &
 \hspace*{-5mm}
 \fontsize{6.5}{6.5}\textbf{Wavelength 1323.5} &
 \hspace*{-5mm}
 \fontsize{6.5}{6.5}\textbf{Wavelength 1428.5} &
 \hspace*{-5mm}
 \fontsize{6.5}{6.5}\textbf{Wavelength 1533.5} &
 \hspace*{-5mm}
 \fontsize{6.5}{6.5}\textbf{Wavelength 1638.5} \\
 \hspace*{-5mm}
 \includegraphics[width=0.9in]{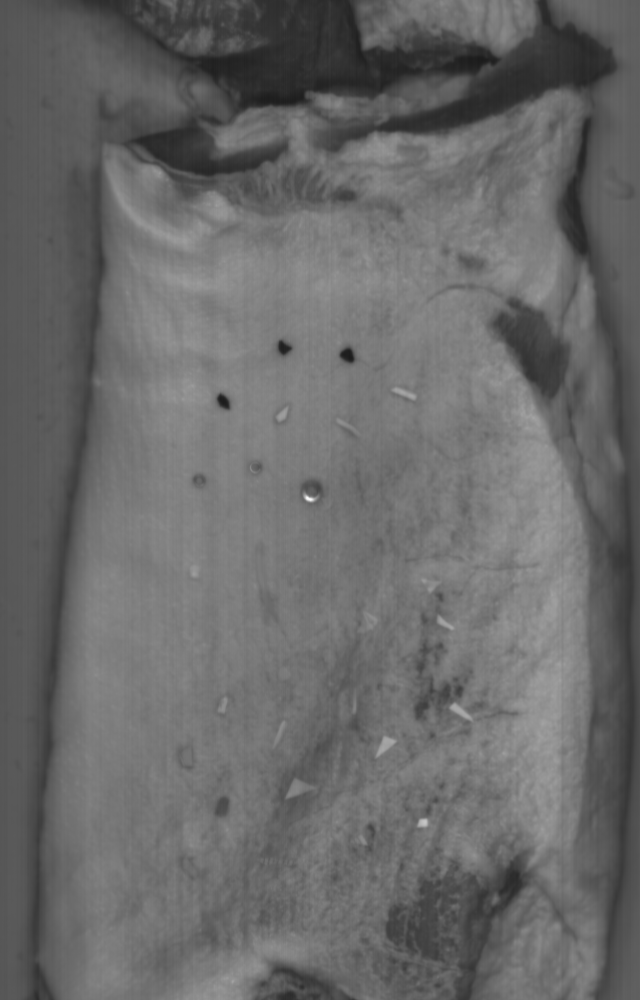} &
 \hspace*{-5mm}
 \includegraphics[width=0.9in]{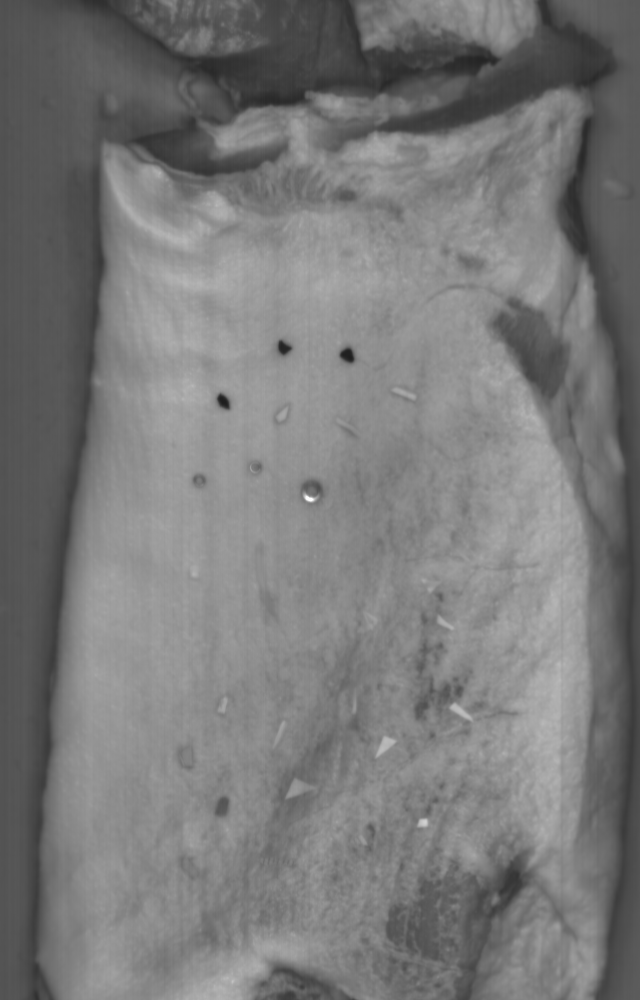} &
 \hspace*{-5mm}
 \includegraphics[width=0.9in]{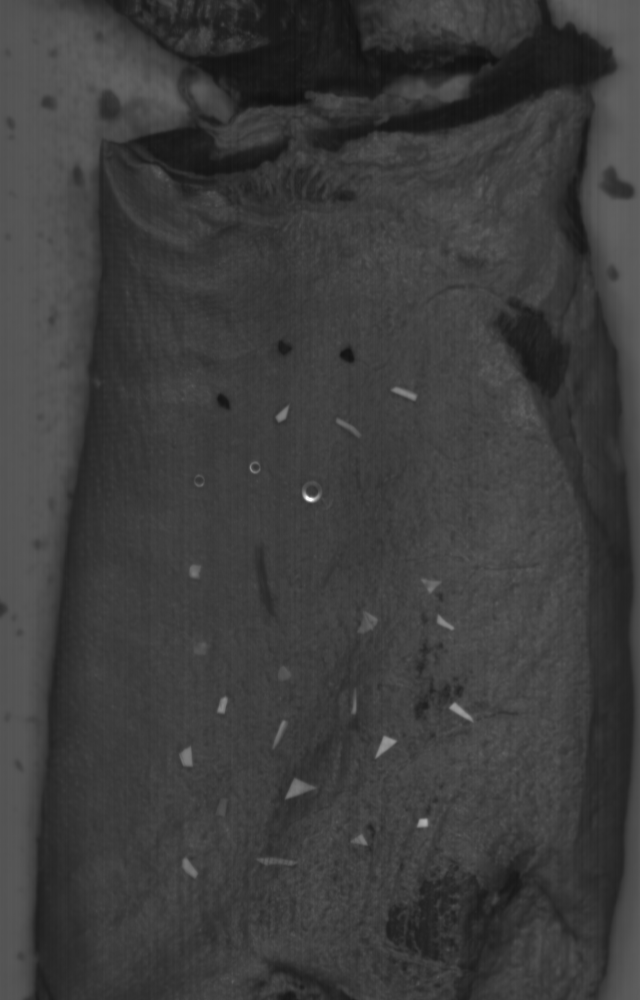} &
 \hspace*{-5mm}
 \includegraphics[width=0.9in]{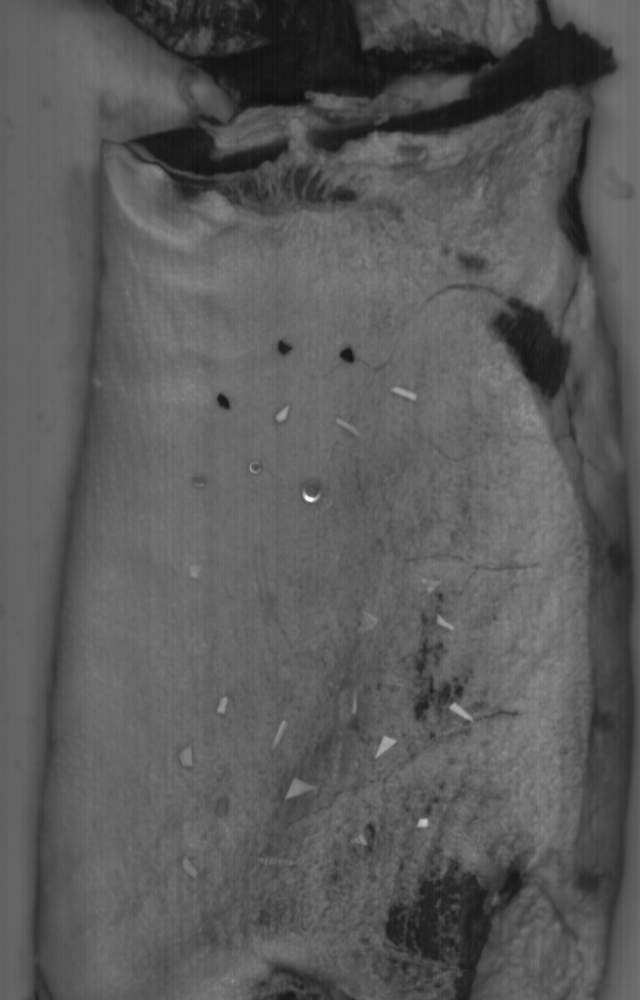} &
 \hspace*{-5mm}
 \includegraphics[width=0.9in]{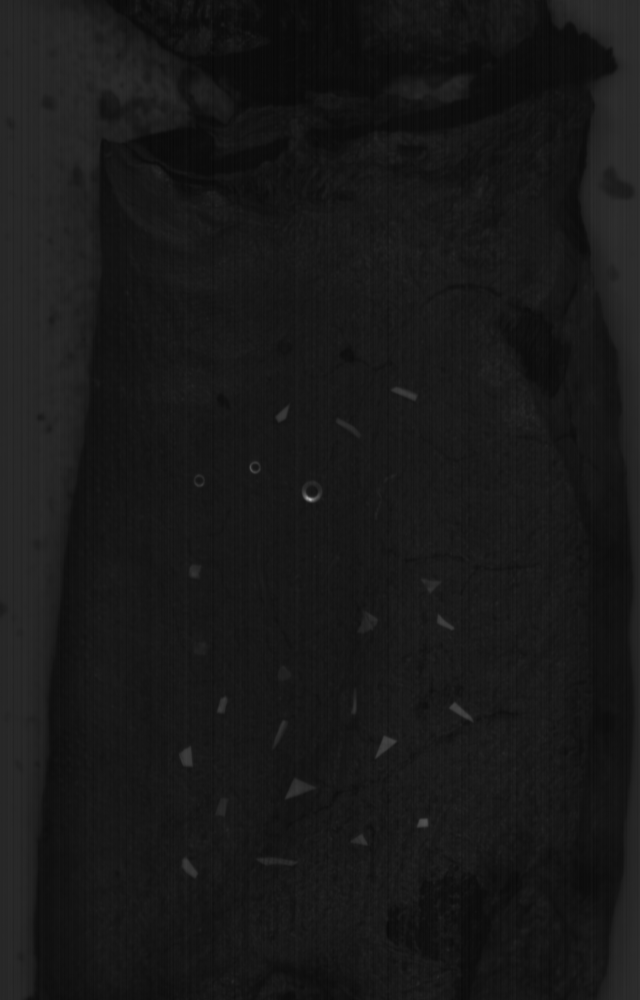} &
 \hspace*{-5mm}
 \includegraphics[width=0.9in]{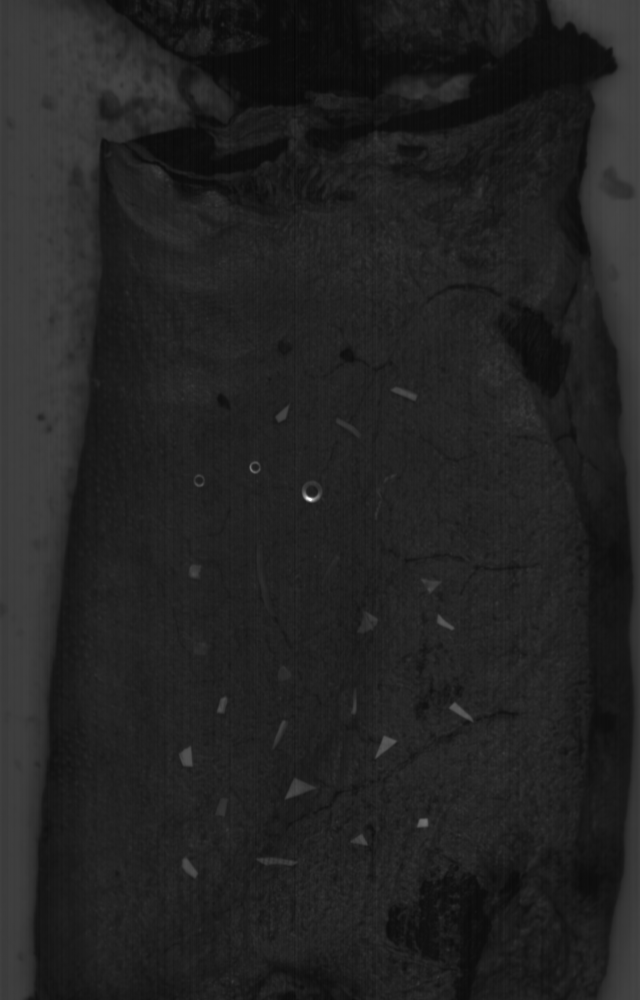} &
 \hspace*{-5mm}
 \includegraphics[width=0.9in]{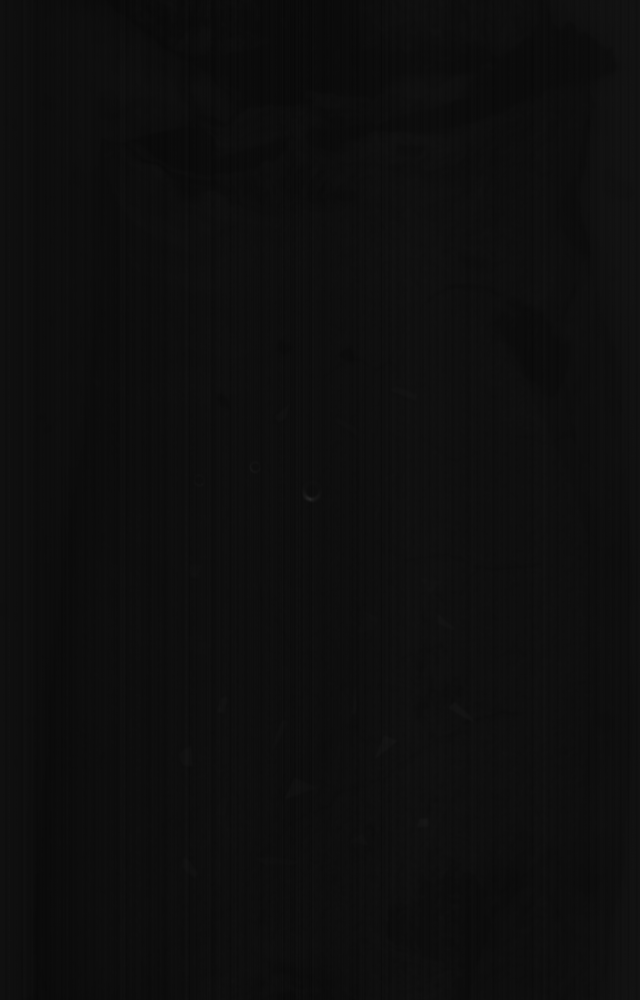} 
 \vspace*{-0.8mm}
 \\ 
 \hspace*{-5mm}
 \includegraphics[width=0.9in]{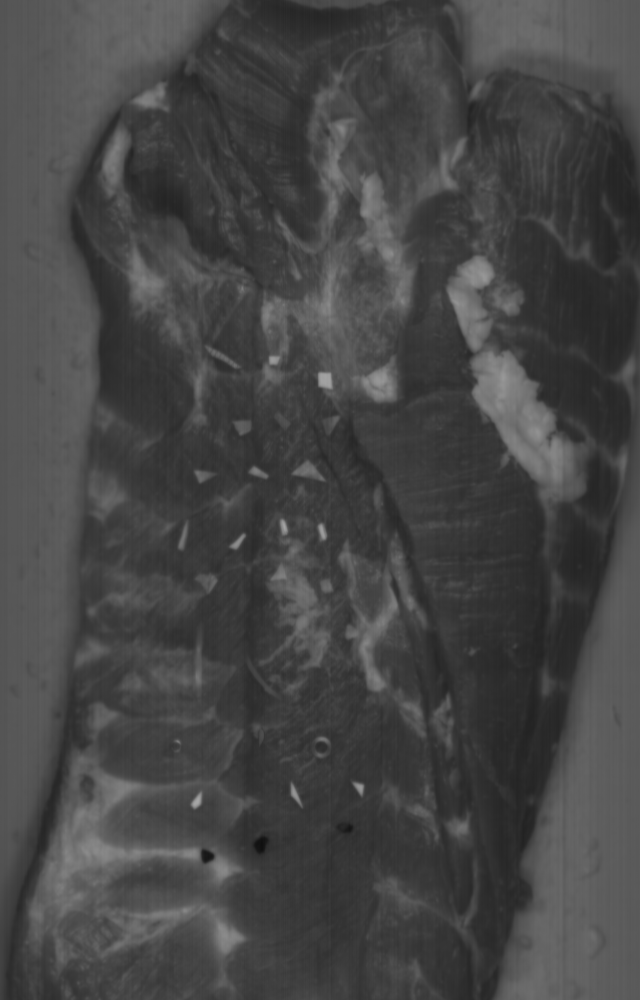} &
 \hspace*{-5mm}
 \includegraphics[width=0.9in]{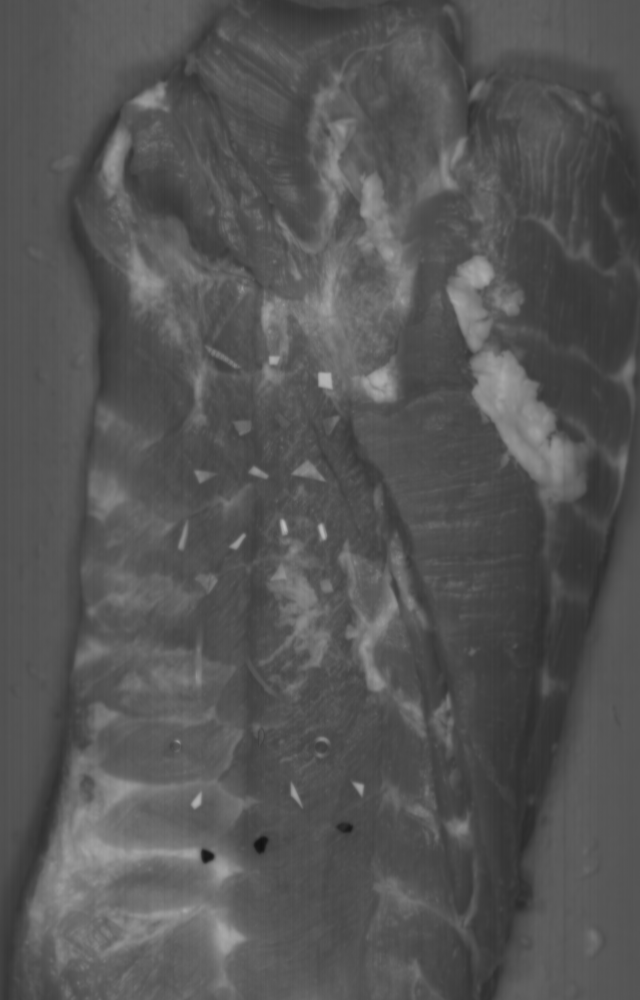} &
 \hspace*{-5mm}
 \includegraphics[width=0.9in]{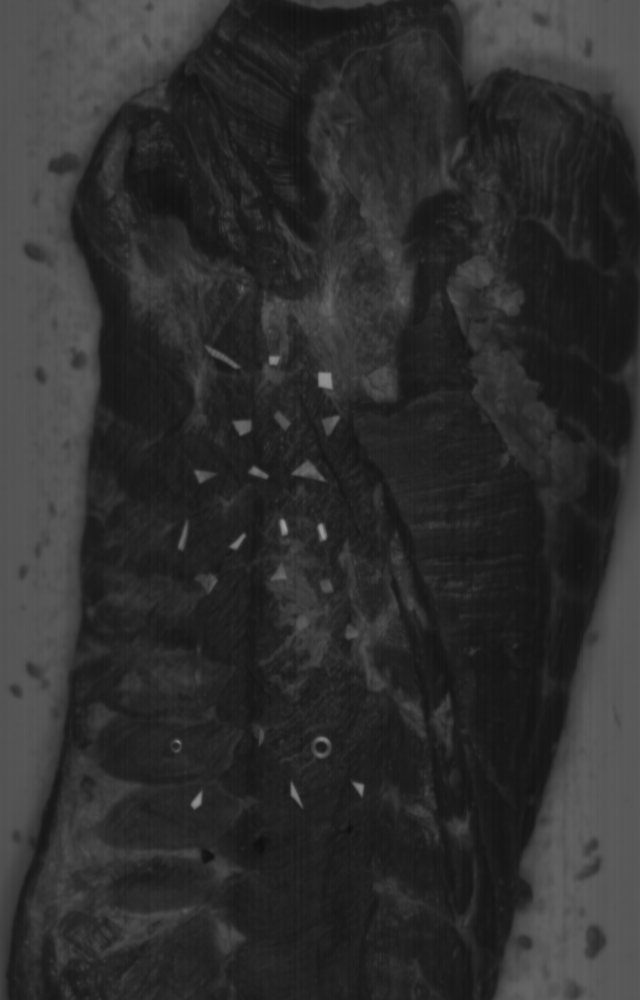} &
 \hspace*{-5mm}
 \includegraphics[width=0.9in]{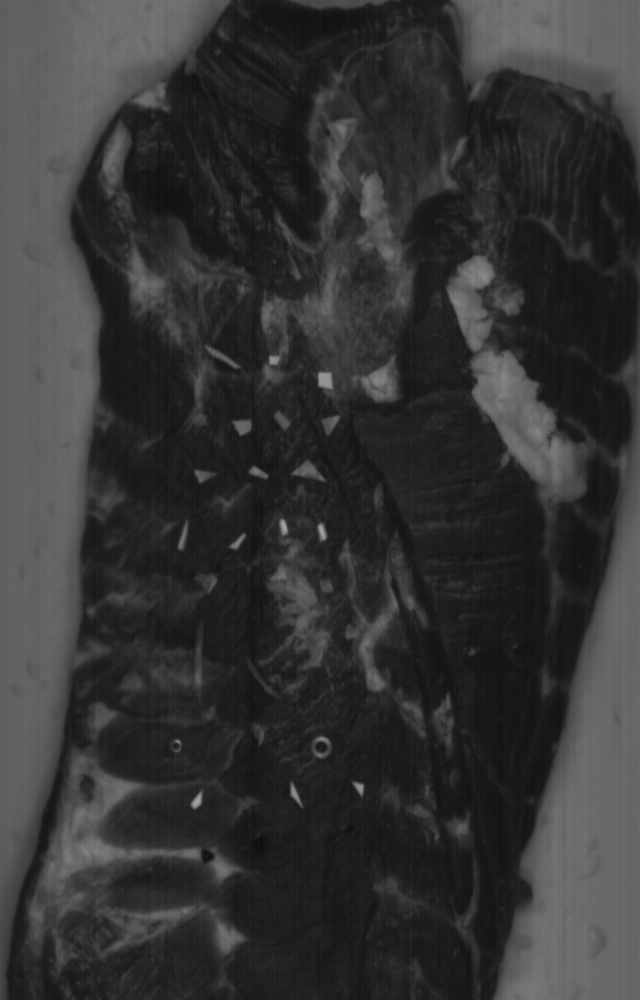} &
 \hspace*{-5mm}
 \includegraphics[width=0.9in]{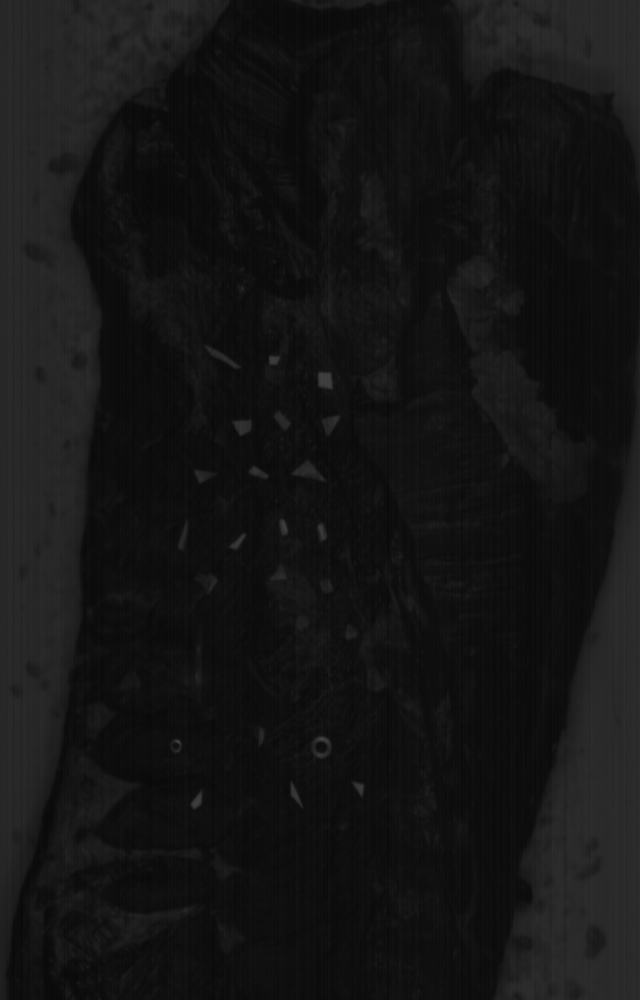} &
 \hspace*{-5mm}
 \includegraphics[width=0.9in]{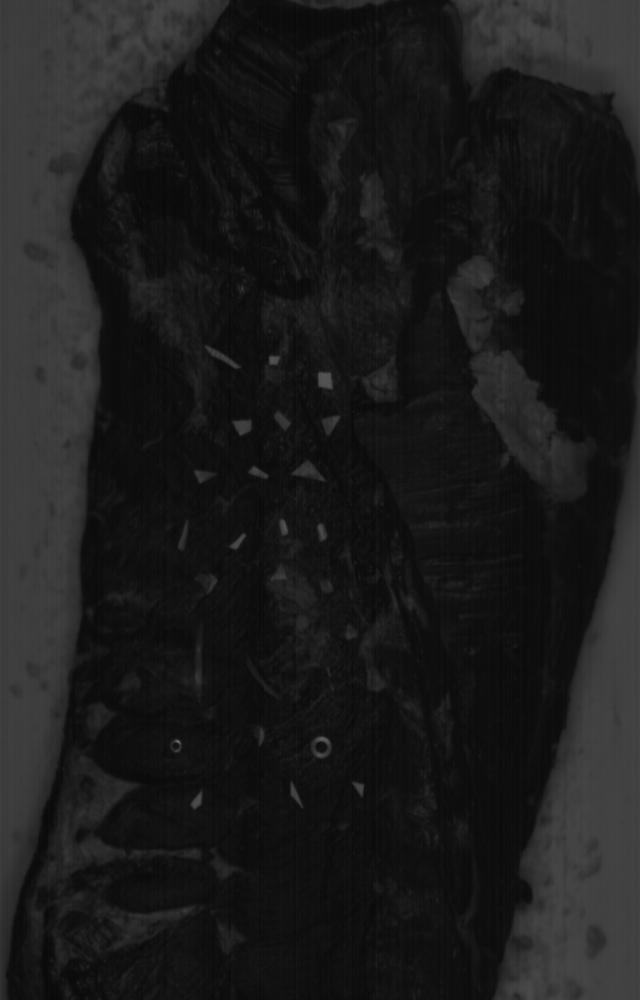} &
 \hspace*{-6mm}
 \includegraphics[width=0.9in]{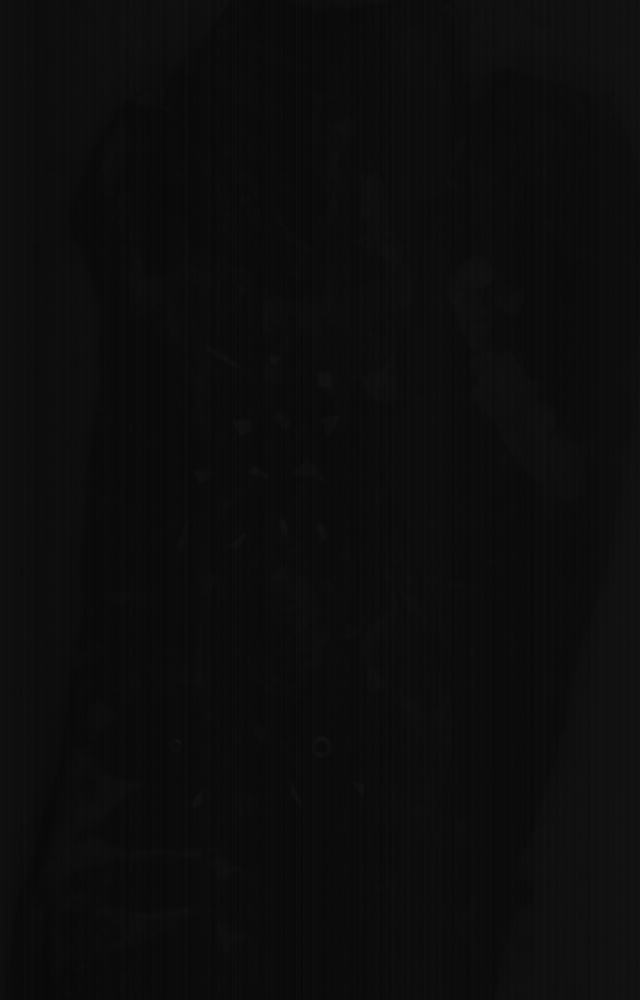}
 \\
\end{tabular}
\caption{Examples of two hyperspectral images: one with fat side up (top row) and the other with meat side up (bottom row). Each image depicts seven of the 184 spectral bands, acquired at wavelengths \SI{1012}{\nano\meter}, \SI{1113.5}{\nano\meter}, \SI{1218.5}{\nano\meter}, \SI{1323.5}{\nano\meter}, \SI{1428.5}{\nano\meter}, \SI{1533.5}{\nano\meter} and \SI{1638.5}{\nano\meter}.}
\label{fig:spectral_example}
\end{figure}

We collected 78 hyperspectral images of pork belly with contaminants and 183 images without contaminants, in order to generate the training and testing sets for the machine learning models. The images were captured under the same setup as described in \secref{sec:image_acquisition_setup}. Both orientations of the pork belly \textendash fat side up and meat side up \textendash were considered, as depicted in Figure \ref{fig:spectral_example}. 

From the 78 hyperspectral images of pork belly with contaminants, we manually annotated 22 images to train the segmentation model. The remaining 56 images were used for testing. It should be noted that the detection of fine contaminants was particularly challenging due to the surrounding material often influencing their spectral response. The mixed spectral signature between the contaminant and the surrounding material made it difficult to accurately determine the extent of the contaminant, leading to some imprecision in the annotated regions. Each image was divided into $20 \times 16$ patches with a 50\% overlap for the patch-based segmentation approach, resulting in a total of around 80,000 patches in the training set.

\begin{figure}[b]
\centering
\begin{tabular}{cccc}
 \textbf{RGB} &
 \textbf{Wavelength \SI{1082}{\nano\meter}} &
 \textbf{Annotated} &
 \\
 \vstretch{1.176}{\includegraphics[width=0.875in]{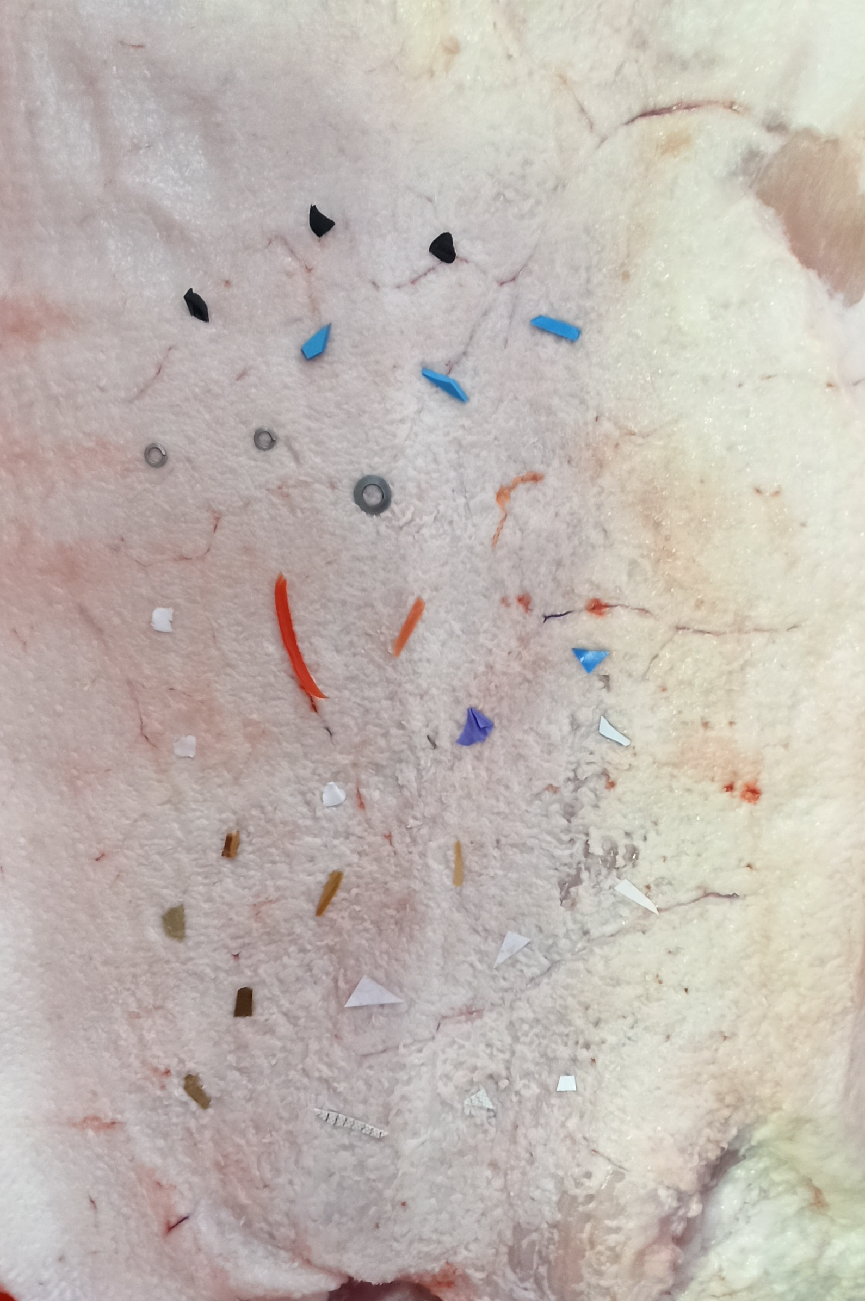}} &
 \vstretch{1.2}{\includegraphics[width=1in]{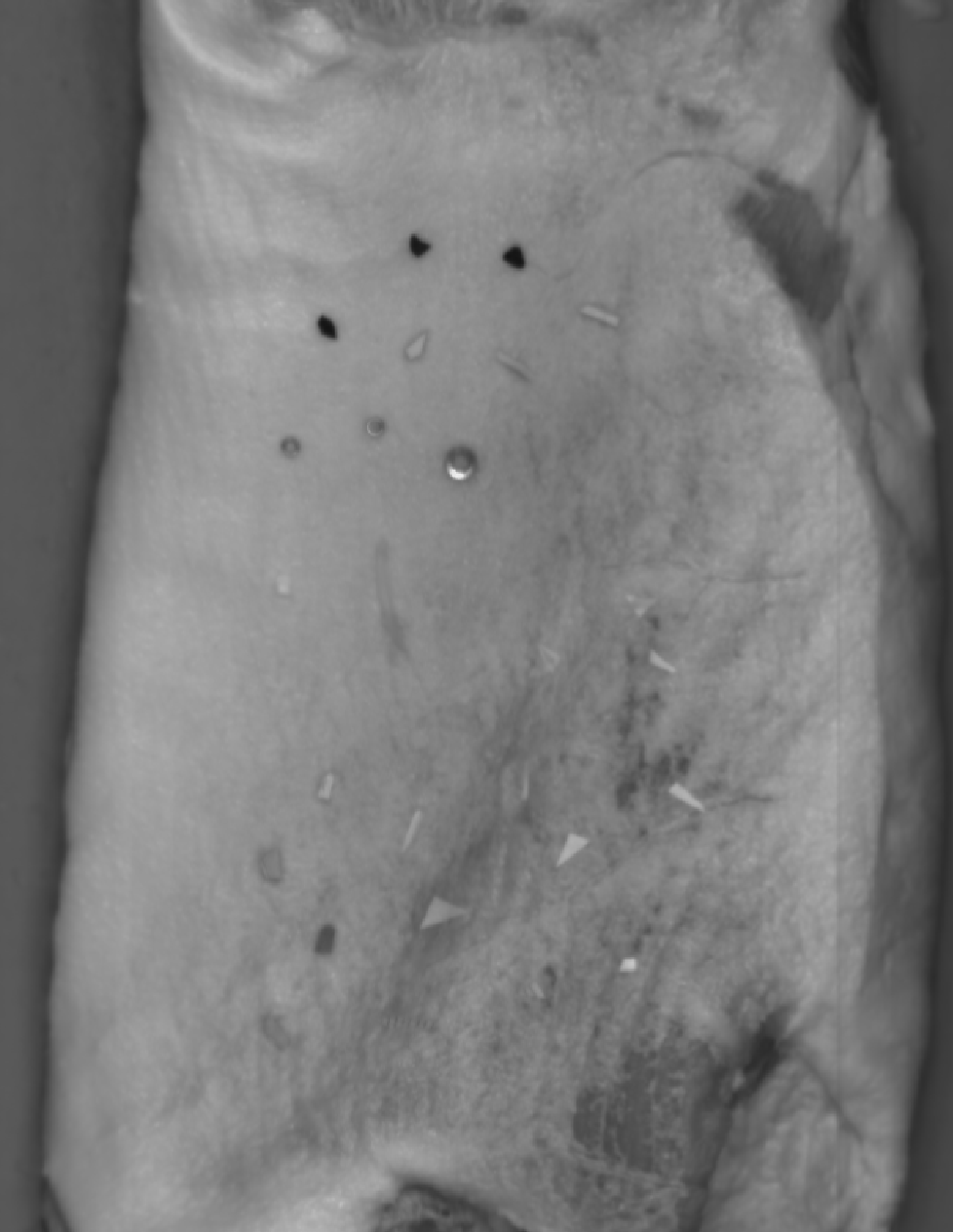}} &
 \vstretch{1.2}{\includegraphics[width=1in]{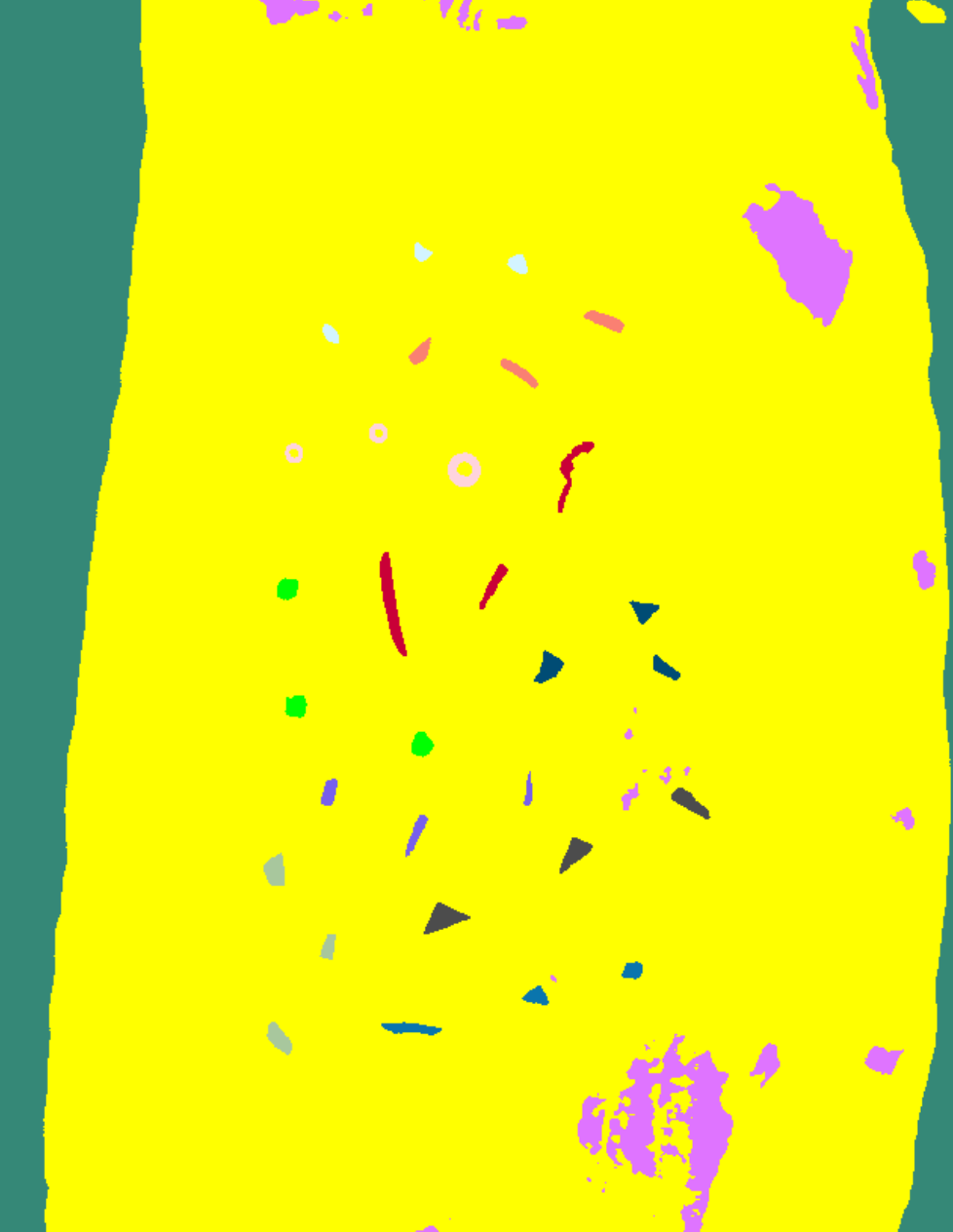}} & 
  \\
 \vstretch{1.128 }{\includegraphics[width=0.875in]{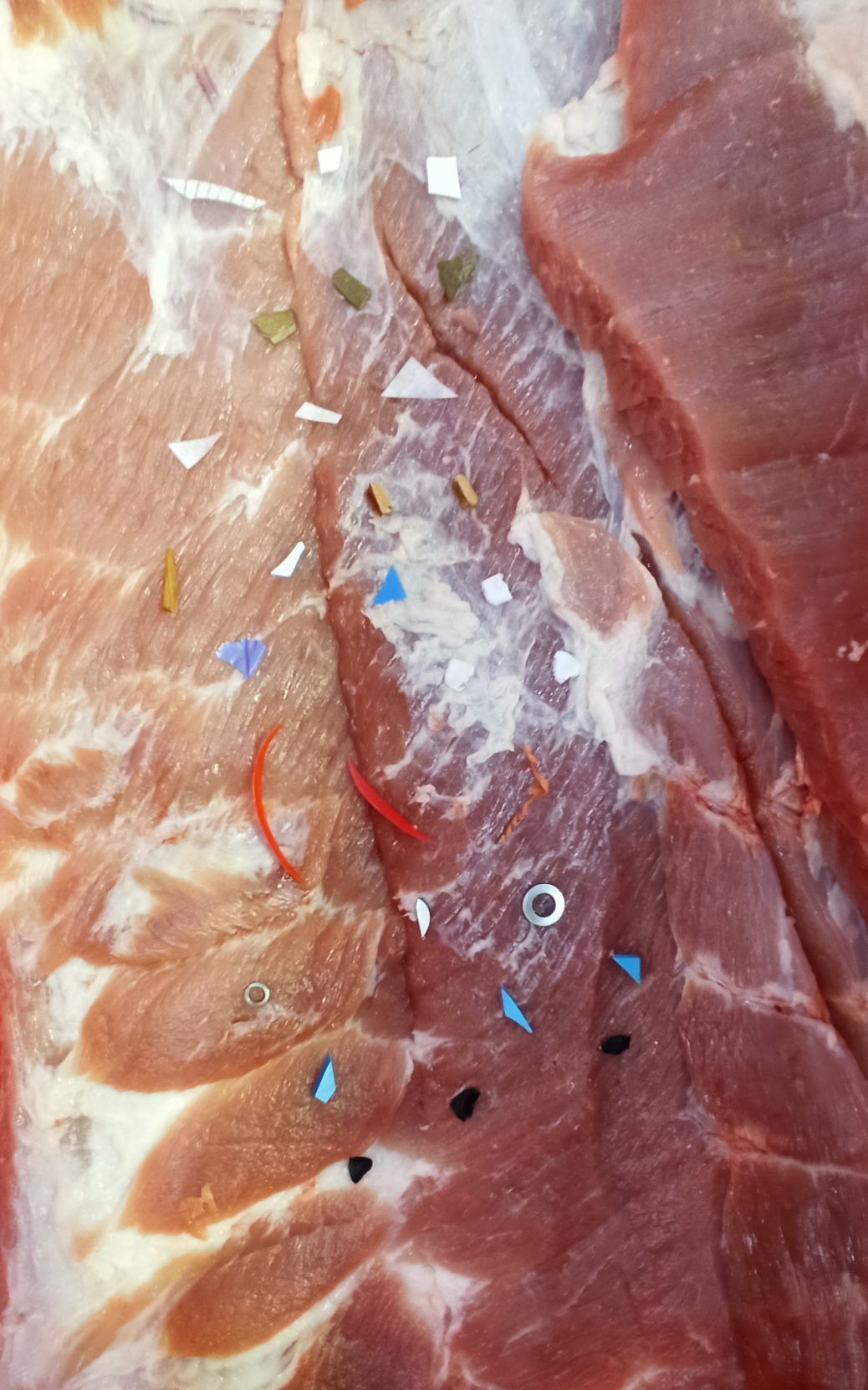}} &
 \vstretch{1.044}{\includegraphics[width=1in]{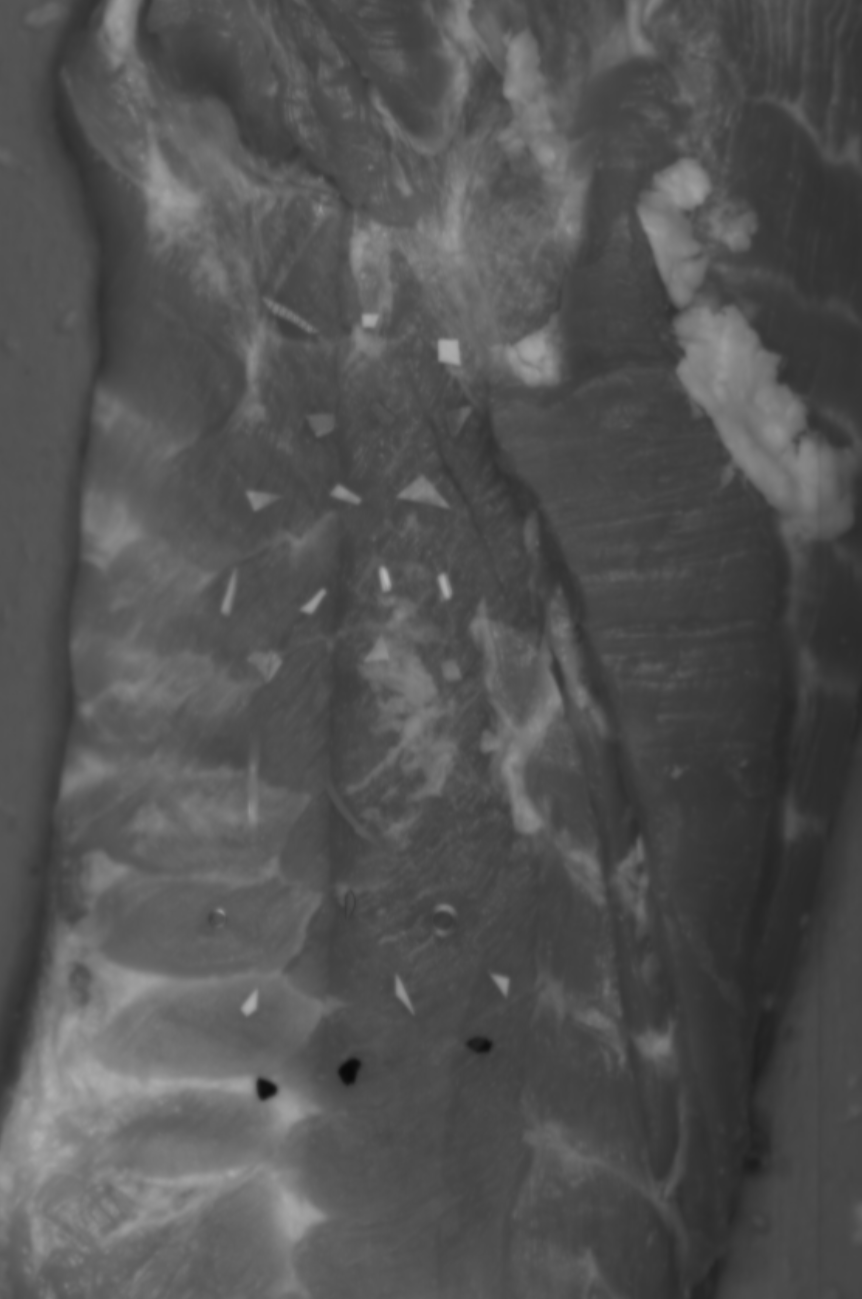}} &
 \vstretch{1.044}{\includegraphics[width=1in]{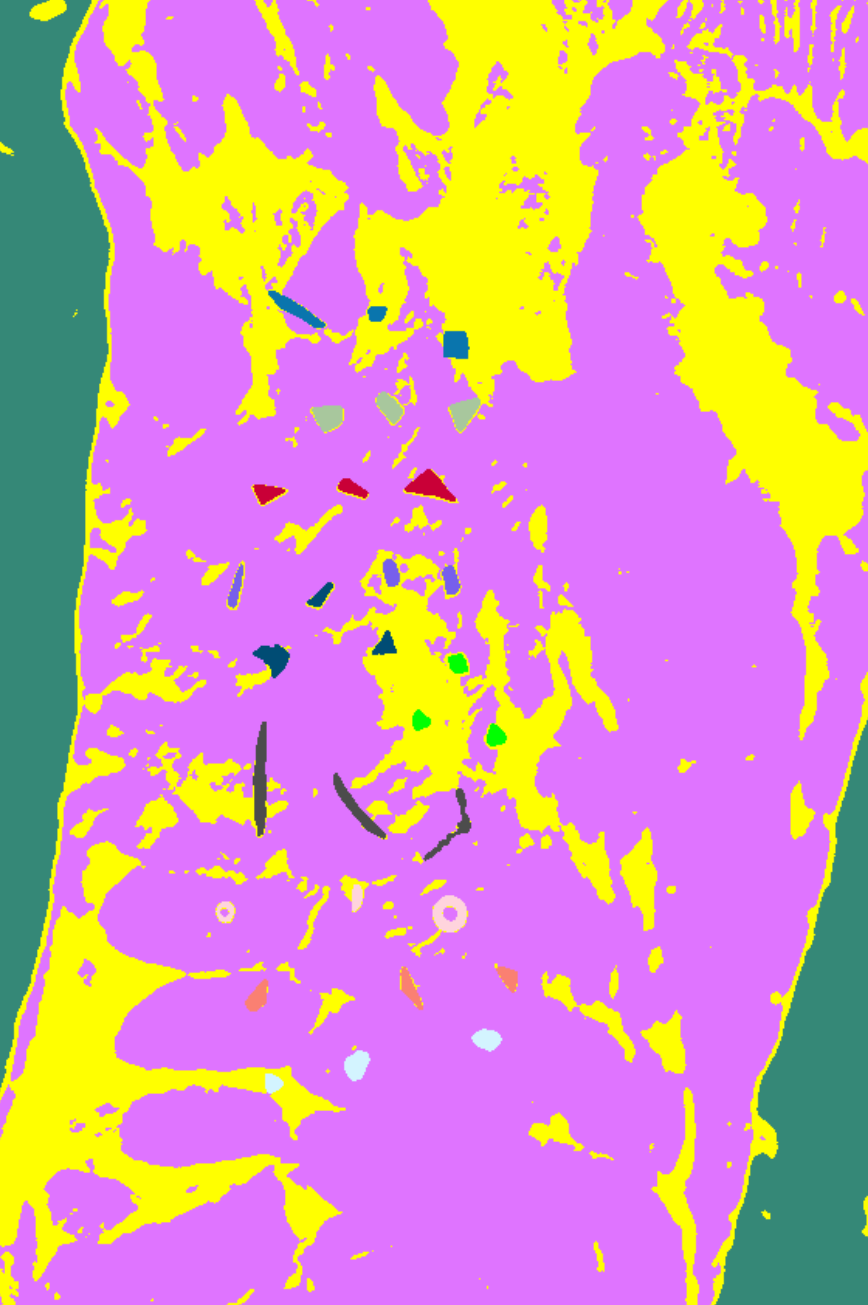}}
 \\
\end{tabular}
\hspace*{-6mm}
 \centering
\begin{tabular}{c}
 \includegraphics[width=1.2in]{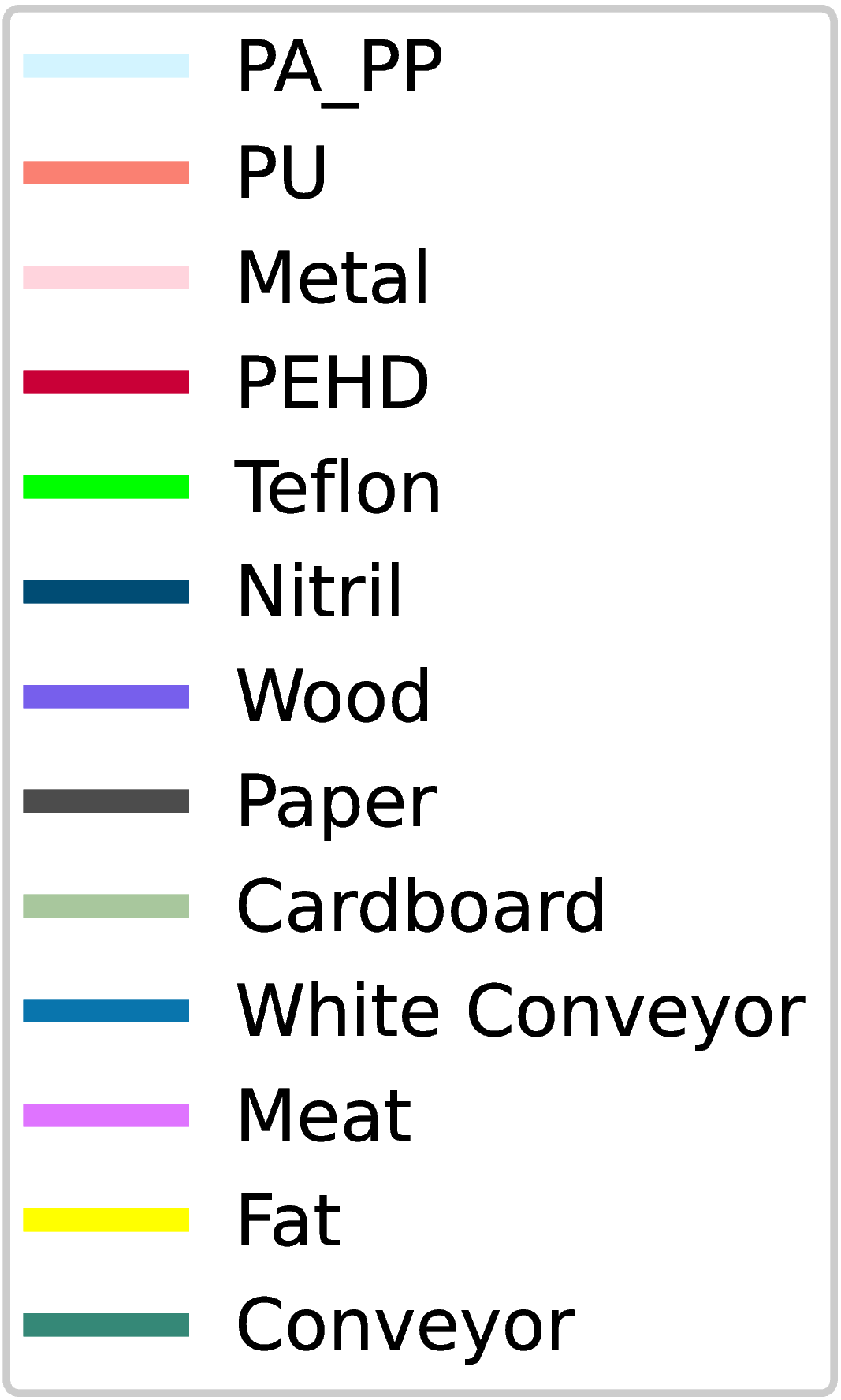}
 \end{tabular}
\caption{Examples of annotated data.}
\label{fig:annotations}
\end{figure}

\figref{fig:annotations} illustrates examples of the annotated images. The first column depicts RGB images for better visualization, captured using a standard consumer-grade camera. The second column shows one of the 184 spectral frequency gray-level images that made up a hyperspectral cube, and the corresponding annotated segmentation masks in the third column. The last column introduces the color associated with each material for annotation.

\begin{figure}[t]
\centering
\begin{tabular}{cc}
 \hspace*{-6mm}
 \includegraphics[width=3.33in]{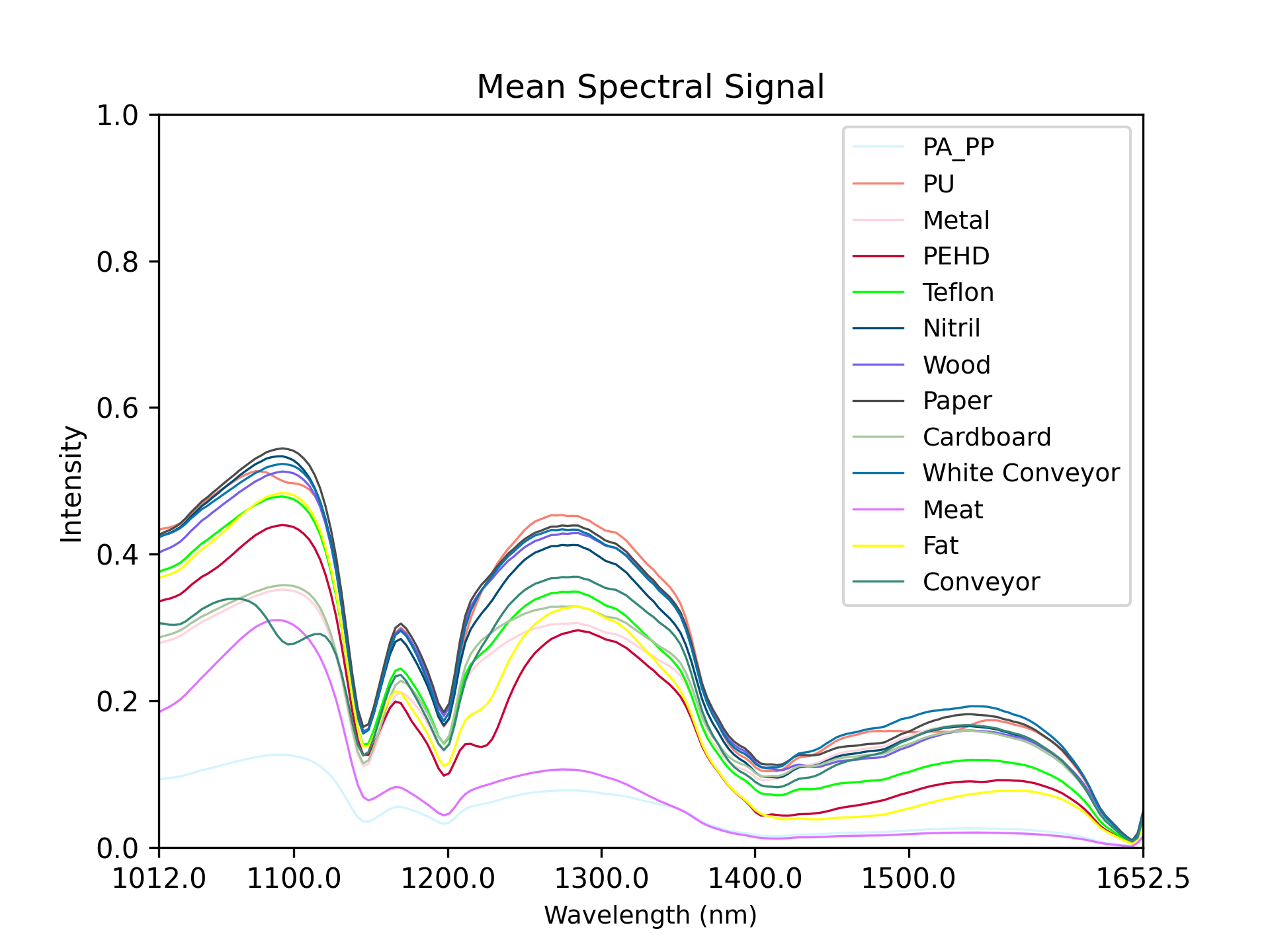} &
 \hspace*{-12mm}
 \includegraphics[width=3.33in]{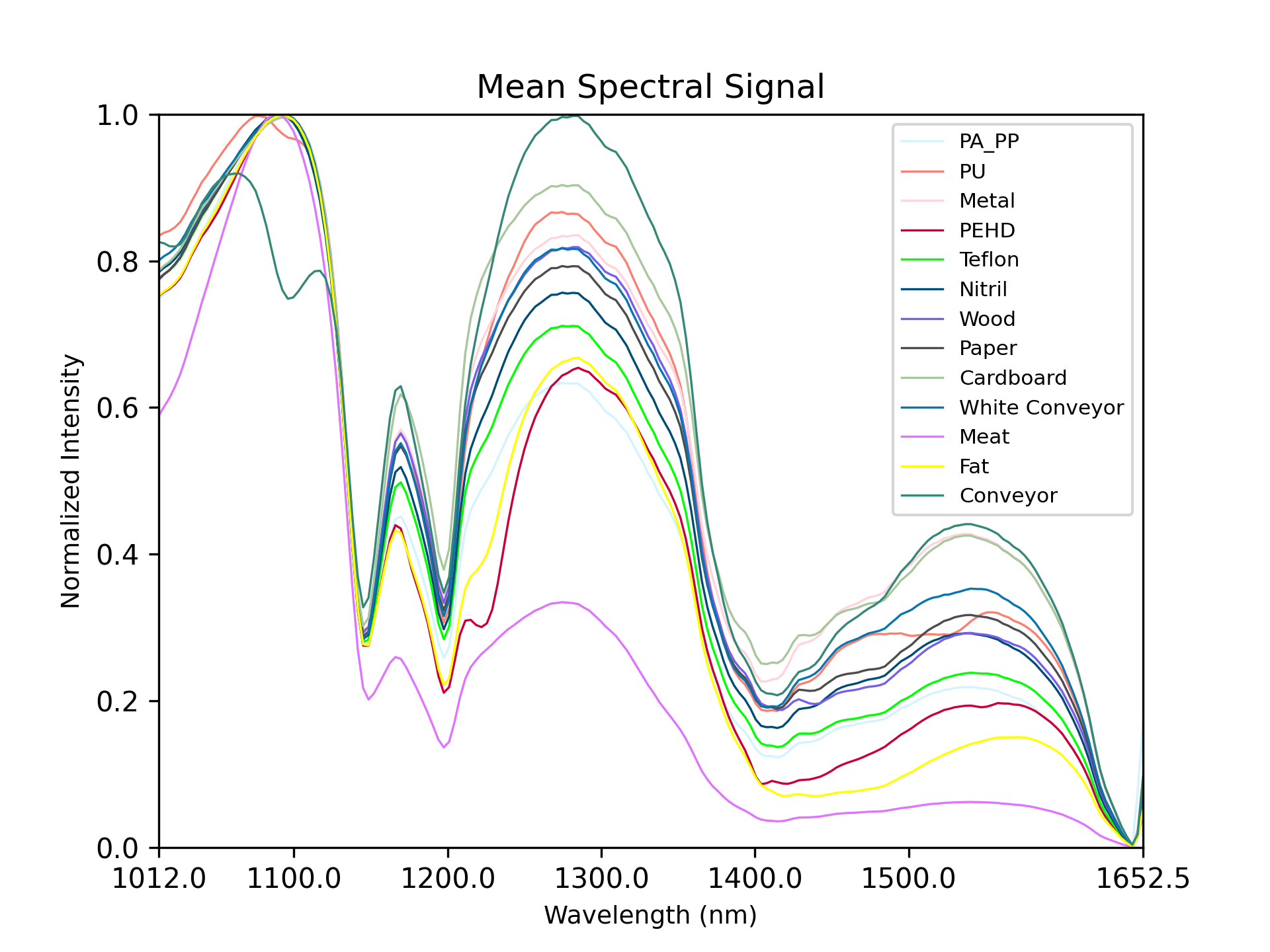} \\ 
\end{tabular}
\caption{Hyperspectral mean intensity curve plots for our 13 test materials after flat-field correction, before (left) and after normalization (right).}
\label{fig:plot_materials_hyperspectral_curves}
\end{figure}

\figref{fig:plot_materials_hyperspectral_curves} presents the mean spectral signatures of the annotated materials.

\subsection{Hyperspectral image processing}
\label{sec:image_processing}

Before feeding the hyperspectral images to the machine learning model, they were processed in two stages. The first stage involved flat-field correction to eliminate fixed-pattern vertical stripe noise, followed by pixel-wise normalization. These preprocessing steps ensured invariance under variations in camera sensor temperature and exposure parameters.

Fixed pattern noise is temporally coherent and occurs due to pixel-to-pixel sensitivity variations in the sensor array, which are caused by factors such as differences in the cell response to incoming light, read-out and amplification electronics or thermal noise, making system calibration a challenge. Image stripes can also be caused by dirt or dust on the sensor. Flat field correction was applied to address these issues. The process involved capturing two frames: a dark one, acquired with the shutter closed, to measure the response of the detector in the absence of incident light, and a flat field image (gain frame) from a homogeneous surface to measure the response of each photo-diode to incoming light. The dark frame helped estimate dark currents for calibration, while the flat-field image allowed for the correction of variations in illumination and detector response. A Teflon tile was used as a homogeneous surface. After subtracting the dark field (one image or an average of several), each pixel should output the same black value. 
After flat-field correction, the image of a homogeneous surface captured under uniform illumination should appear flat. The flat-field correction process is described by the following equation \citep{Seibert1998}:

\begin{math}
C = \frac{(I-D)m}{F-D},
\end{math}
\\
where $C$ is the calibrated image, $I$ is the acquired image to be calibrated, $D$ is the dark frame, $F$ is the flat field image and $m$ is the average pixel value of the corrected flat field image $F-D$.  

Both dark and gain frames must be acquired using the same ISO, exposure settings, and illumination conditions as the image being calibrated. Using $G$ to denote gain, the equation for flat-field correction can be rewritten as:

\begin{math}
G = \frac{m}{F-D},
\end{math}
\\
which simplifies the flat-field correction to:

$ C = (I-D) G $.

The gain, $G$, measures how much the corrected flat field, $F-D$, deviates from its mean and from being flat. 

\begin{figure}[t]
\centering
\begin{tabular}{cccc}
 \includegraphics[width=1in]{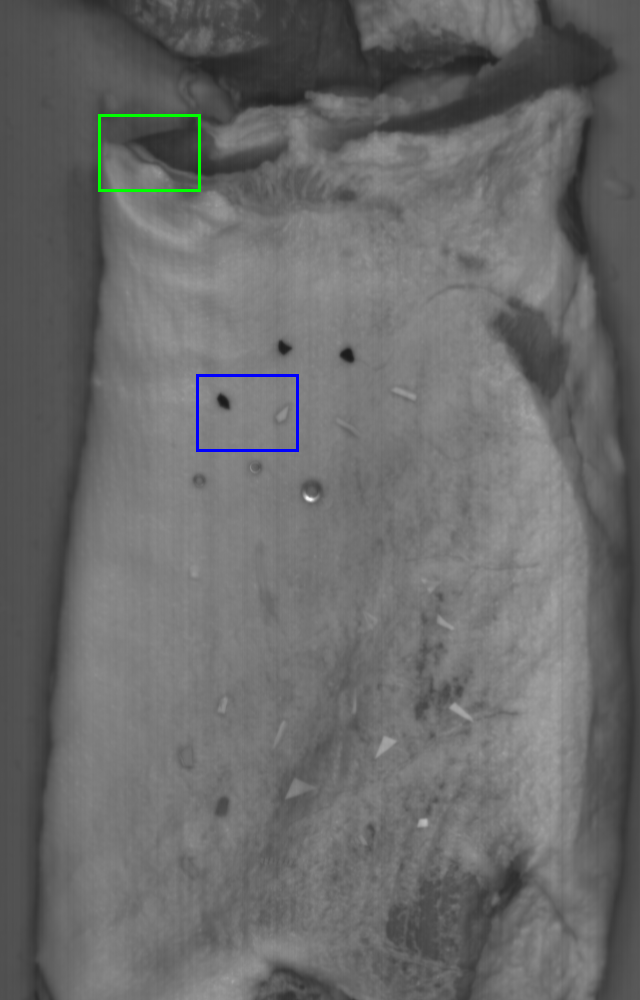} &
 \includegraphics[width=1in]{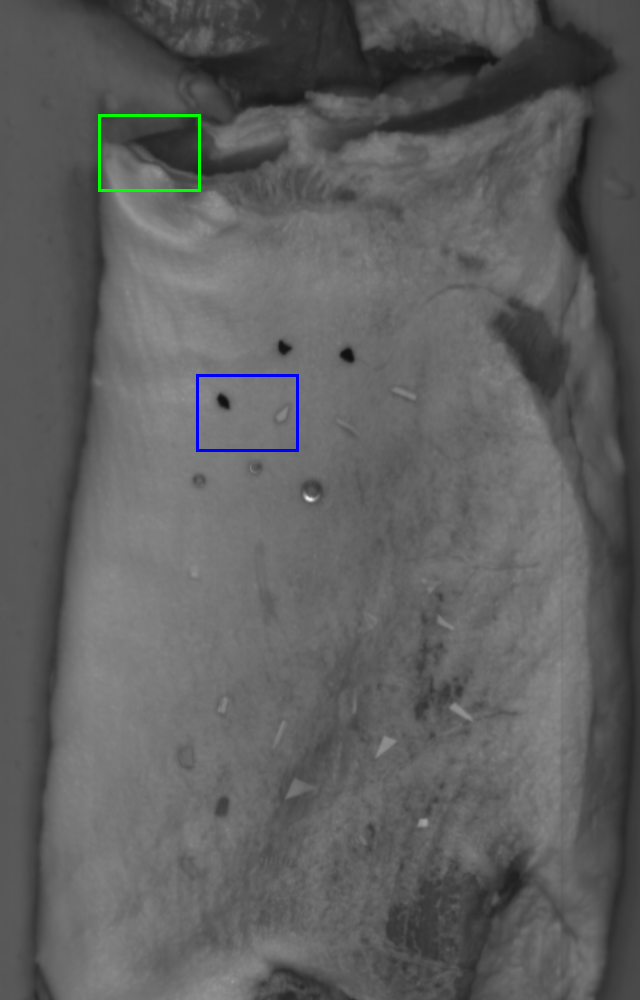} &
 \includegraphics[width=1in]{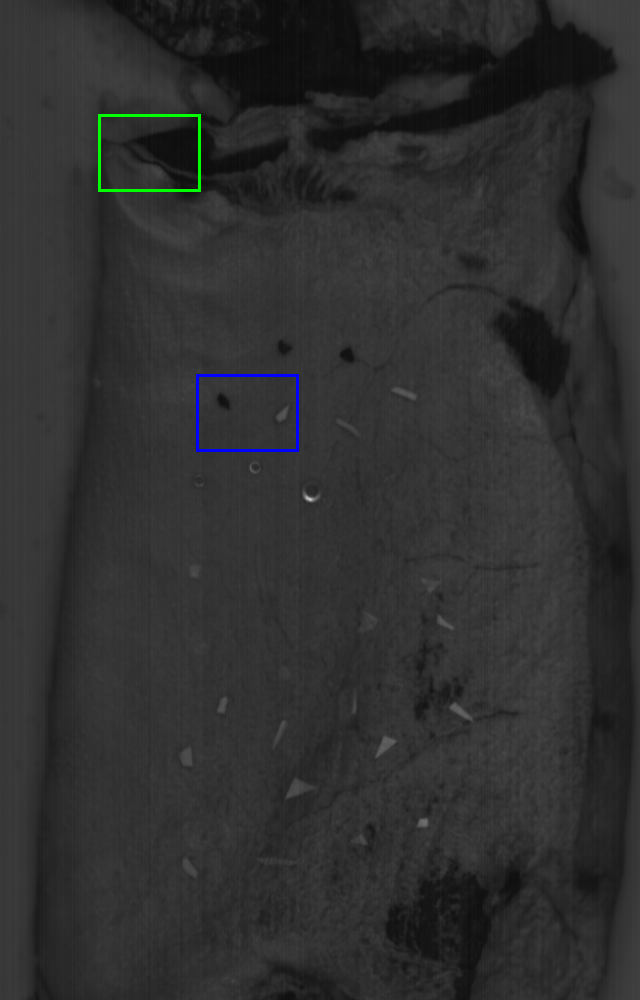} &
 \includegraphics[width=1in]{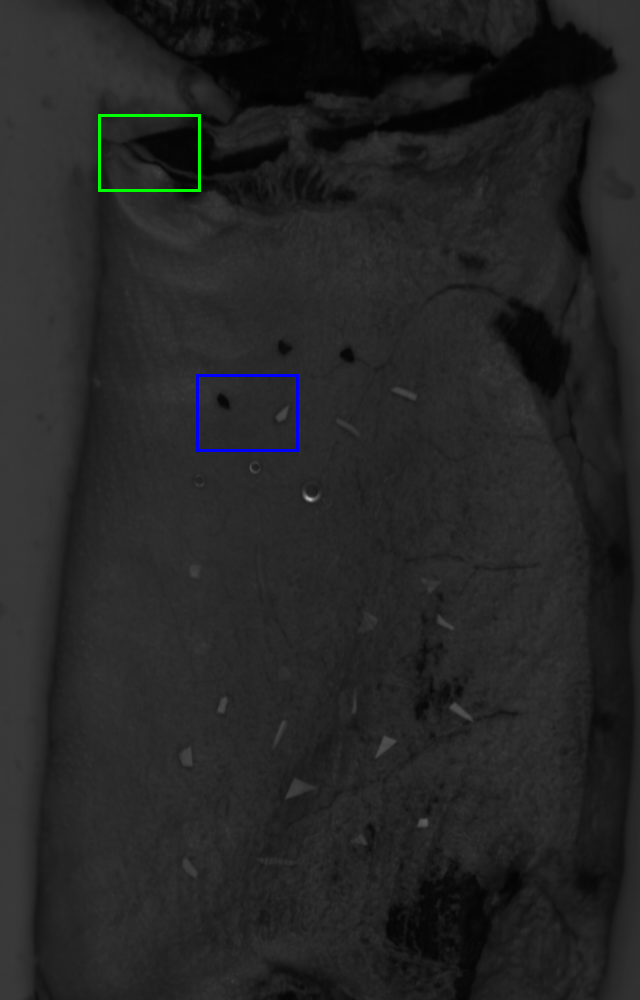}
 \\
\includegraphics[width=1in]{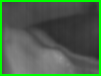} &
 \includegraphics[width=1in]{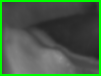} &
 \includegraphics[width=1in]{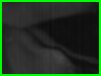} &
 \includegraphics[width=1in]{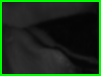}
 \\
 \includegraphics[width=1in]{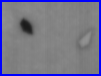} &
 \includegraphics[width=1in]{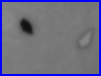} &
 \includegraphics[width=1in]{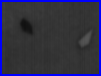} &
 \includegraphics[width=1in]{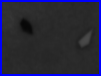}
 \\
\textbf{(a)} &
 \textbf{(b)} &
 \textbf{(c)} &
 \textbf{(d)} 
 \\
\fontsize{7}{7}\textbf{Wavelength \SI{1124}{\nano\meter}} &
 \fontsize{7}{7}\textbf{After FFC} &
 \fontsize{7}{7}\textbf{Wavelength \SI{1369}{\nano\meter}} &
 \fontsize{7}{7}\textbf{After FFC} 
 \\
\end{tabular}
\caption{Flat-Field Correction eliminates vertical fixed pattern stripe noise.}
\label{fig:FFC}
\end{figure}

\figref{fig:FFC} illustrates an example of a hyperspectral image with visible fixed-pattern noise in columns (a) and (c), appearing as vertical stripes. Columns  (b) and (d) depict the results after applying flat-field correction, with a reduced noise pattern. The crops highlight how the noise pattern varies across different wavelength images in columns (a) and (c).

\begin{figure}[b]
\centering
\fontsize{7}{7}\textbf{Wavelength \SI{1176.5}{\nano\meter} processed with Flat-Field Correction}
\\
\vspace*{1mm}
\begin{tabular}{ccc}
  \includegraphics[width=1in]{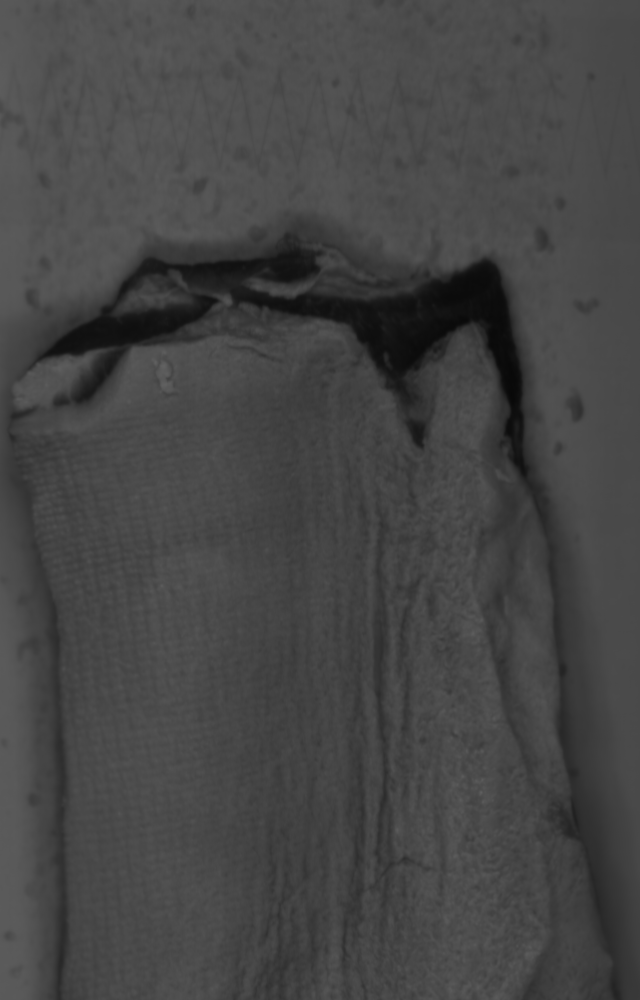} &
 \includegraphics[width=1in]{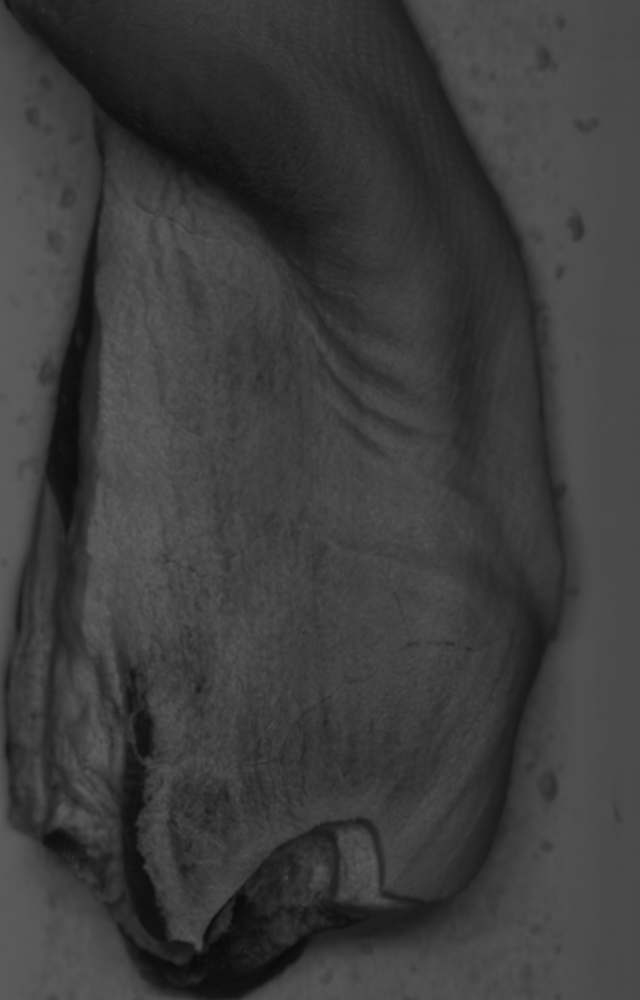} &
 \includegraphics[width=1in]{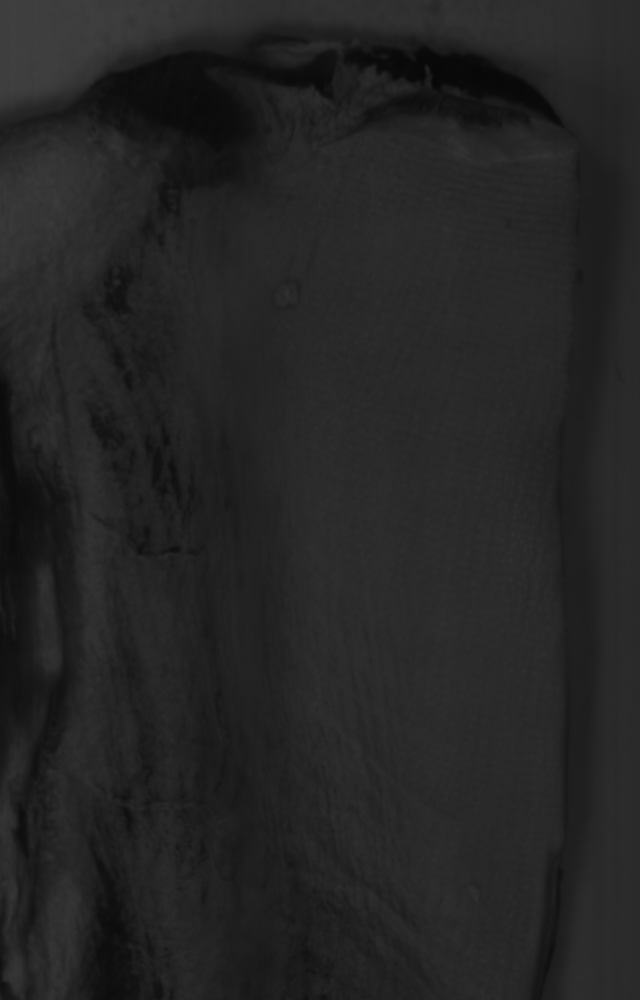}
 \\
 \textbf{(a)} &
 \textbf{(b)} &
 \textbf{(c)}
 \\
 \end{tabular}\\
 \vspace*{3mm}
 \fontsize{7}{7}\textbf{After Normalization}
  \\
  \vspace*{1mm}
  \begin{tabular}{ccc}
  \includegraphics[width=1in]{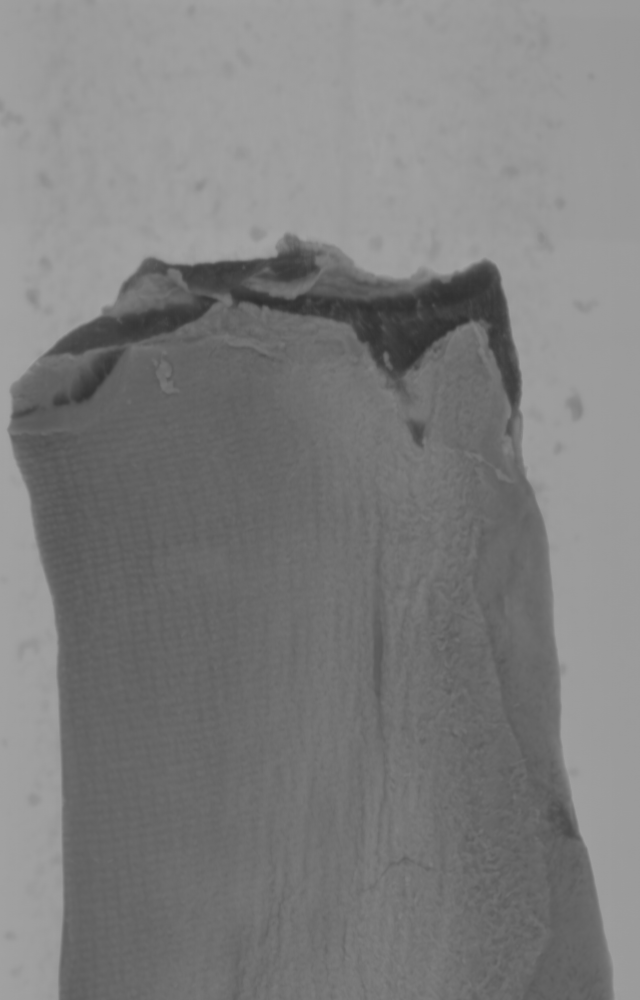} &
 \includegraphics[width=1in]{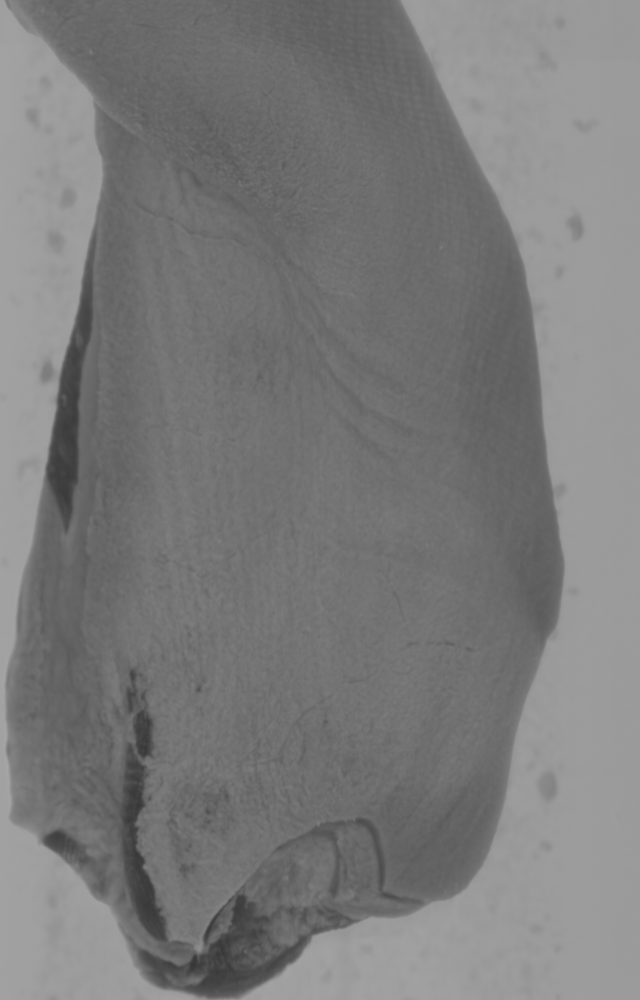} &
 \includegraphics[width=1in]{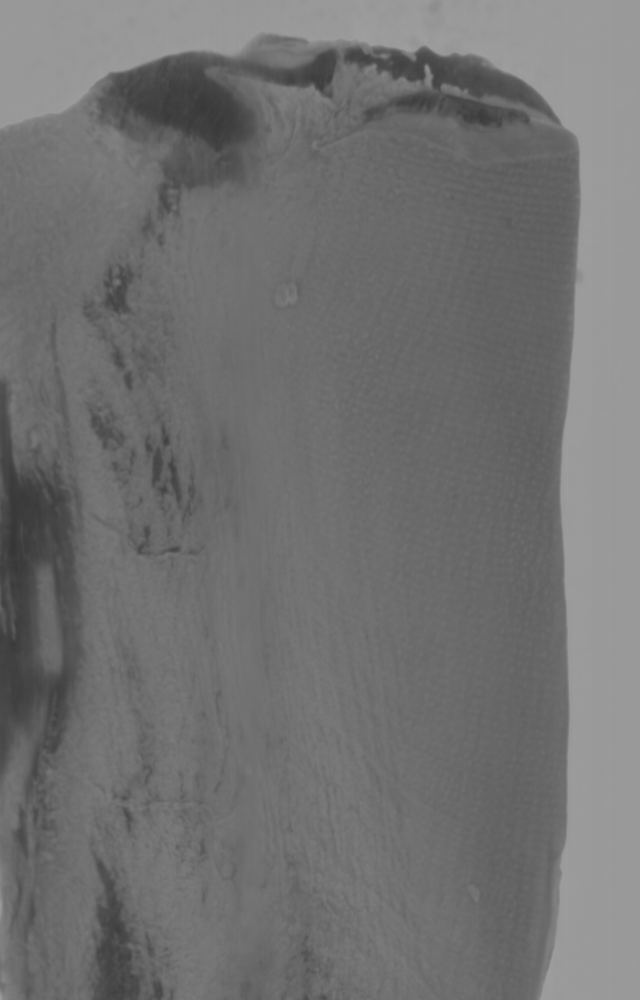}
 
\end{tabular}\\
\caption{Normalization produces similar hypercubes for hyperspectral images acquired under the same lighting conditions but different camera parameters and sensor temperatures. Normalization attenuates shadows, as seen in (b).}
\label{fig:normalize}
\end{figure}

\figref{fig:normalize}, on the other hand, presents examples of slices from three hypercubes taken under the same illumination settings but at different camera temperatures. Columns (a) and (b) used the same camera exposure settings, while column (c) had a different exposure. In the top row, images captured at \SI{1176.5}{\nano\meter} wavelength reveal significant differences in gray-level intensity between samples under different conditions. Shadows and variations in illumination also caused noticeable discrepancies. To address this, we applied pixel-wise normalization by subtracting the minimum spectral value across the 184 bands and then dividing by the maximum spectral value. The resulting normalized images are illustrated in the bottom row of \figref{fig:normalize}, demonstrating that after normalization, intra-class variability was reduced for fat, meat, and conveyor belt samples. Shadows and fine illumination differences were also attenuated, making the model more robust to variations in sensor temperature and exposure settings. Without normalization, there was a higher variability within each class, which would require more annotated samples to obtain better generalization. Without this step, distinguishing between samples of meat and fat became more challenging in the presence of shadows. Spectral normalization thus simplified the annotation task, reducing the need for extensive data labeling.

\subsection{Machine learning model}
\label{sec:models}
Several neural network architectures were developed, starting with a fully connected network designed to classify each 184-dimensional pixel in the hyperspectral image independently. However, this pixel-wise approach ignored the spatial context, losing the dependencies between neighboring pixels. As a result, it produced higher rates of false positives, as the classification did not consider how the hyperspectral signal of each pixel was influenced by the properties of the underlying materials in neighboring pixels. 

\begin{figure}[t]
    \centering
    \includegraphics[width=0.9\columnwidth]{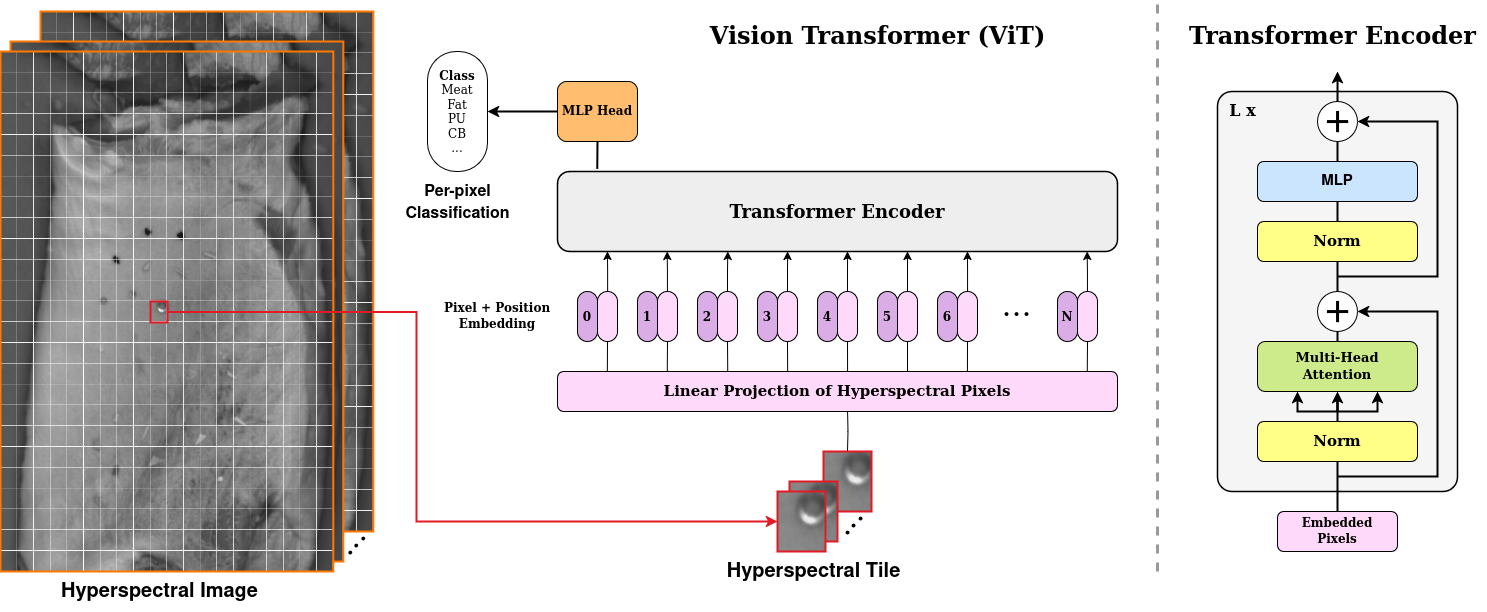}
    \caption{Overview of the ViT-based Architecture for HS Classification.}
    \label{fig:nn_arch}
\end{figure}

To address this limitation, we adopted a patch-based approach, dividing the hyperspectral image into overlapping patches of $20 \times 16$ pixels in size. Each patch is treated as an independent input to the network, allowing the model to leverage both spectral and spatial information. The architecture consists of a series of four cascaded transformer blocks, inspired by the Vision Transformer (ViT) introduced by \cite{Dosovitskiy2020}. Each transformer block is composed of a self-attention module followed by a two-layer MLP, with each component preceded by a normalization layer and containing residual connections. The self-attention module consists of eight attention heads that, unlike the original ViT \textendash which operates on non-overlapping image patches treated as tokens \textendash computes self-attention across all pixels within a patch, progressively refining the learned spatial-spectral representations of the pixels. The embedding length remains fixed at 184 throughout the network, corresponding to the dimensionality of the hyperspectral pixel. This design choice enables the model to capture the aforementioned dependencies and interactions between neighboring pixels, thereby improving classification accuracy. Finally, a fully connected linear layer is applied to the output of the last transformer block to classify each pixel, as depicted in \figref{fig:nn_arch}.

All our models were trained on an NVIDIA A100 Tensor Core GPU for 24 epochs with a batch size of 520. We utilized the AdamW optimizer with a weight decay of $2e^{-4}$ and learning rate of $1e^{-3}$, decayed using a polynomial learning rate scheduler with a warm-up of 3 epochs. A custom loss function was built on top of the standard cross-entropy loss to address the issue of false positives, particularly when meat or fat was misclassified as a contaminant. This loss function applied higher penalties for such misclassifications, reducing the false positive rate, while assigning lower penalties when the conveyor belt was misclassified as a contaminant as this was not as critical. Additionally, label smoothing was incorporated with a rate of 0.3 to soften the hard labels, which helped mitigate the impact of imprecise annotations, especially at the boundaries of contaminants. The models were implemented in PyTorch 1.11.0 and Python 3.8.10. The source code with all hyper-parameter configurations and pre-trained models will be made available at \url{https://github.com/hayatrajani/spectral_pork/}.

All trained models were then evaluated on a standard laptop equipped with two NVIDIA GeForce GTX4060 GPU and an Intel Core i7-9700TE CPU operating at 3.80 GHz running Ubuntu 20.04.5, Python 3.8.10 and PyTorch 1.9.0+cu111. Model performance is reported in terms of mean Intersection over Union (mIoU) and inference speed in the number of images processed per second (FPS).


\section{Results}
\label{sec:results}

This section presents the results of two of our best-performing models: one incorporating spectral normalization as a preprocessing step and one without it. For a model to be considered well-performing, it required reliable true positive detection (contaminants correctly identified as contaminants) and no false positives (misclassification of negative classes such as meat, fat, or conveyor belt as contaminants). Despite their overall accuracy, both models yielded some false positives, prompting an investigation into post-processing strategies to mitigate these cases. However, these steps also reduced the number of true positives, particularly near the boundaries of the contaminants, which are inherently more challenging to annotate and predict with confidence.

Given the industrial nature of this application, minimizing false positives was critical because they incur high costs during a full production day. False positives result in unnecessary manual inspections, increasing operational costs, while undetected contaminants pose serious consumer safety risks. Consequently, the design of the model and the loss function were central to achieving high contaminant detection accuracy while avoiding false positives.

One of the primary challenges in this application was the lack of reliable ground truth data. Contaminant annotations and predictions were reliable only at the center of foreign objects, while border regions were often ambiguous and difficult to label accurately.



To address false positives, a post-processing step was applied involving morphological erosion of the detected contaminants with a $3 \times 3$ square as the structuring element. Although this step eliminated small blobs of false positives, but it also removed low-confidence predictions near the borders of contaminants and excluded fine contaminants such as elongated or spiral-shaped PEHD, with a width of 1-2 pixels. Contaminants without at least one $3 \times 3$ blob were reclassified as fat for such cases. 
    
In addition to erosion, three pixel-wise post-processing conditions were applied to separate true from false positives based on their spectral characteristics, especially for PEHD and PA-PP:

\begin{enumerate}
    \item If the model predicts that the current sample is PEHD but $I(x,y,1225.5) - I(x,y,1211.5) > I(x,y,1026) - I(x,y,1012)$ and $I(x,y,1411) / I(x,y,1407.5)<1.04)$ then that sample is labeled as fat.
    \item If the model predicts that the current sample is PEHD but $I(x,y,1117) > I(x,y,1099.5)$ then that sample is labeled as a conveyor belt.
    \item If the model predicts that the current sample is PA-PP but $I(x,y,1225.5) - I(x,y,1215) < I(x,y,1232.5) - I(x,y,1012)$ then that sample is labeled as meat.
\end{enumerate}
where $I(x,y,w)$ denotes the gray-level intensity at the spectral wavelength $w$ and pixel coordinates $(x,y)$.

\begin{figure}
\centering
\begin{tabular}{cc}
 \hspace*{-6mm}
 \includegraphics[width=3.33in]{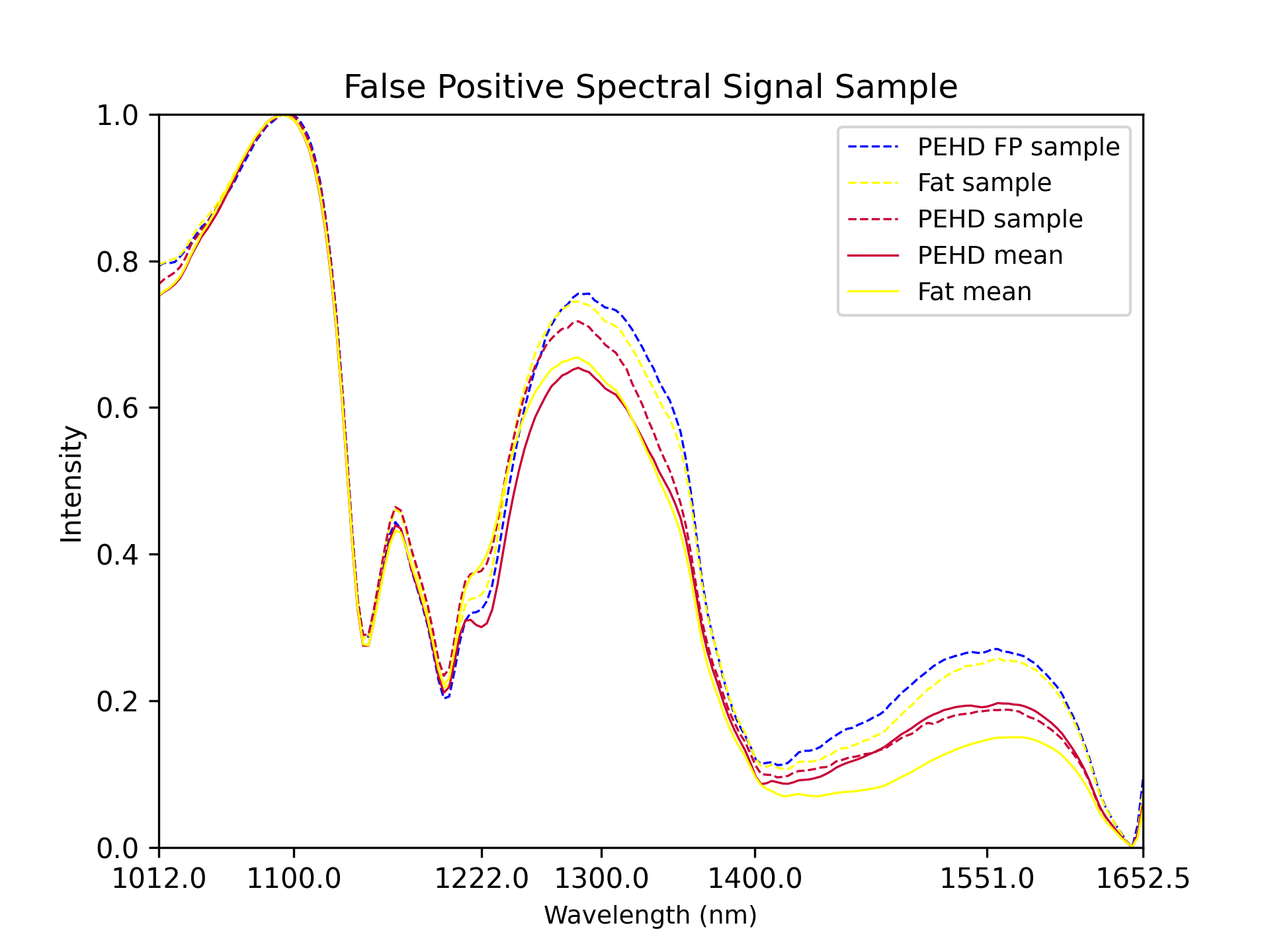} &
 \hspace*{-12mm}
 \includegraphics[width=3.33in]{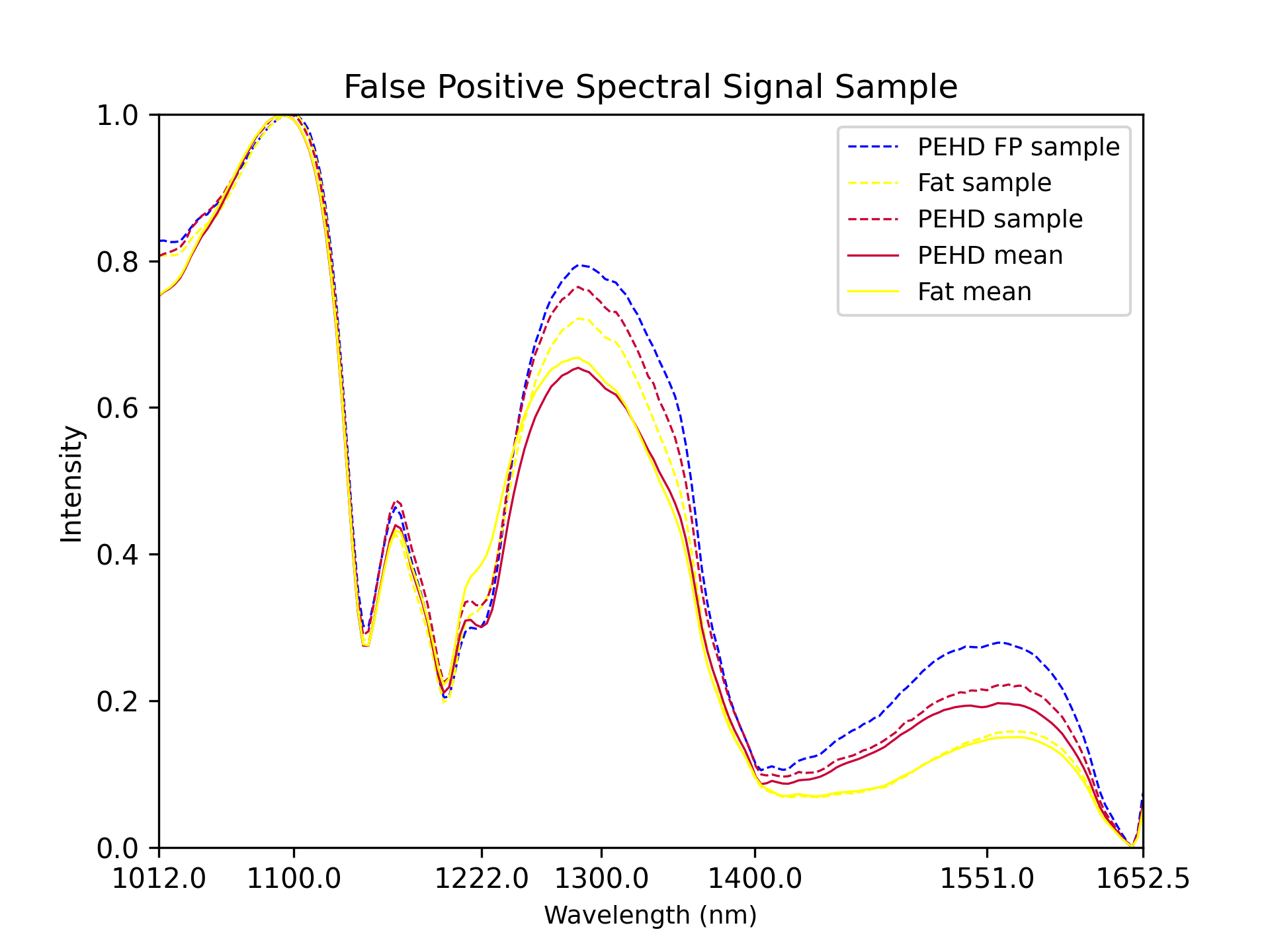} \\
  \hspace*{-6mm}
 \includegraphics[width=3.33in]{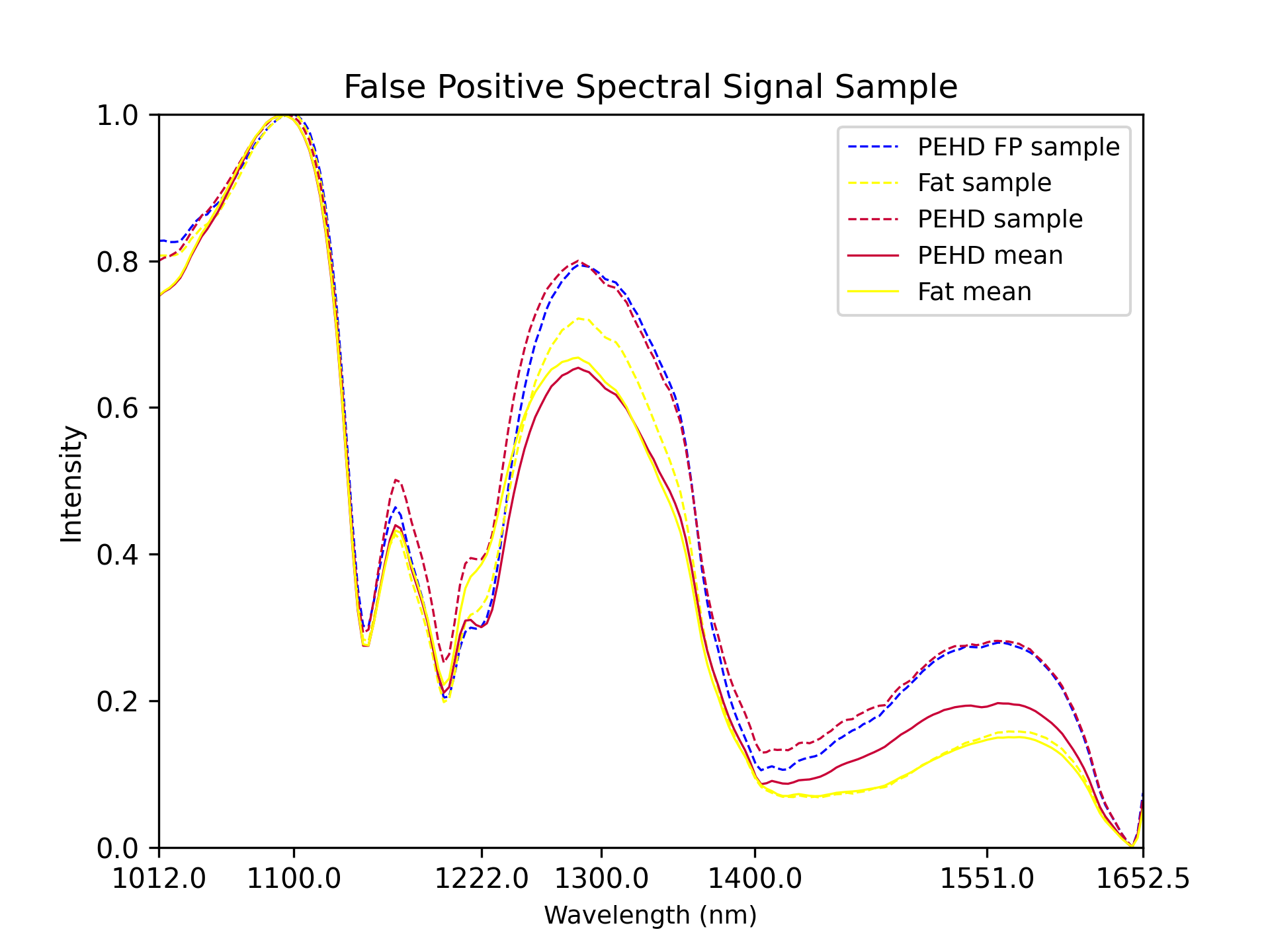} &
 \hspace*{-12mm}
 \includegraphics[width=3.33in]{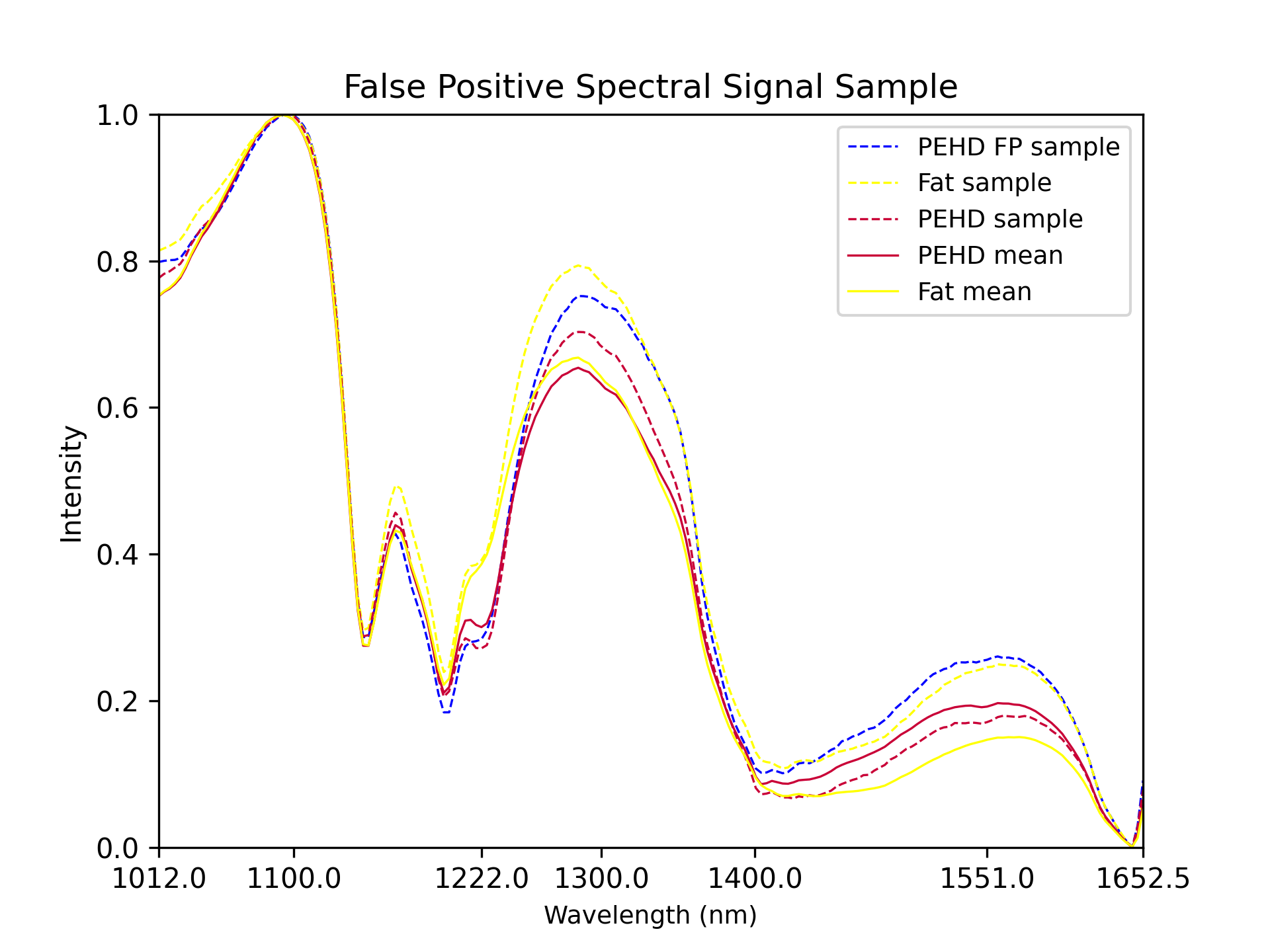} \\ 
\end{tabular}
\caption{Plots of fat samples labeled as PEHD by the model with spectral curves that look more similar to PEHD samples than fat.}
\label{fig:plot_fp_samples}
\end{figure}

\figref{fig:plot_fp_samples} illustrates cases where spectral signatures of PEHD false positives closely resembled those of true PEHD contaminants. Similarly, PEHD and fat sometimes exhibited extremely similar spectral curves, complicating their separation.

\begin{figure}
\centering
\fontsize{9}{9}\textbf{True Positives}
\\
\vspace*{1mm}
\begin{tabular}{cc}
  \includegraphics[scale=0.35]{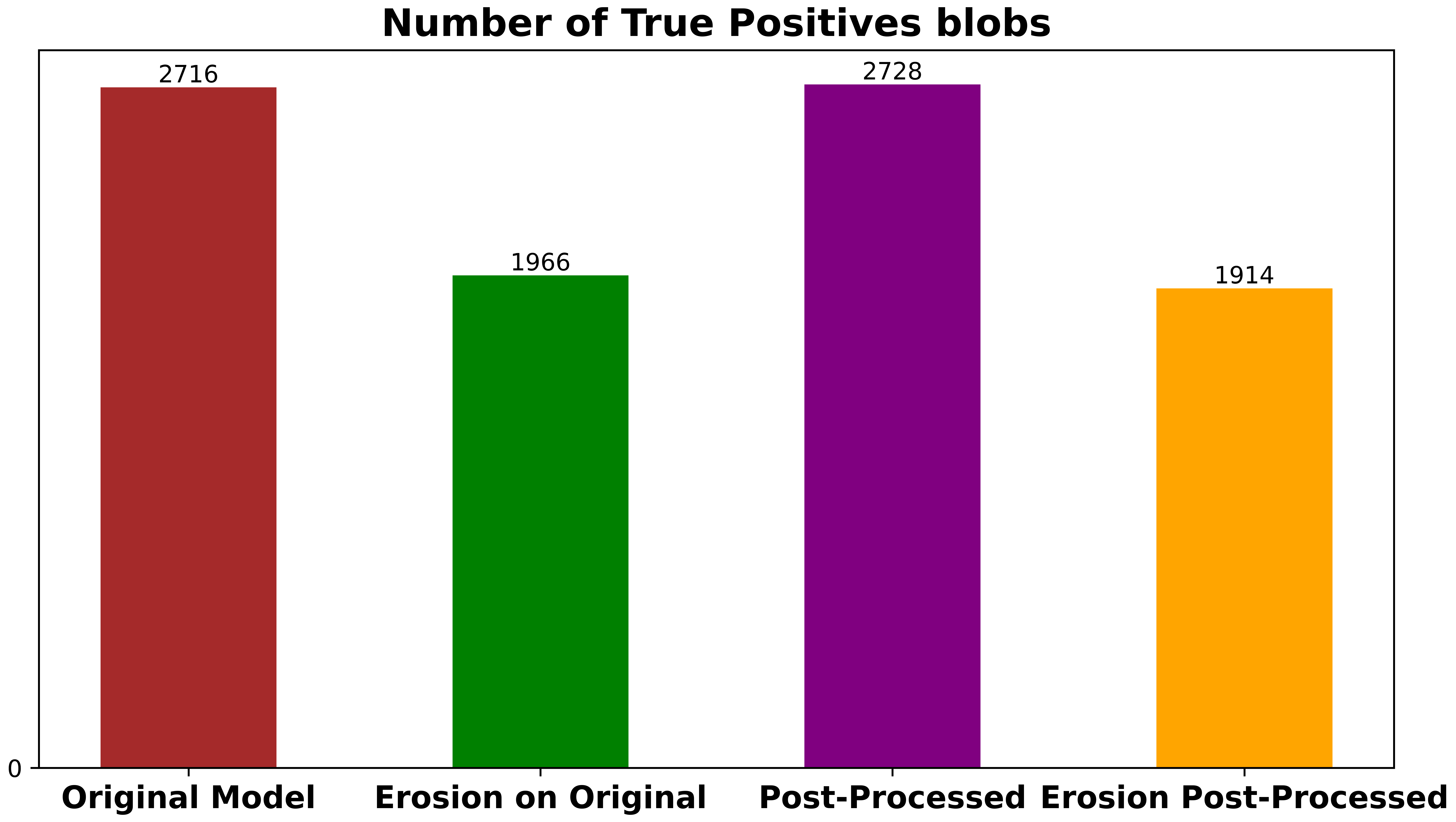} &
 \includegraphics[scale=0.35]{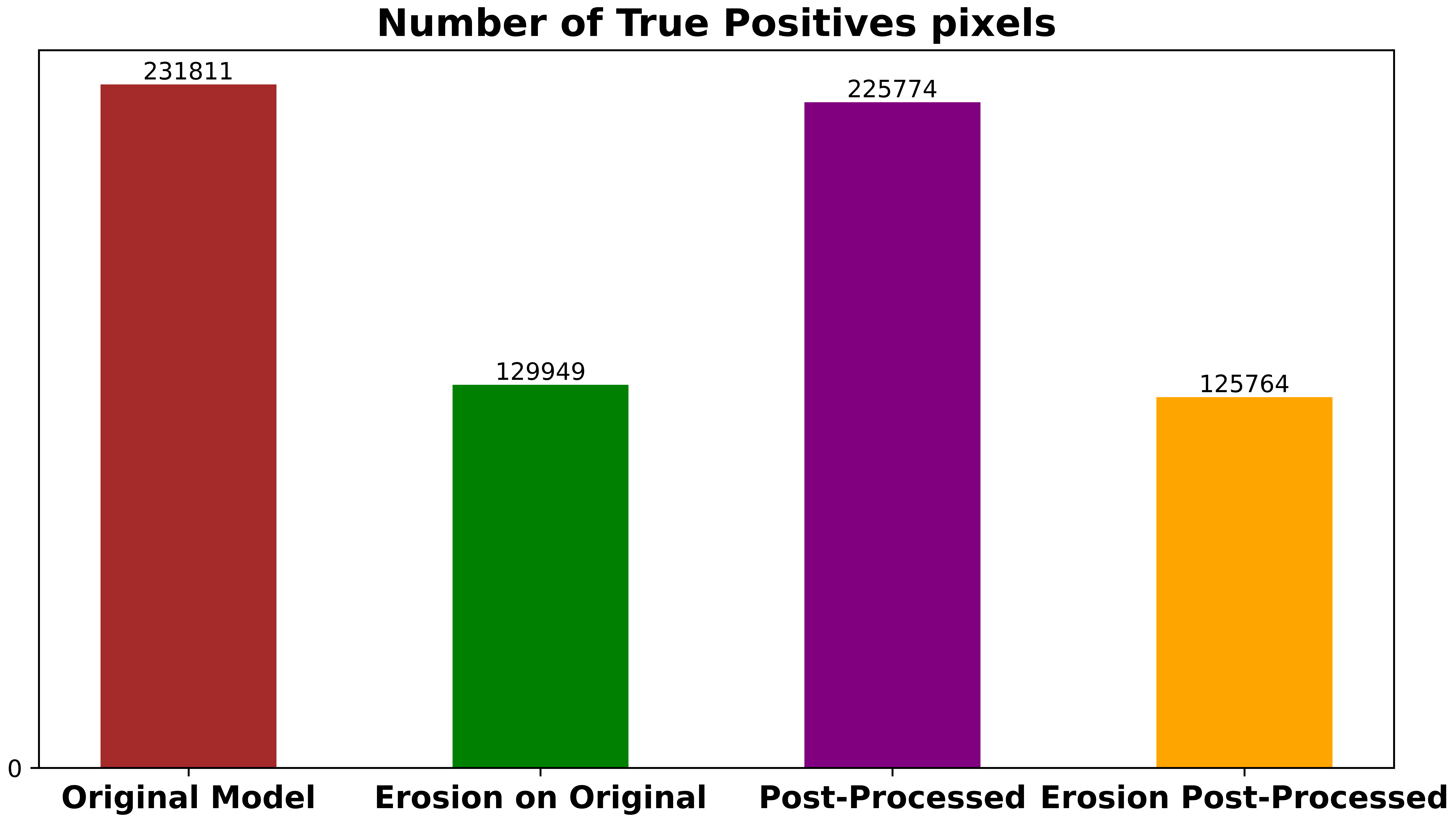} 
  \vspace*{-1mm}\\

 \fontsize{7}{7}\textbf{(a)} & \fontsize{7}{7}\textbf{(b)} 
 \end{tabular}\\
 
 \vspace*{6mm}
 \fontsize{9}{9}\textbf{False Positives}
  \\
  \vspace*{1mm}
  \begin{tabular}{cc}
  \includegraphics[scale=0.35]{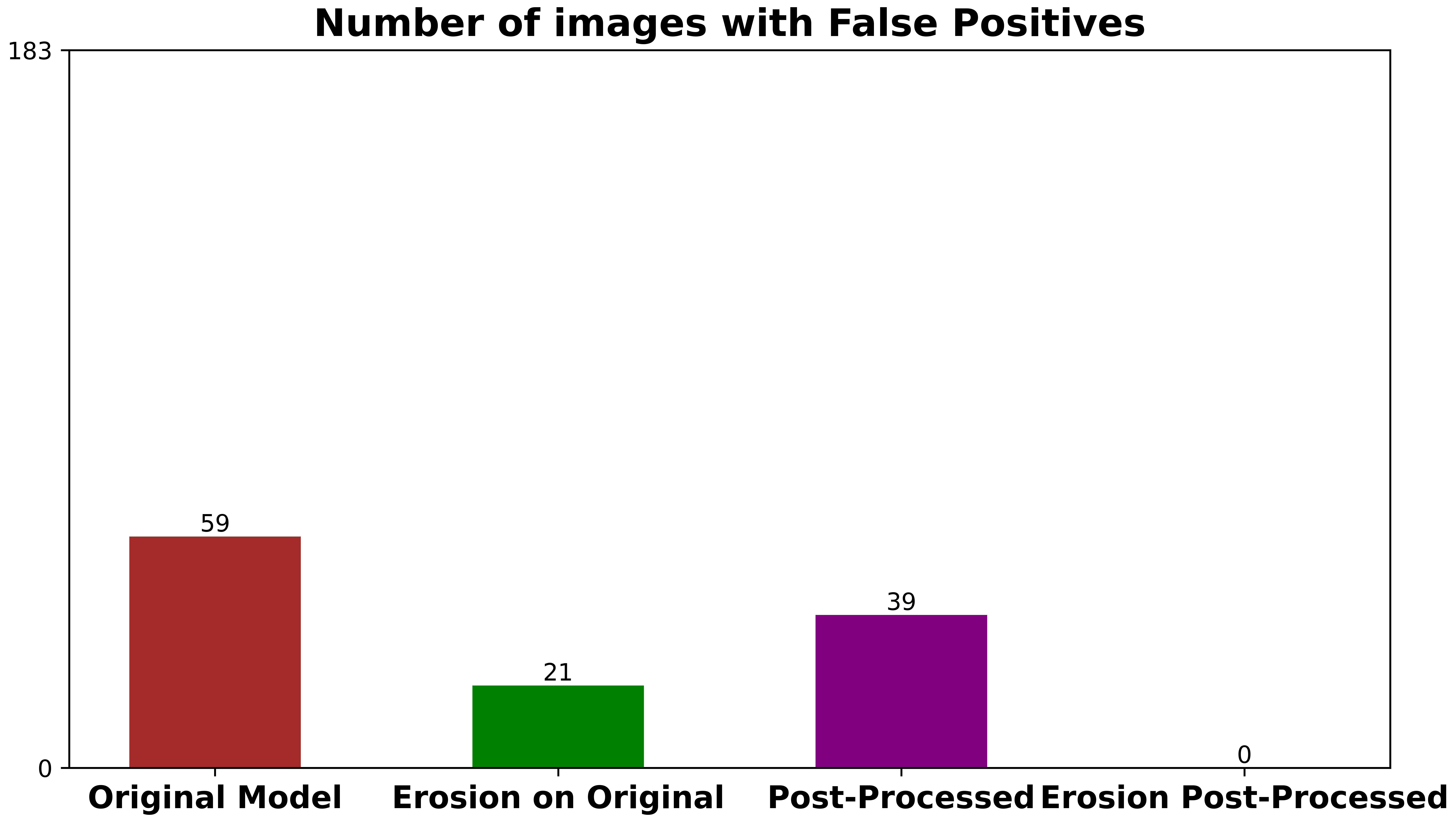} &
 \includegraphics[scale=0.35]{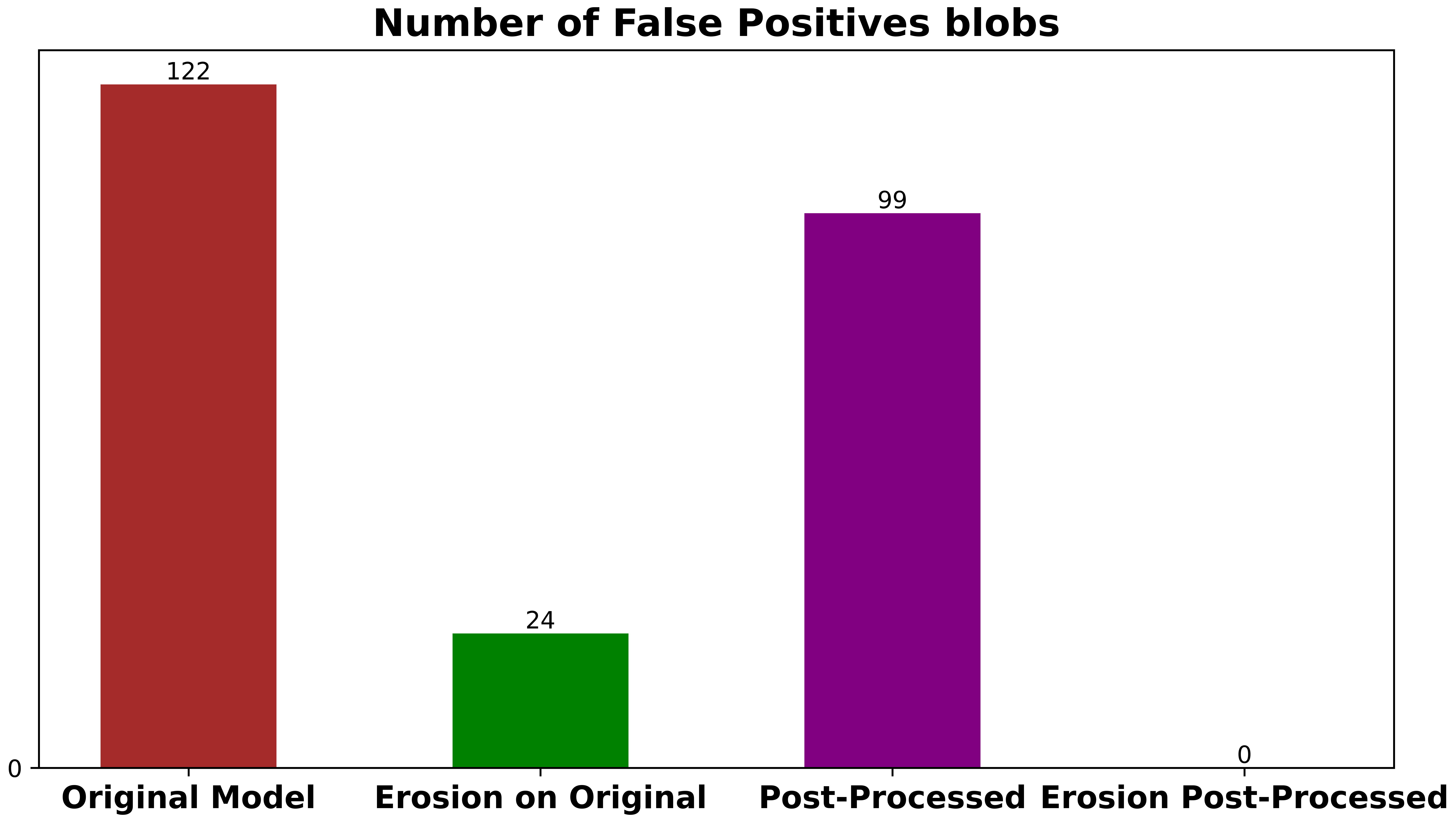}
  \vspace*{-1mm}\\

 \fontsize{7}{7}\textbf{(c)} & \fontsize{7}{7}\textbf{(d)} 
 \end{tabular}\\
\vspace*{1mm}

\begin{tabular}{cccc}
\vspace*{3mm}
\raisebox{-.5\height}{\includegraphics[scale=0.35]{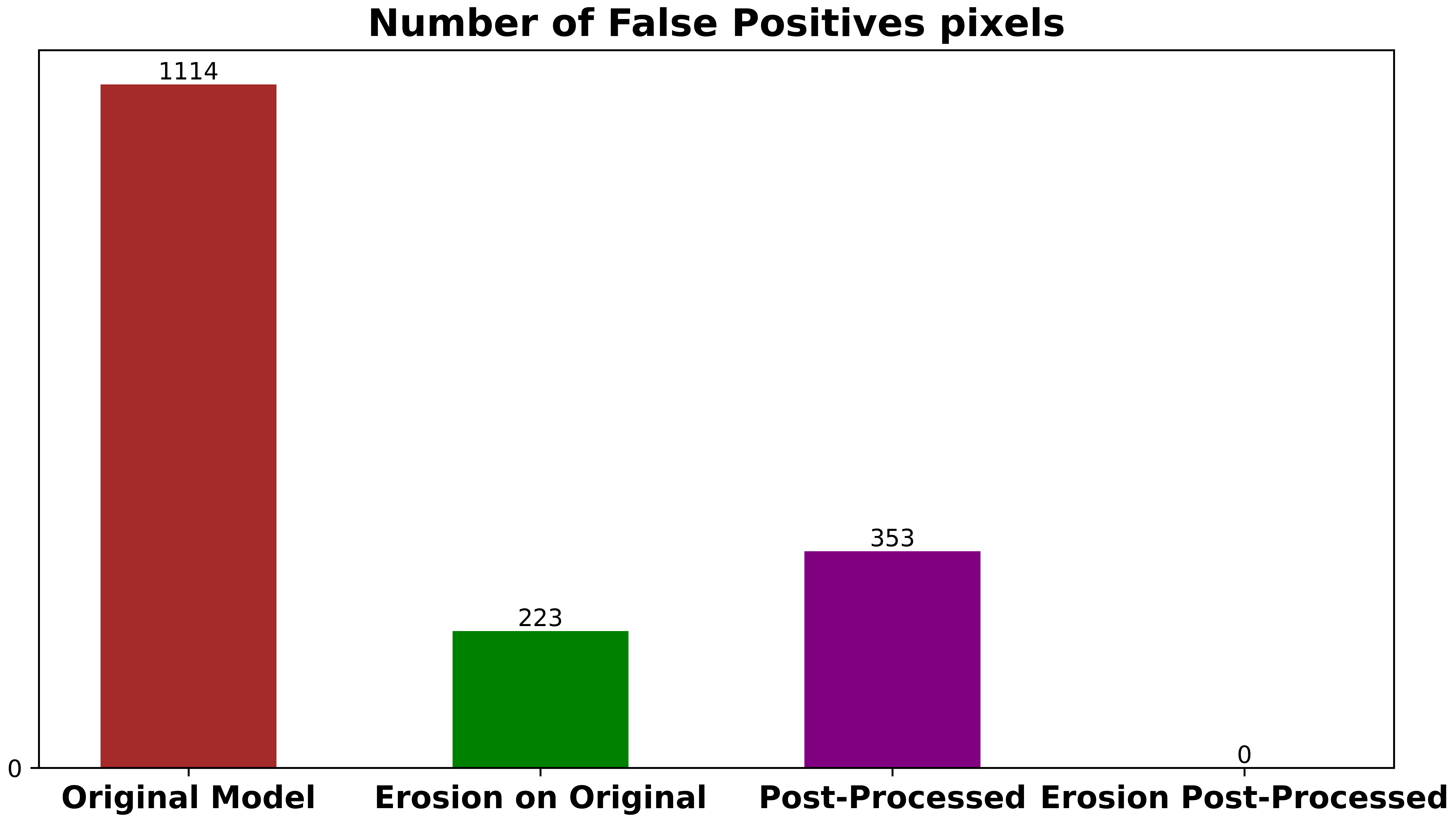}} &
\hspace*{-6mm}
\fontsize{7}{7}\textbf{Where:} &
\hspace*{-5mm}
\raisebox{-.5\height}{\includegraphics[scale=0.25]{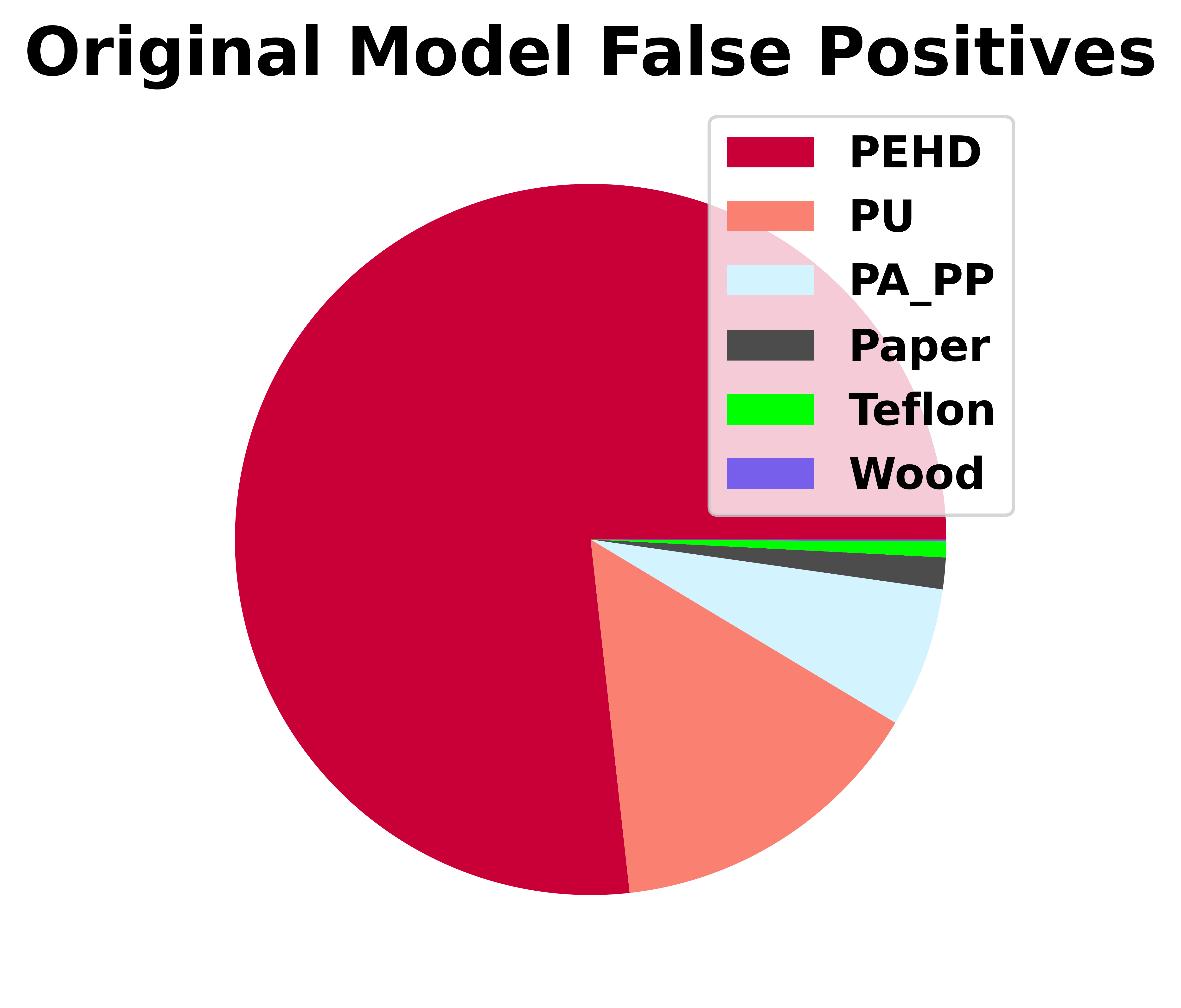}} &
\hspace*{-3mm}
\raisebox{-.5\height}{\includegraphics[scale=0.25]{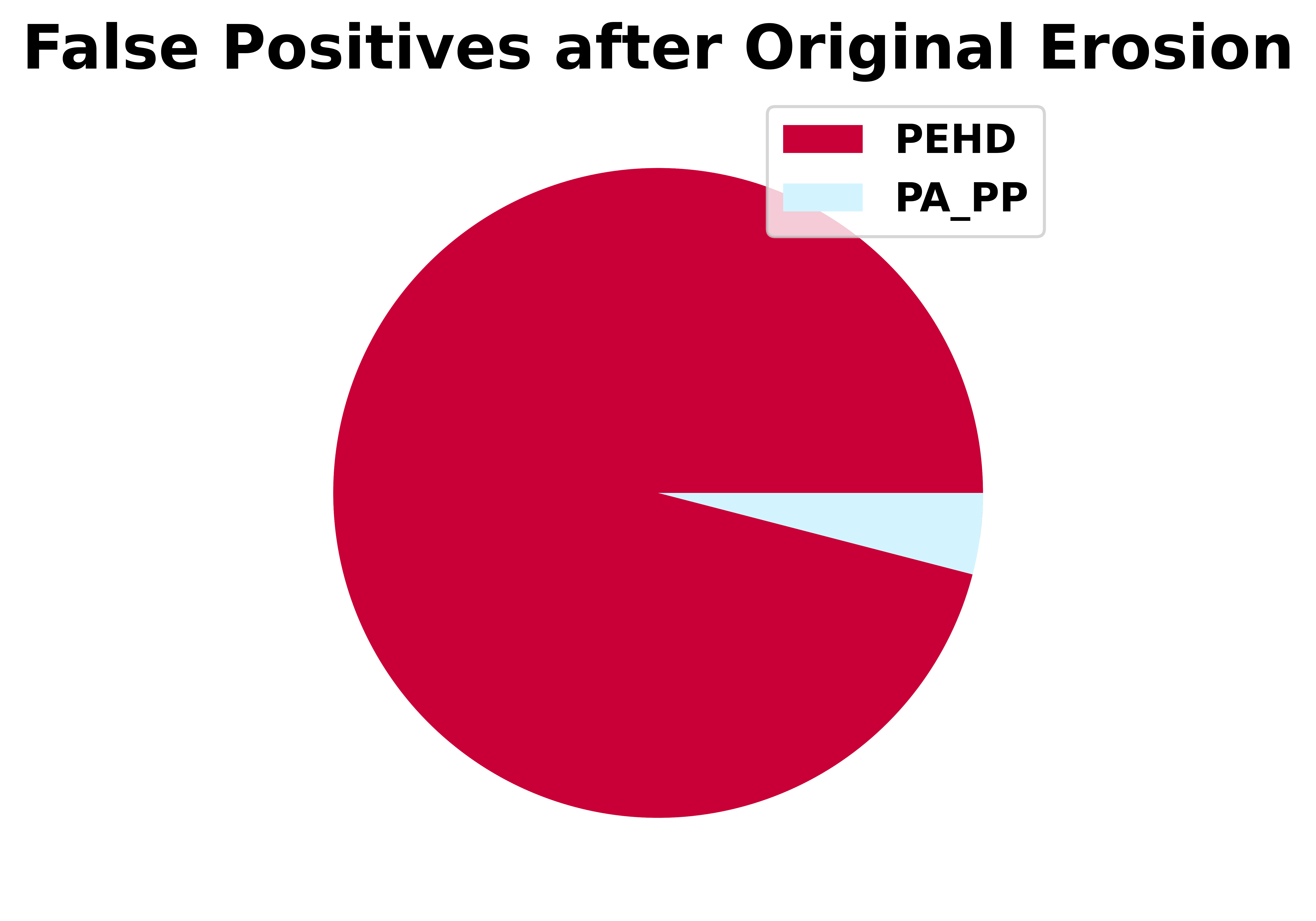}} 
\vspace*{-3mm}\\
 \fontsize{7}{7}\textbf{(e)} & & \fontsize{7}{7}\textbf{(f)} & \fontsize{7}{7}\textbf{(g)}
\end{tabular}\\

\caption{Prediction analysis of the results for the proposed model using the normalizing step, showing true positives as blob count in (a) and pixel count in (b), followed by false positives as image count in (c), blob count in (d) and pixel count in (e). The false positive pixels counted in the brown bar of (e) have the material distribution from (f) before erosion, while those counted in the green bar of (e) have the material distribution from (g).}
\label{fig:normalized_model_prediction_results}
\end{figure}

The results for the model with spectral normalization are shown in \figref{fig:normalized_model_prediction_results}. True positive detections from the test set of 55 images with contaminants are presented in plots (a) and (b). For comparison purposes, the false positives from 183 images without contaminants are presented in plots (c)-(g). The brown bars represent results from the original model, and the green bars correspond to results obtained after performing morphological erosion. Purple bars show the results after applying the three post-processing conditions to the results of the original model, and orange bars represent the results after applying the three post-processing conditions followed by morphological erosion.

After applying morphological erosion, contaminant predictions were reduced to the most confident regions, primarily the interiors of blobs, leading to a significant drop in the number of predicted contaminant pixels for small objects. This effect is evident when comparing brown bars with green ones and purple bars with orange ones in plots (a)–(b) in \figref{fig:normalized_model_prediction_results}. Importantly, the orange bars of 0 height in plots (c)–(e) show that the three post-processing steps followed by erosion completely eliminated false positives from the test set while retaining most true positives, as seen by the minimal difference between orange and green bars in (a) and (b).    

The false positives predicted by the original model are mostly PEHD, as seen in the pie plot from \figref{fig:normalized_model_prediction_results}(f) representing the distribution of the brown bar in (e). After the erosion step, there are still mostly PEHD false positives but also several PA-PP false positives, as seen in the (g) pie plot representing the distribution of the green bar in (e), motivating us to implement our three post-processing steps.

\begin{figure}
\centering
\fontsize{9}{9}\textbf{True Positives}
\\
\vspace*{1mm}
\begin{tabular}{cc}
  \includegraphics[scale=0.35]{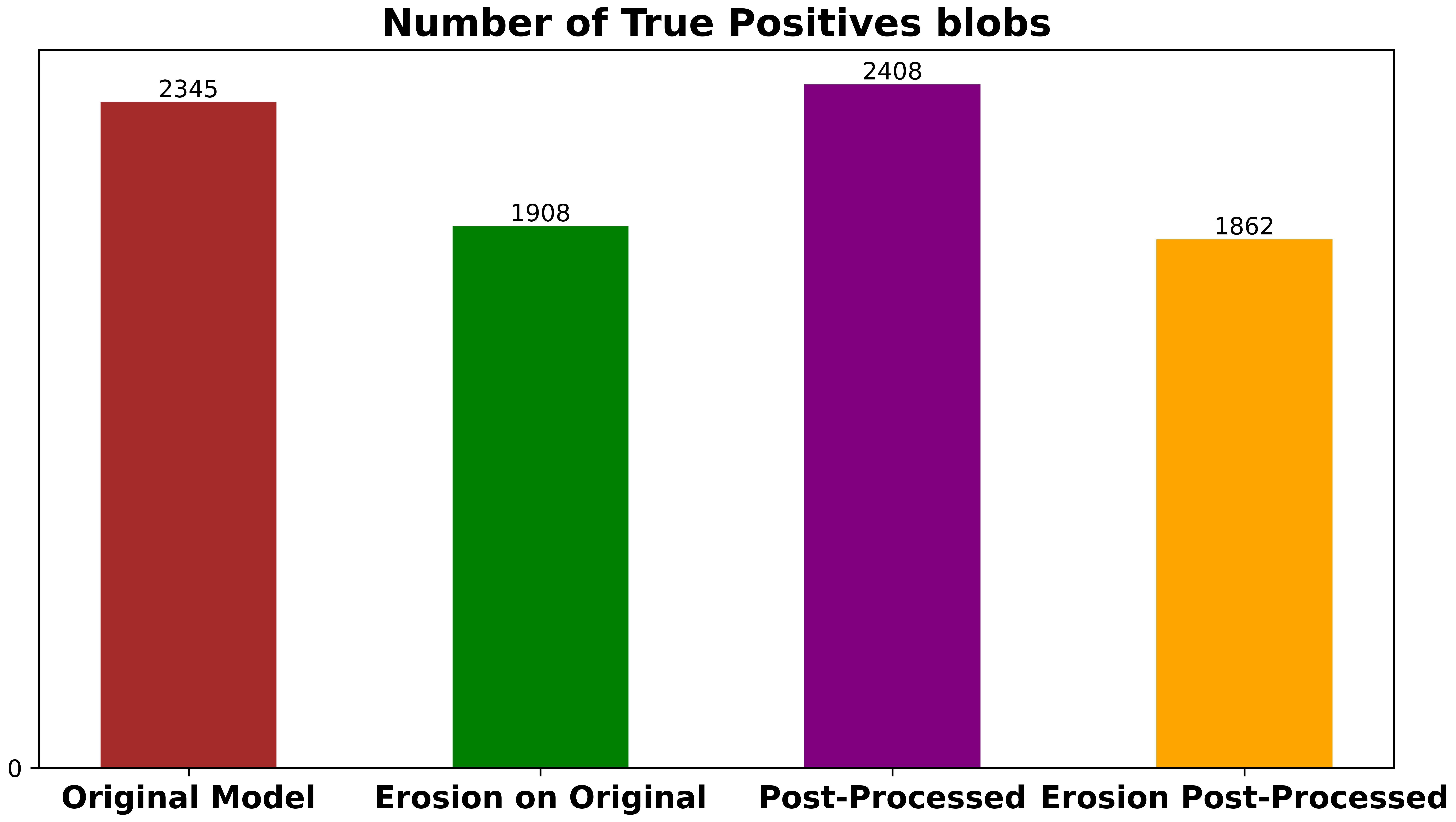} &
 \includegraphics[scale=0.35]{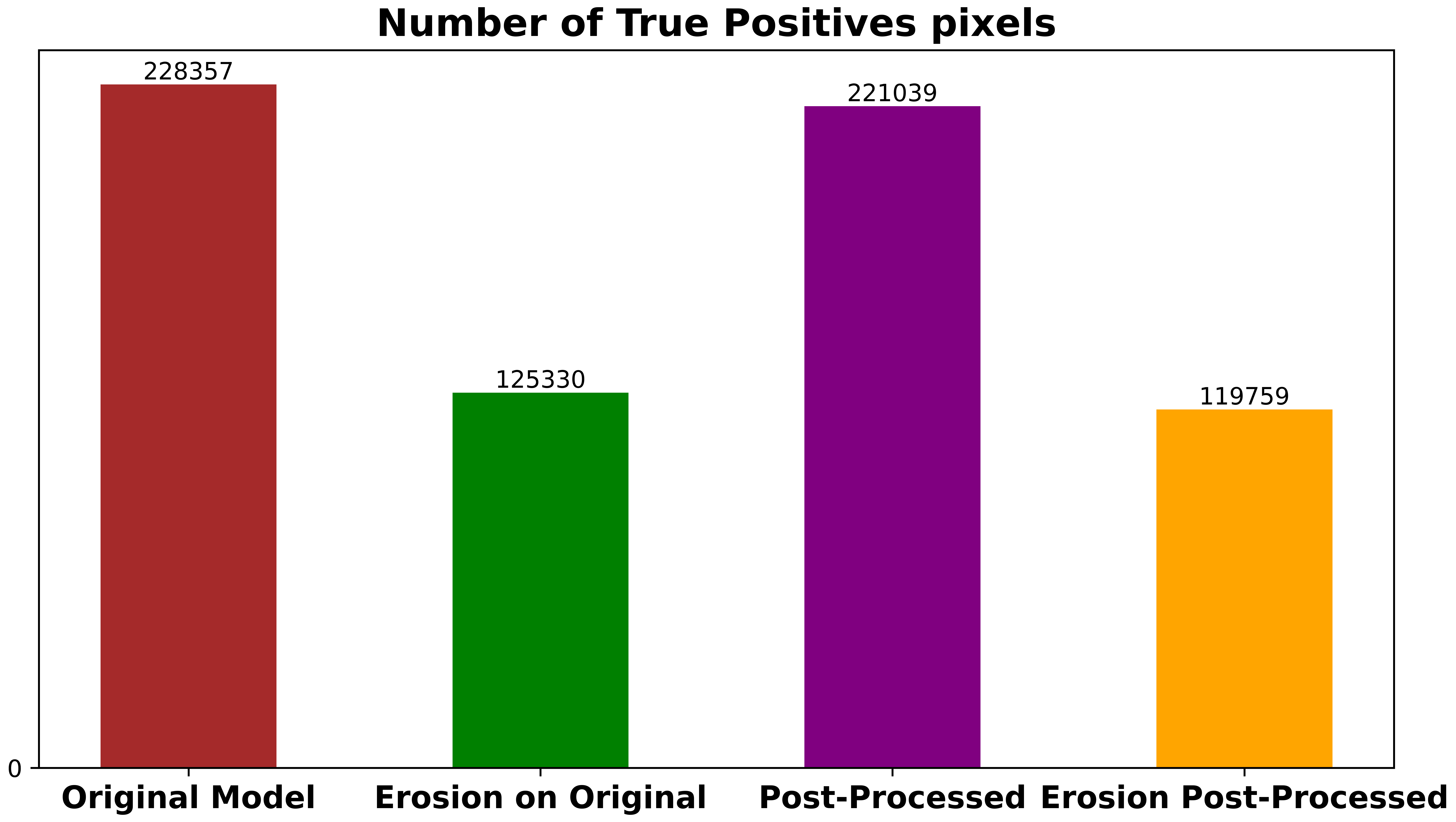} 
  \vspace*{-1mm}\\

 \fontsize{7}{7}\textbf{(a)} & \fontsize{7}{7}\textbf{(b)} 
 \end{tabular}\\
 
 \vspace*{6mm}
 \fontsize{9}{9}\textbf{False Positives}
  \\
  \vspace*{1mm}
  \begin{tabular}{cc}
  \includegraphics[scale=0.35]{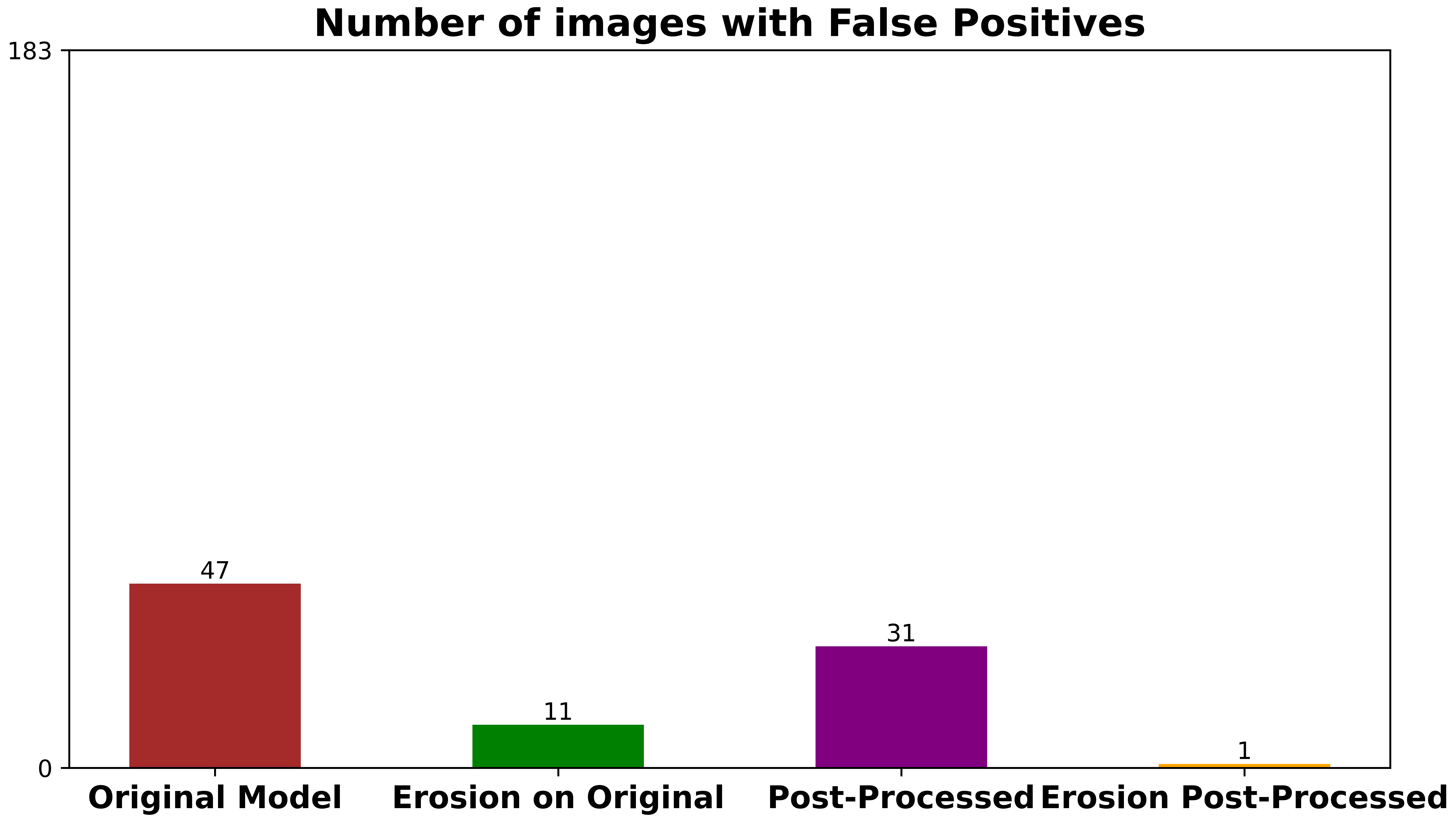} &
 \includegraphics[scale=0.35]{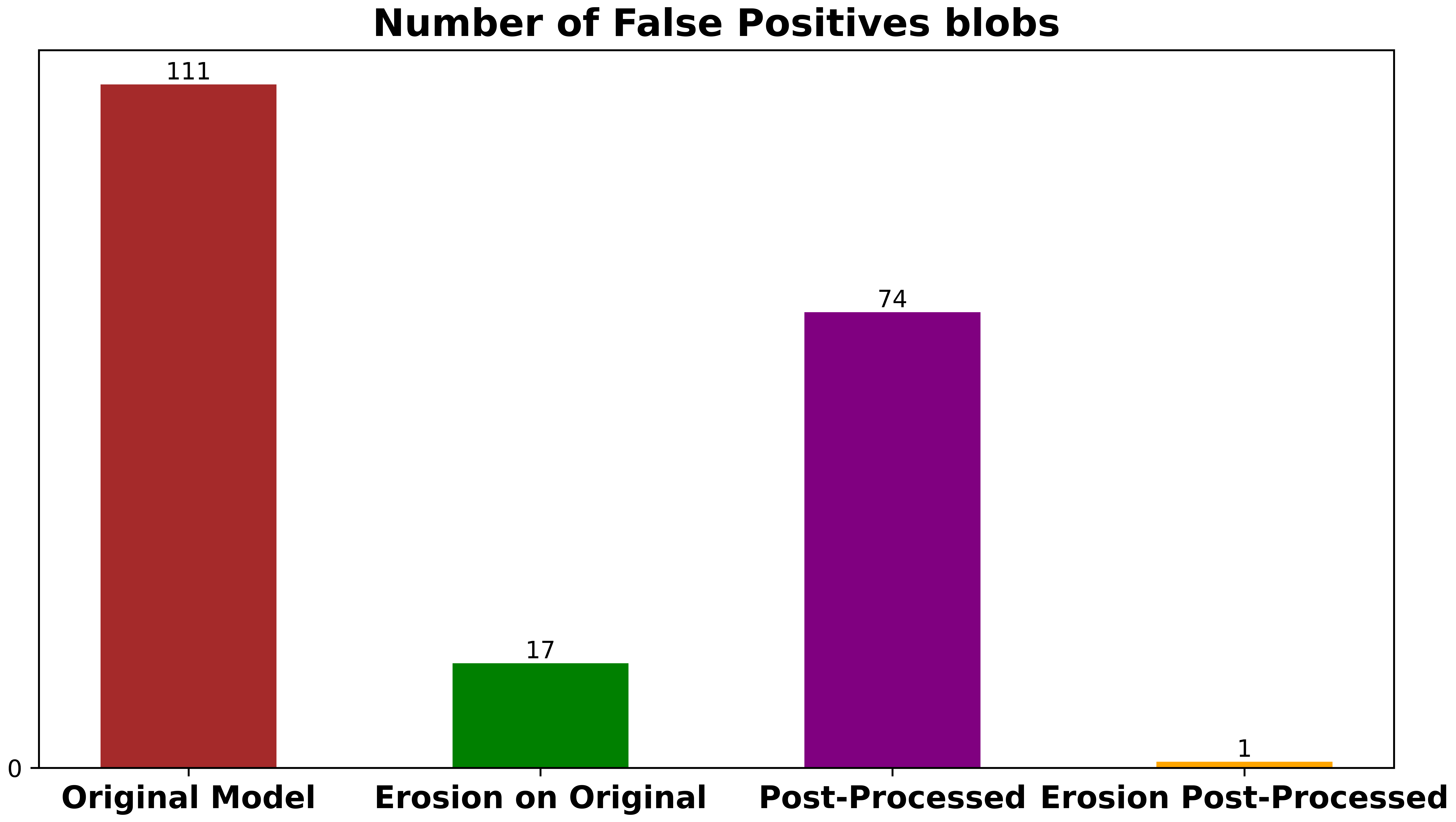}
  \vspace*{-1mm}\\

 \fontsize{7}{7}\textbf{(c)} & \fontsize{7}{7}\textbf{(d)} 
 \end{tabular}\\
\vspace*{1mm}

\begin{tabular}{cccc}
\vspace*{3mm}
\raisebox{-.5\height}{\includegraphics[scale=0.35]{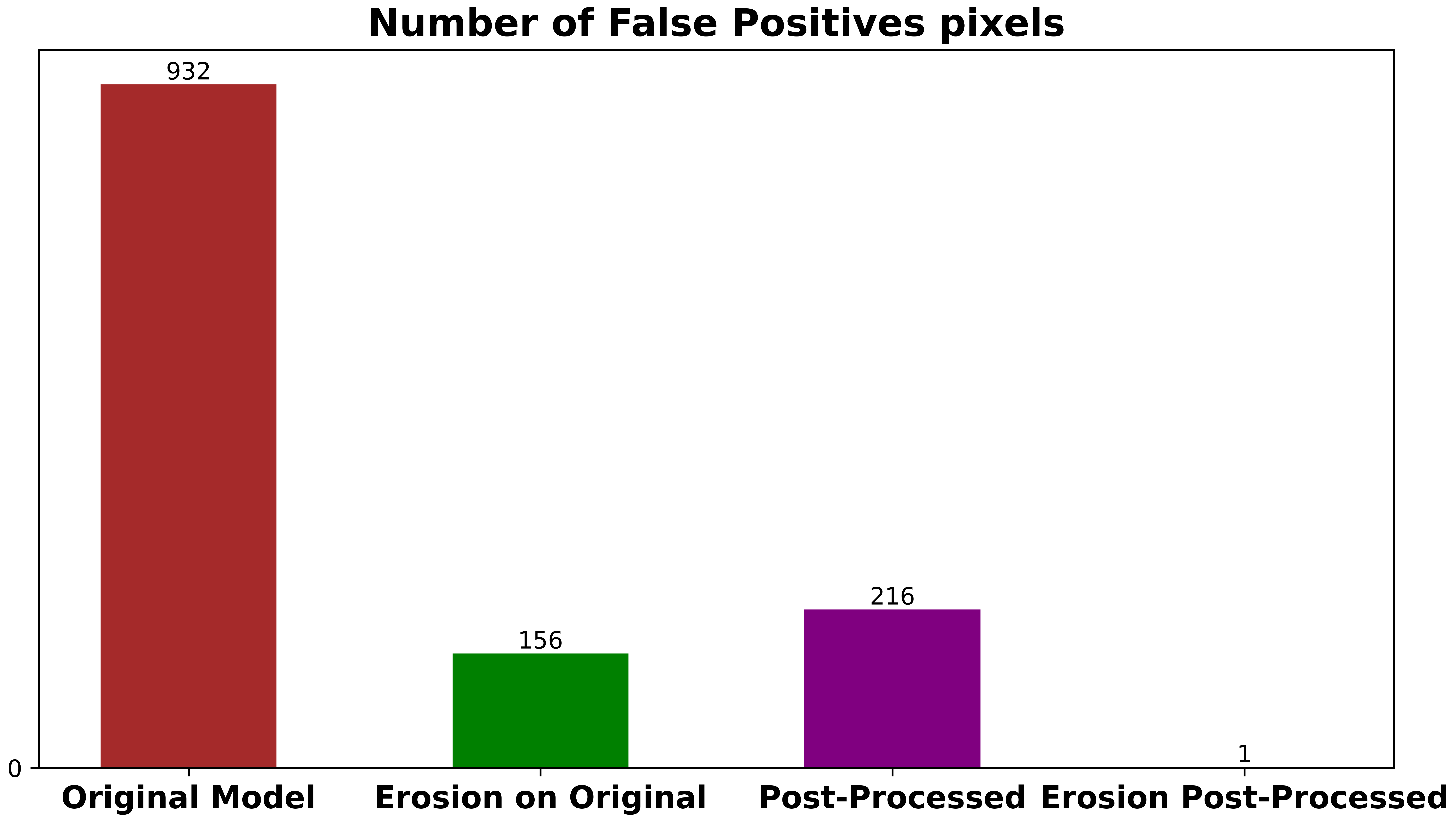}} &
\hspace*{-6mm}
\fontsize{7}{7}\textbf{Where:} &
\hspace*{-5mm}
\raisebox{-.5\height}{\includegraphics[scale=0.25]{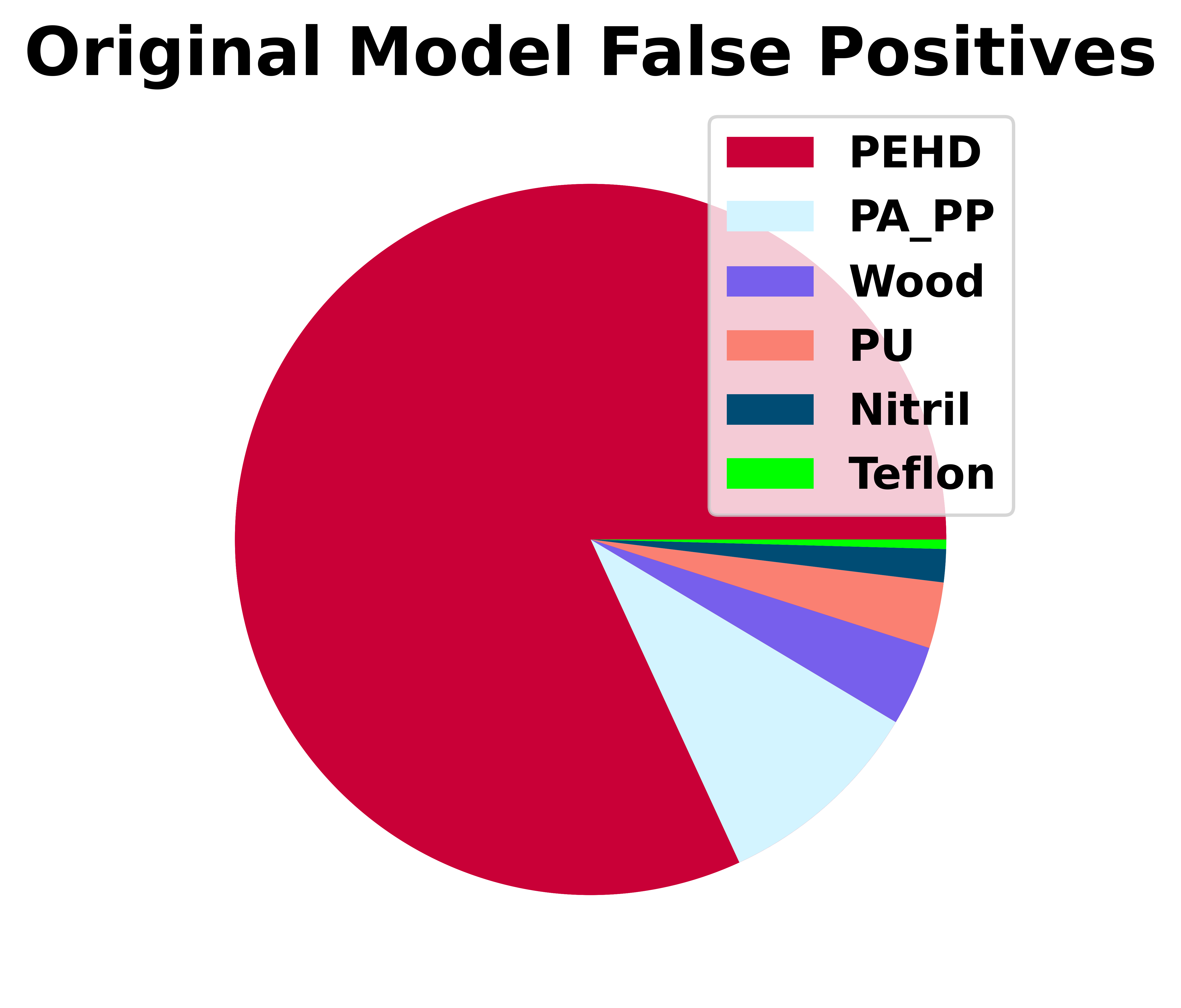}} &
\hspace*{-3mm}
\raisebox{-.5\height}{\includegraphics[scale=0.25]{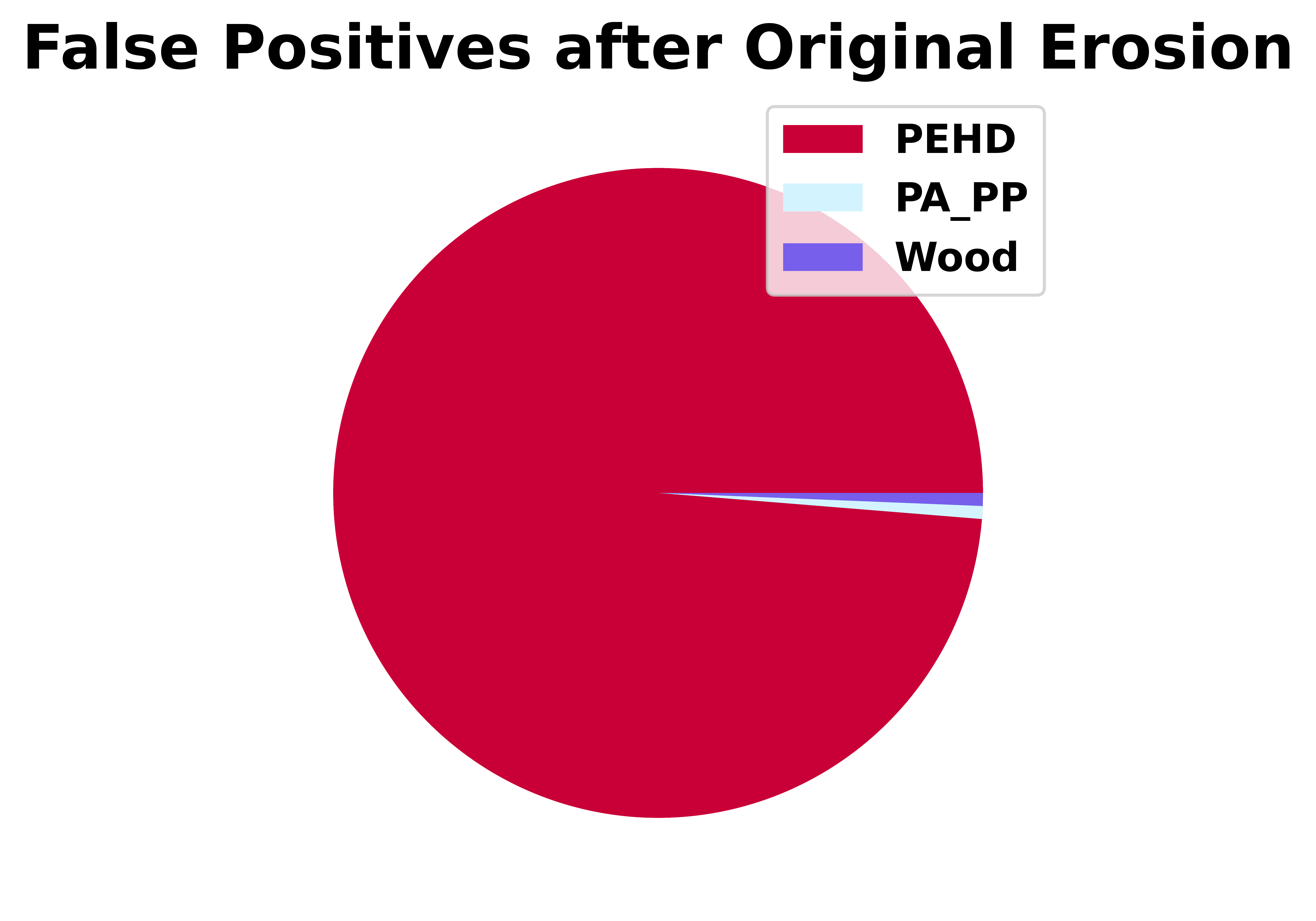}} 
\vspace*{-3mm}\\
 \fontsize{7}{7}\textbf{(e)} & & \fontsize{7}{7}\textbf{(f)} & \fontsize{7}{7}\textbf{(g)}
\end{tabular}\\

\caption{Prediction results for the model without normalizing, showing true positives as blob count in (a) and pixel count in (b), followed by false positives as image count in (c), blob count in (d) and pixel count in (e). The false positive pixels counted in the brown bar of (e) have the material distribution from (f) before erosion, while those counted in the green bar of (e) have the material distribution from (g).}
\label{fig:model_without_normalizing_prediction_results}
\end{figure}

We also evaluated a model without the spectral normalization step, as depicted in \figref{fig:model_without_normalizing_prediction_results}, using the same test set of 55 images with contaminants for evaluating true positive detection as in plots (a)-(b) and 183 images without contaminants to estimate how many false positives appear, as in plots (c)-(e). The results confirmed the model with spectral normalization to be the most robust, followed by the post-processing and erosion steps. It achieved no false positives on the 183-image test set while retaining most true positive predictions from the original model.

\begin{figure}
\centering
\begin{tabular}{ccccc}
 \includegraphics[width=1in]{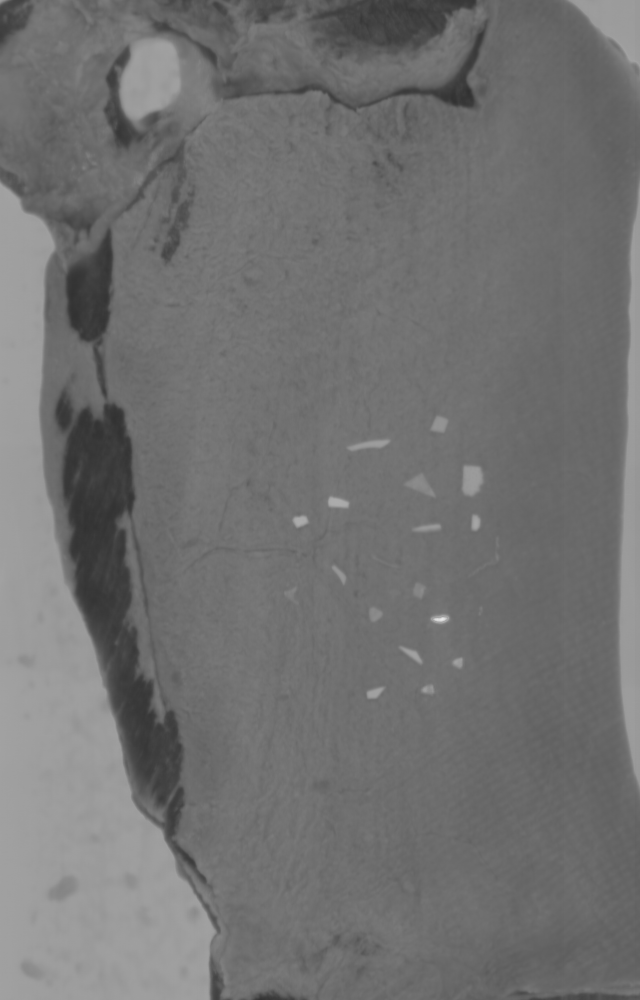} &
 \includegraphics[width=1in]{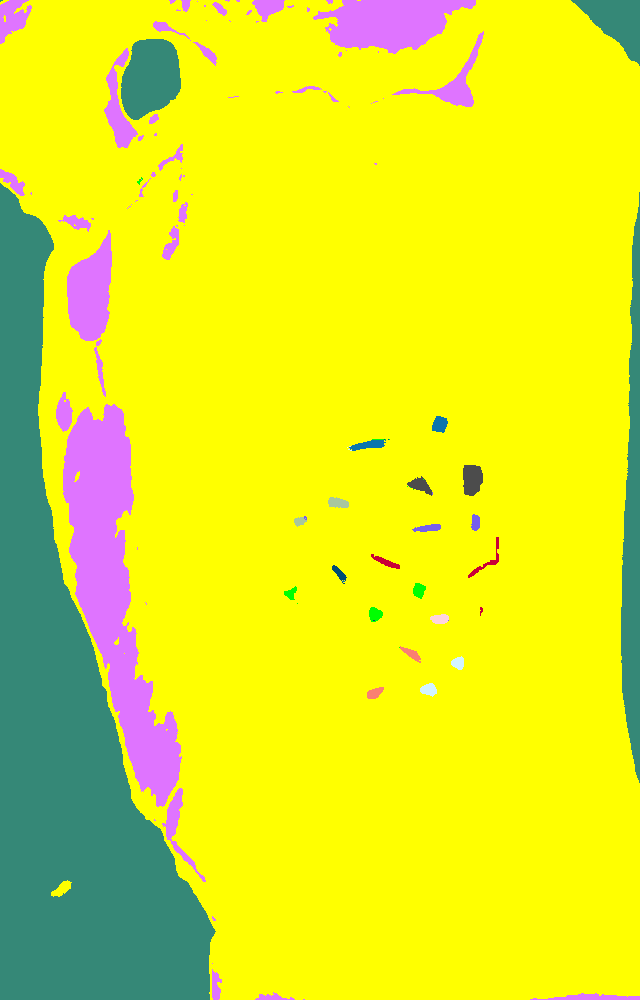} &
 \includegraphics[width=1in]{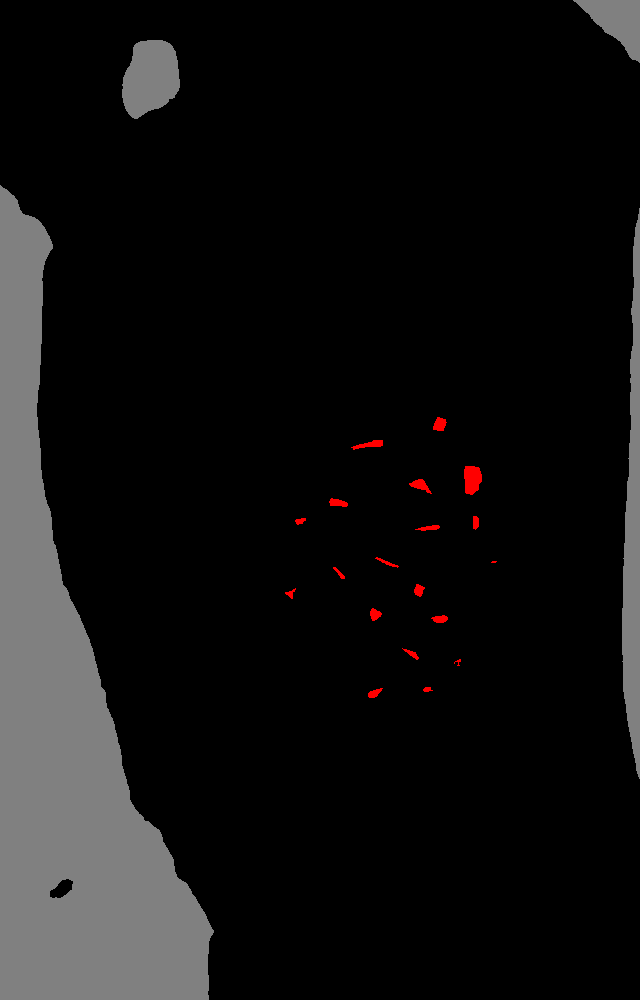} &
 \includegraphics[width=1in]{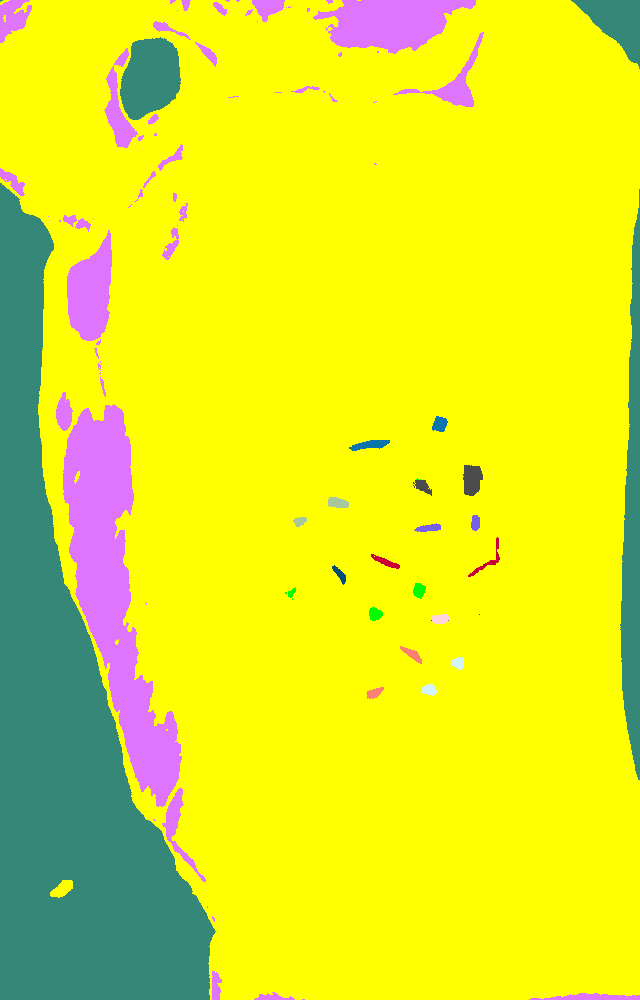} &
 \includegraphics[width=1in]{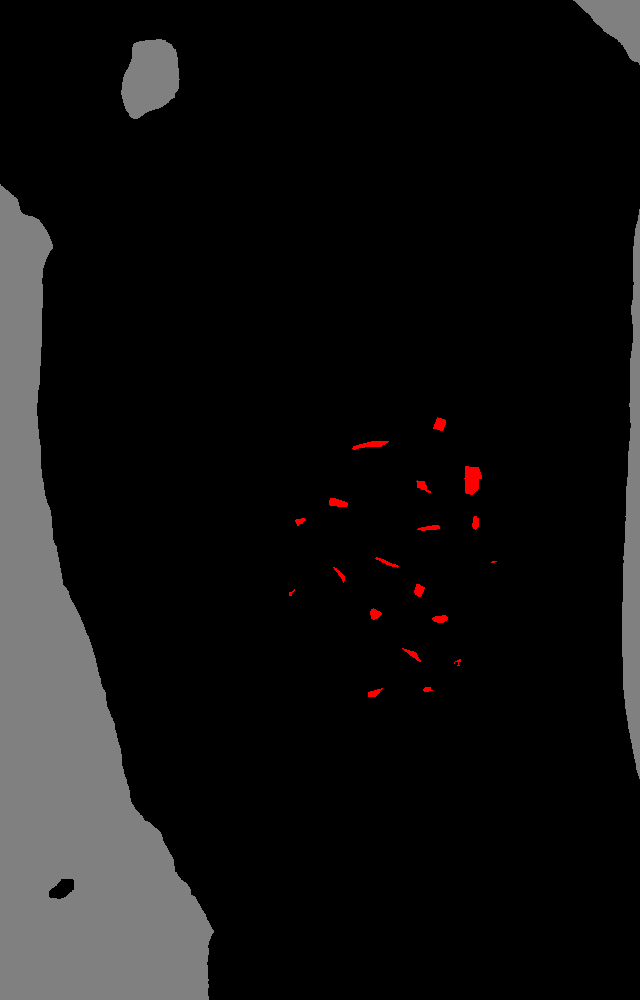}
 \\
 \includegraphics[width=1in]{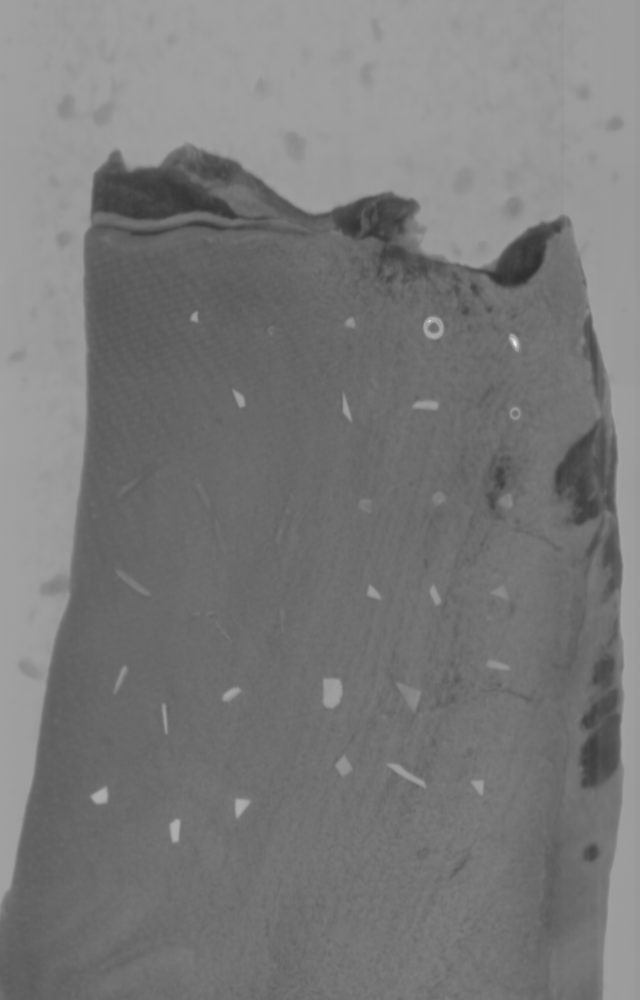} &
 \includegraphics[width=1in]{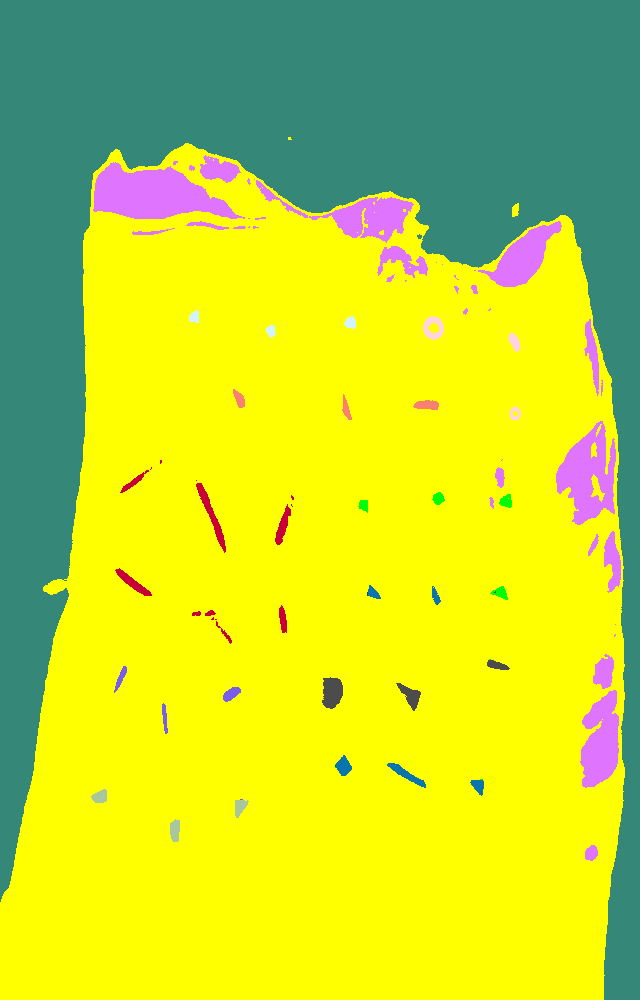} &
 \includegraphics[width=1in]{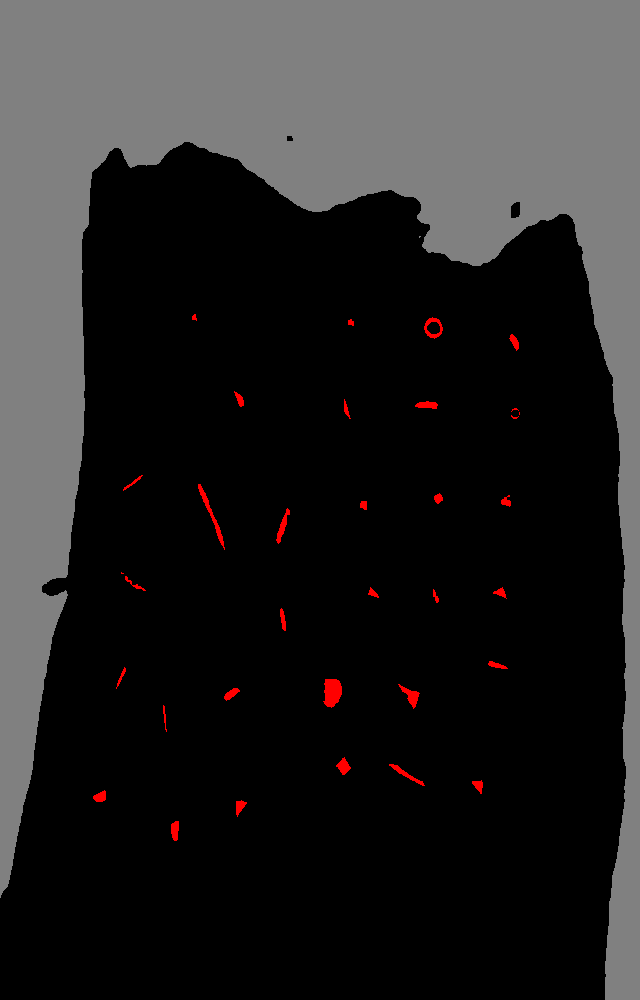} &
  \includegraphics[width=1in]{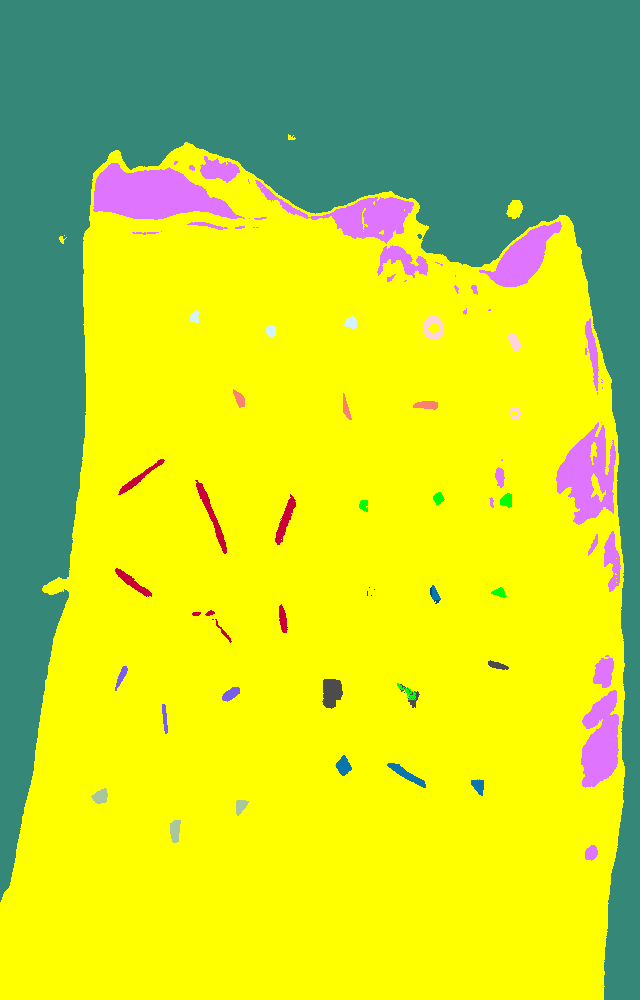} &
 \includegraphics[width=1in]{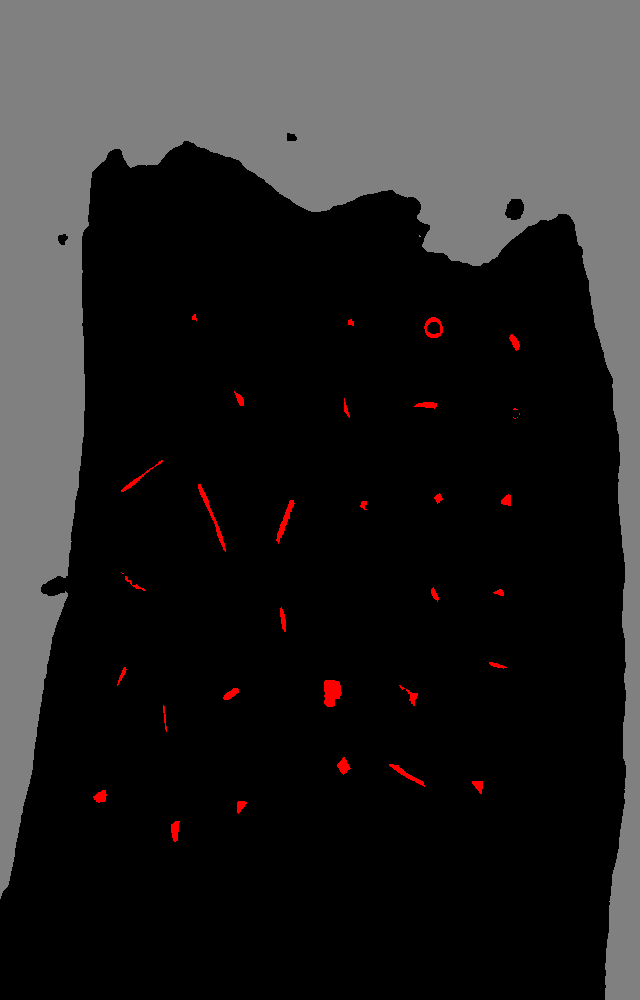}
 \\
 \includegraphics[width=1in]{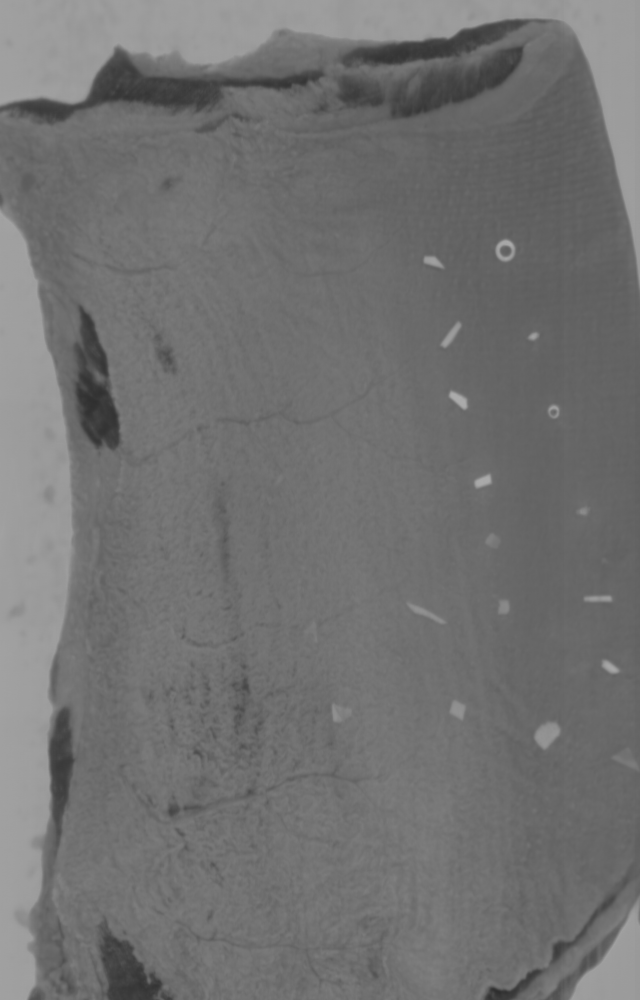} &
 \includegraphics[width=1in]{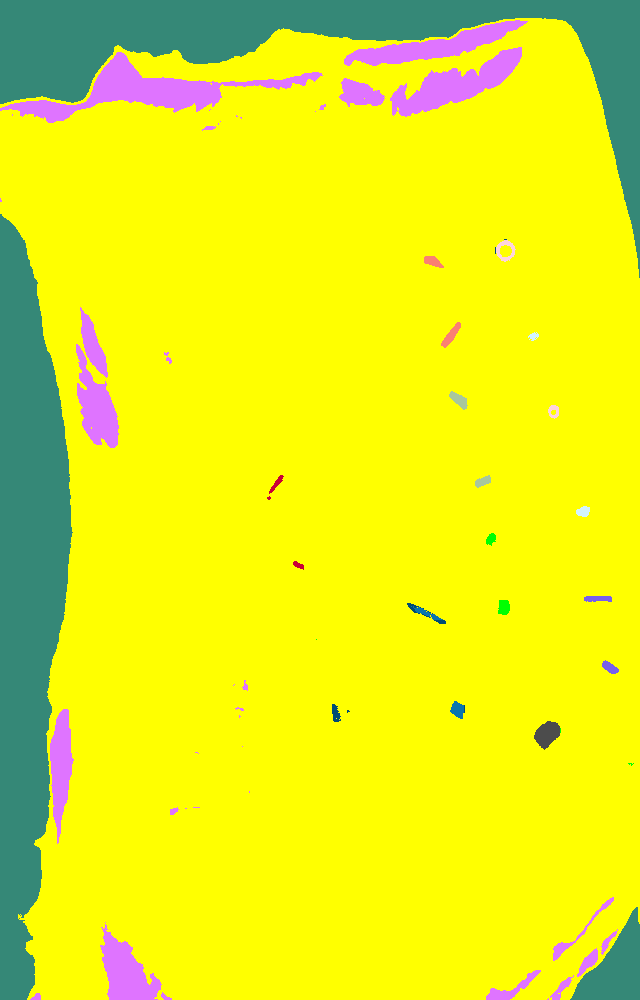} &
 \includegraphics[width=1in]{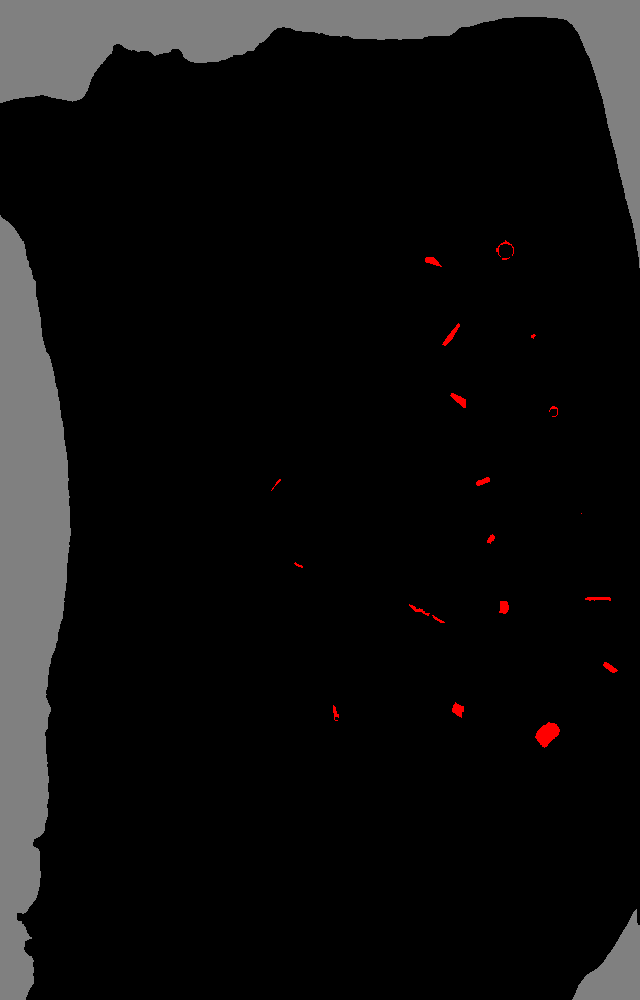} &
   \includegraphics[width=1in]{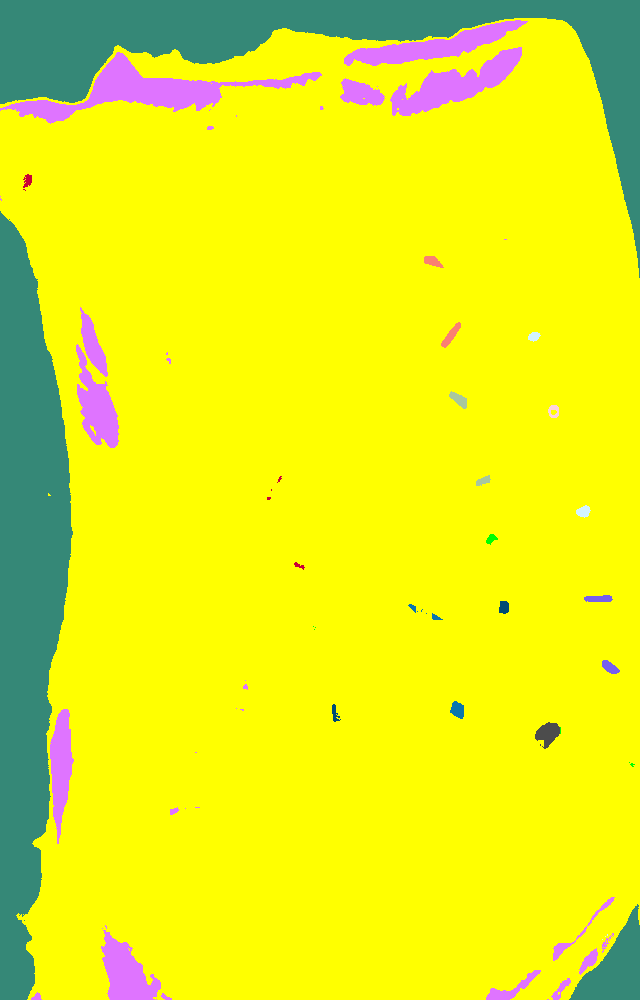} &
 \includegraphics[width=1in]{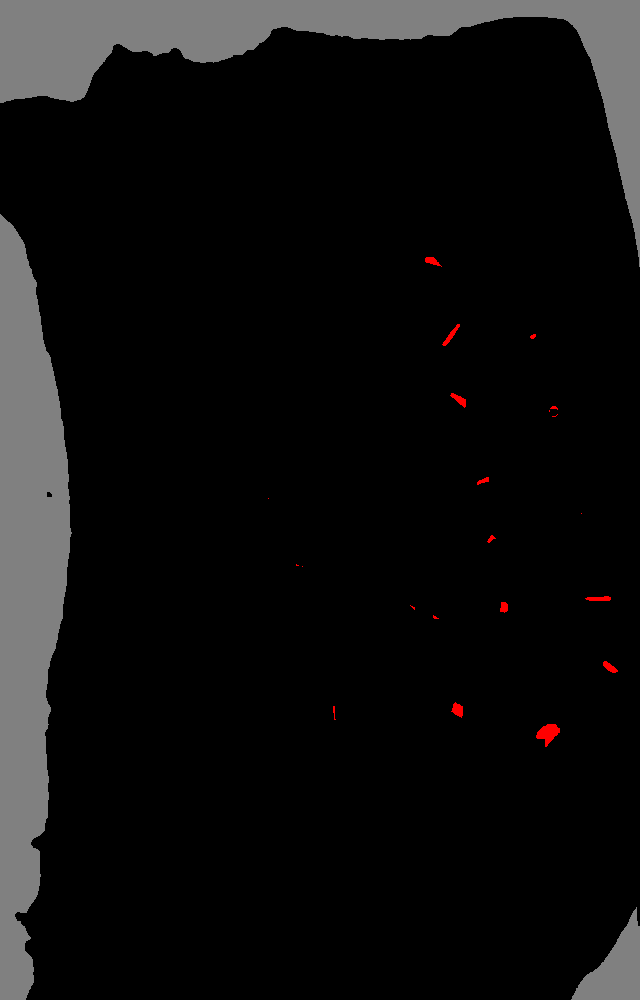}
 \\
  \textbf{(a)} &
 \textbf{(b)} &
 \textbf{(c)} &
 \textbf{(d)} &
 \textbf{(e)} 
 \\
\end{tabular}
\caption{Prediction examples on fat. (a) Spectral wavelength \SI{1169.5}{\nano\meter}; (b) Prediction of the normalizing model; (c) Prediction of the normalizing model followed by Post-Processing and Contaminant Erosion; (d) Prediction of the model without the normalizing step; (e) Prediction of the model without normalizing step followed by Post-Processing and Contaminant Erosion.}
\label{fig:normalizing_model_results_fat}
\end{figure}

\begin{figure}
\centering
\begin{tabular}{ccccc}
 \includegraphics[width=1in]{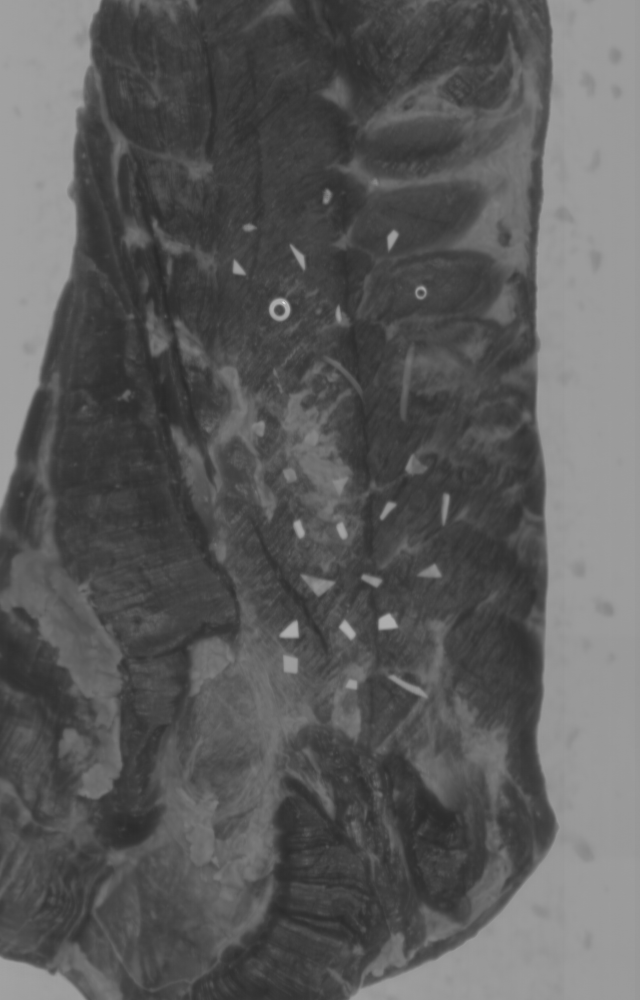} &
 \includegraphics[width=1in]{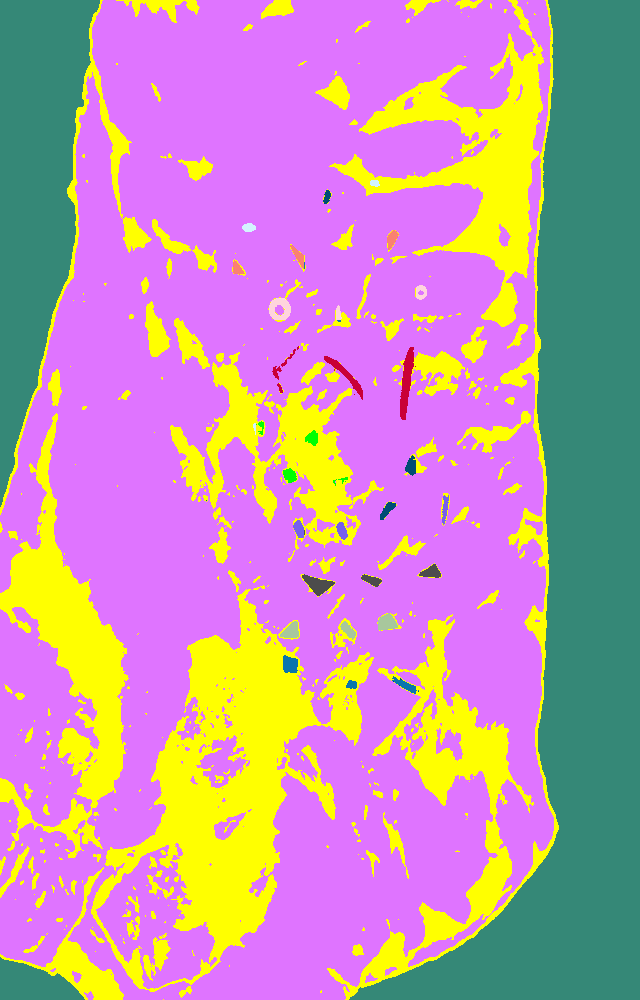} &
 \includegraphics[width=1in]{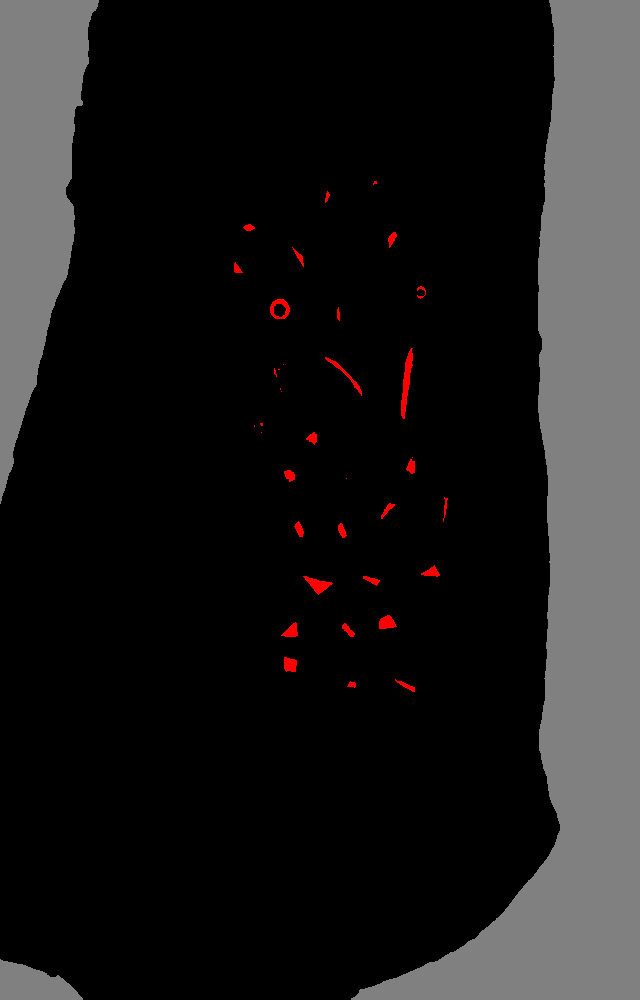} &
 \includegraphics[width=1in]{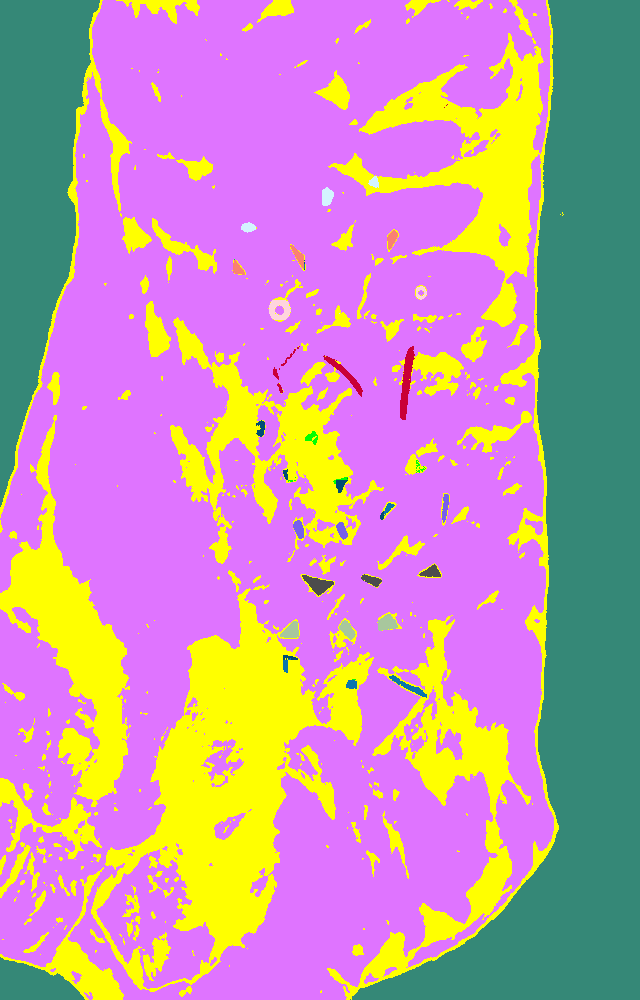} &
 \includegraphics[width=1in]{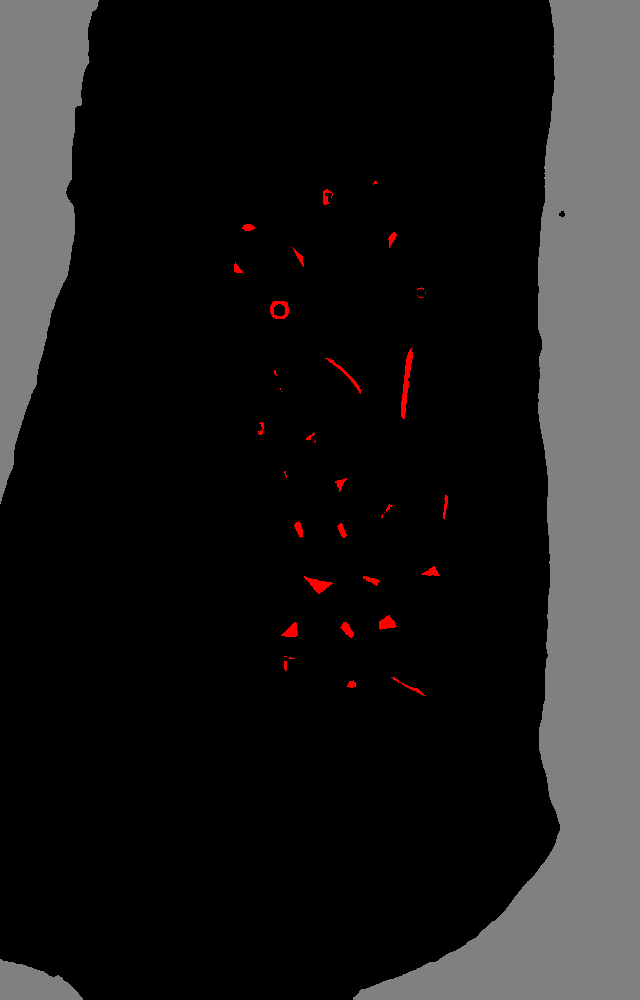}
 \\
 \includegraphics[width=1in]{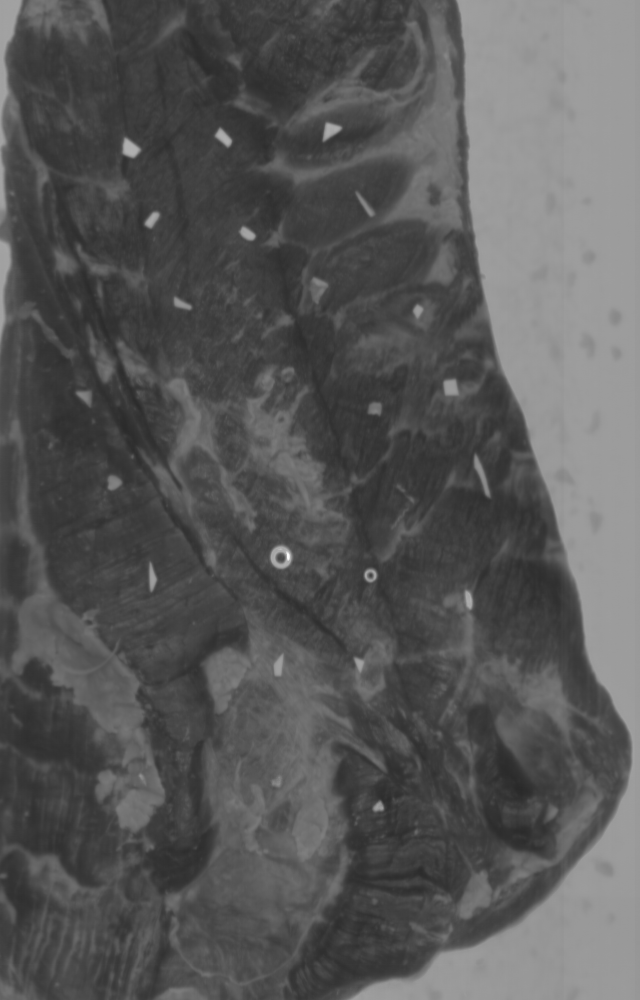} &
 \includegraphics[width=1in]{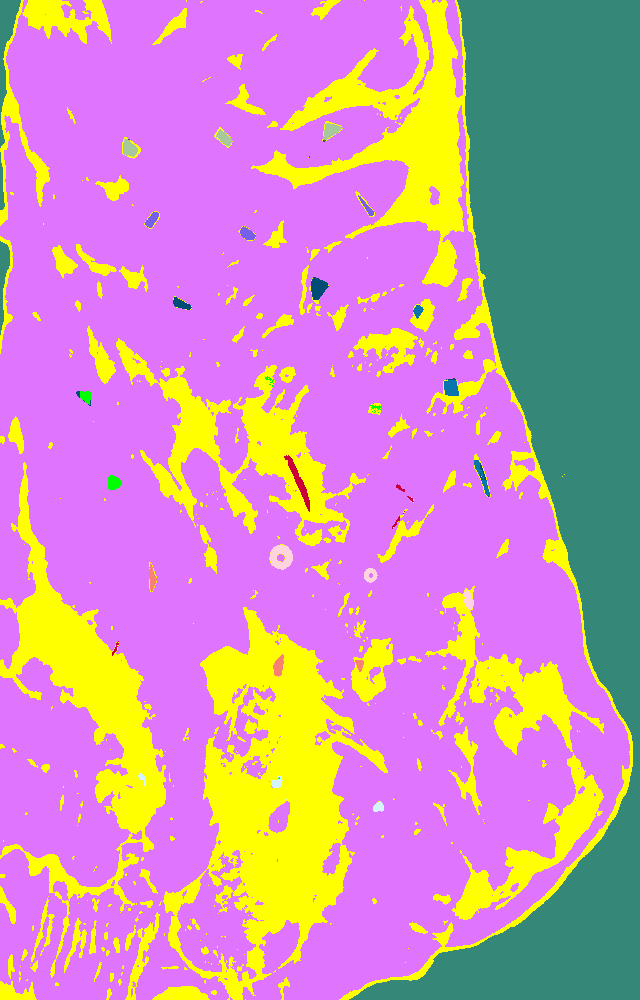} &
 \includegraphics[width=1in]{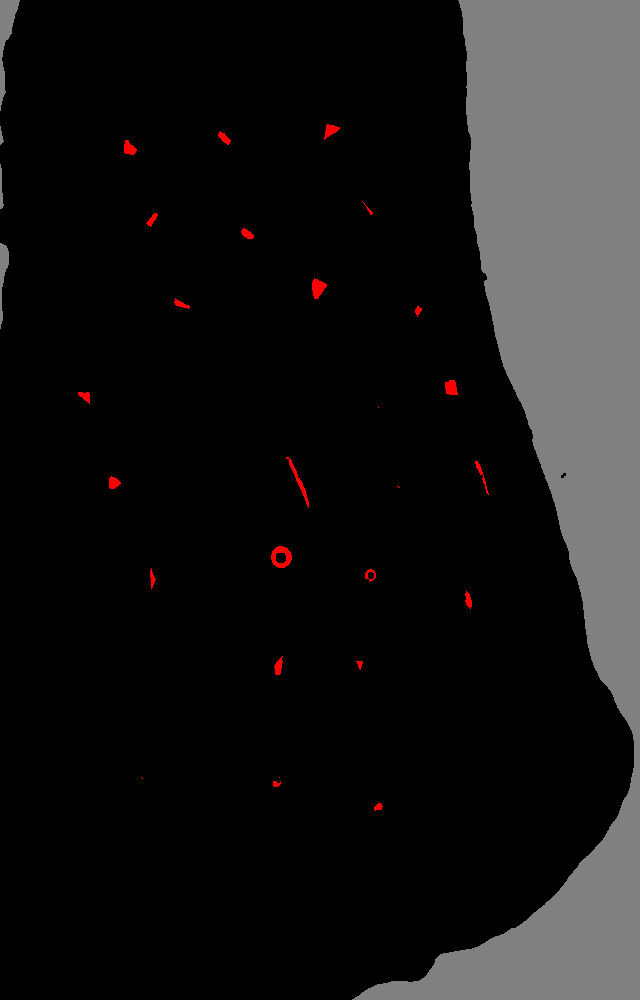} &
  \includegraphics[width=1in]{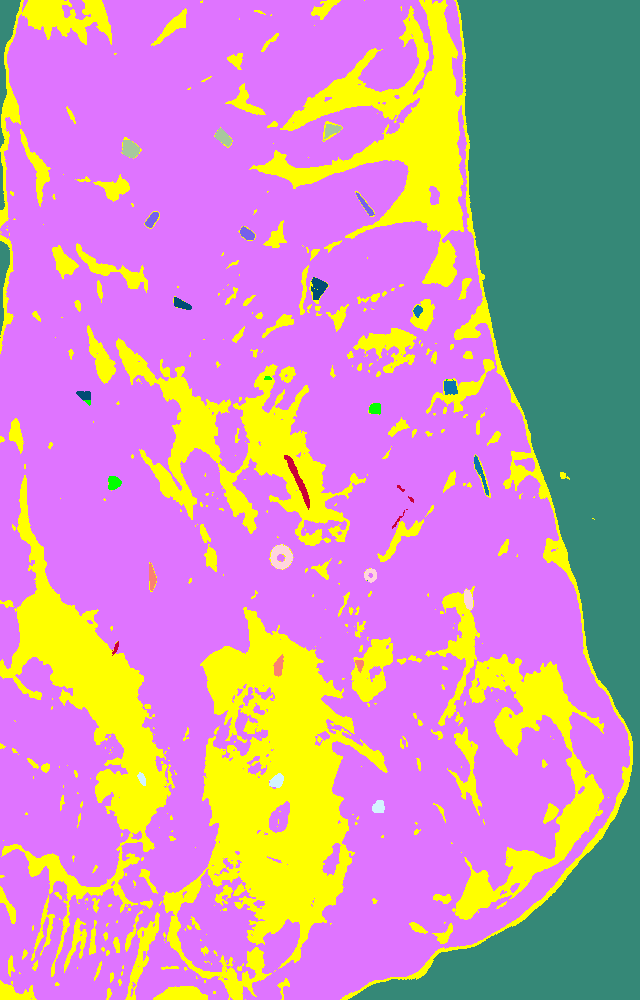} &
 \includegraphics[width=1in]{src_Specim-FX17e-076900055547_12_pred_conveyor_pork_dilation.png}
 \\
 \includegraphics[width=1in]{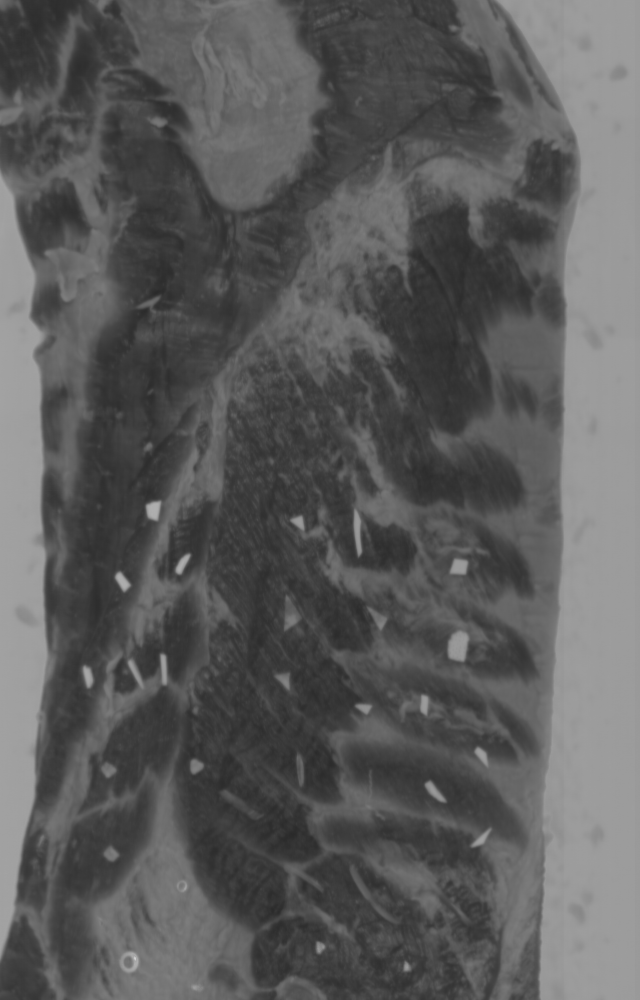} &
 \includegraphics[width=1in]{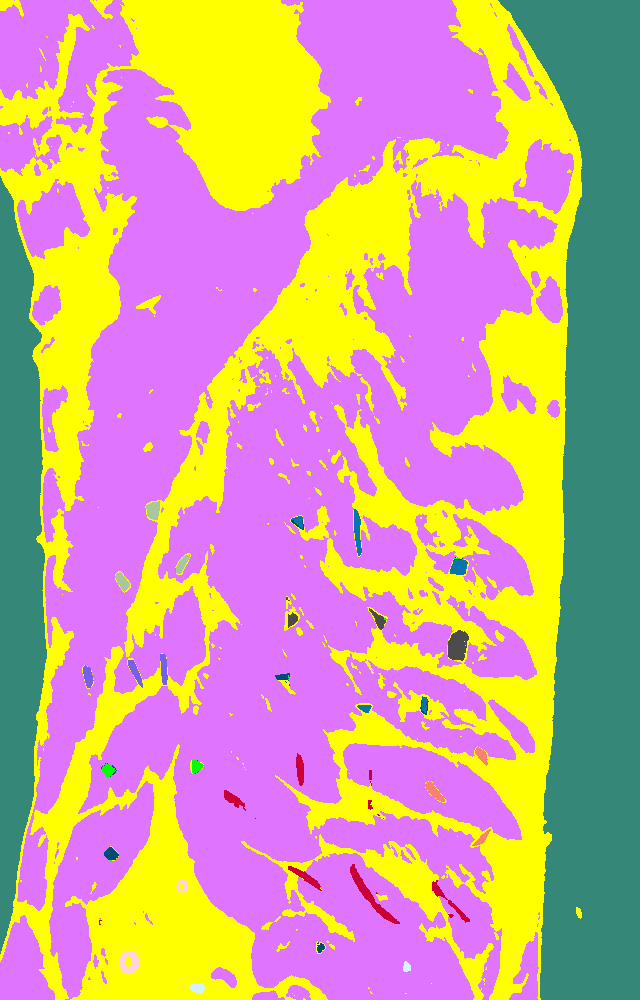} &
 \includegraphics[width=1in]{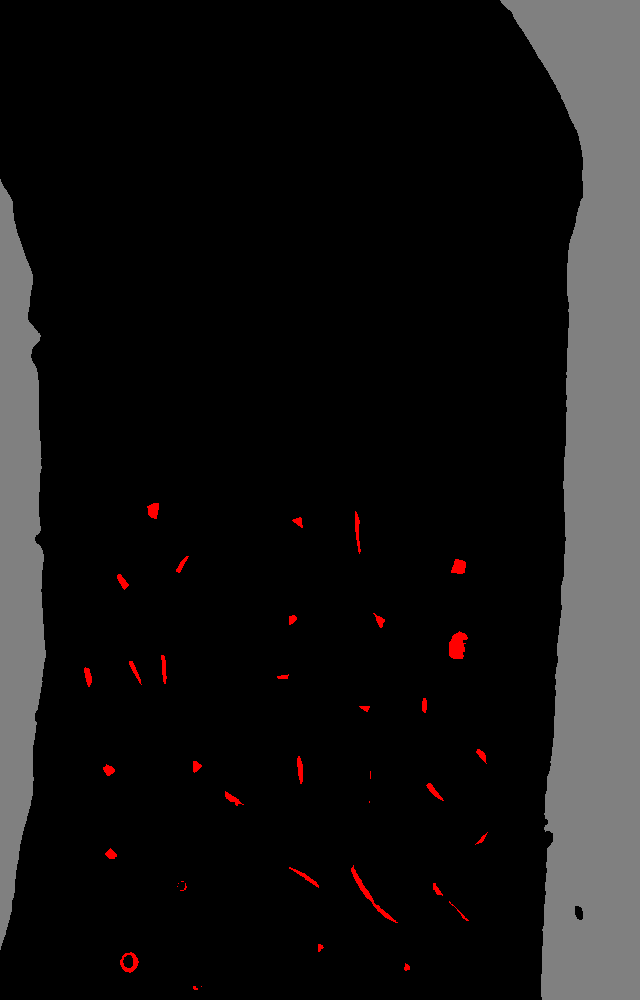} &
   \includegraphics[width=1in]{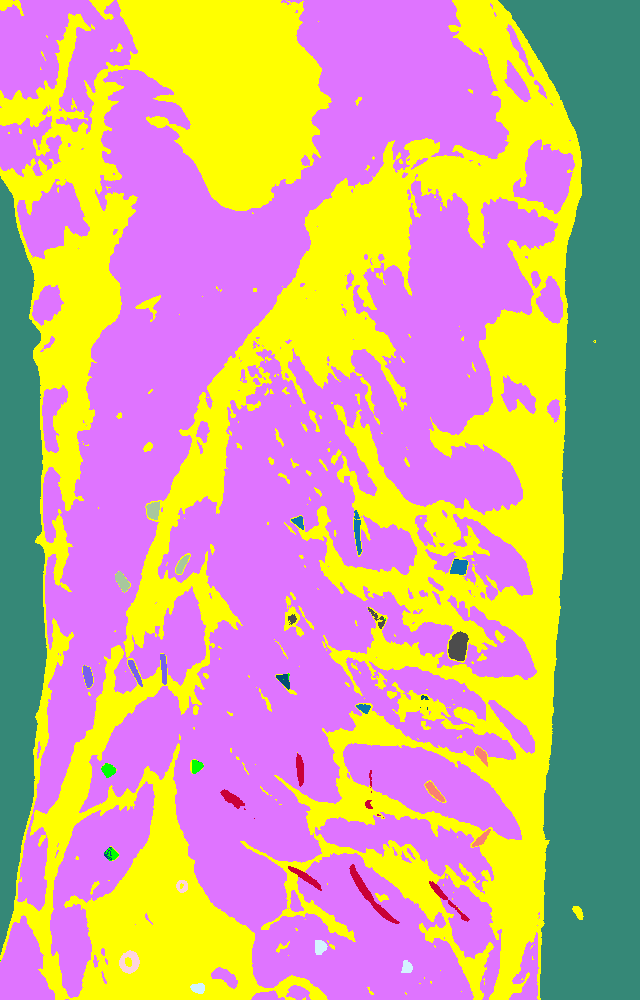} &
 \includegraphics[width=1in]{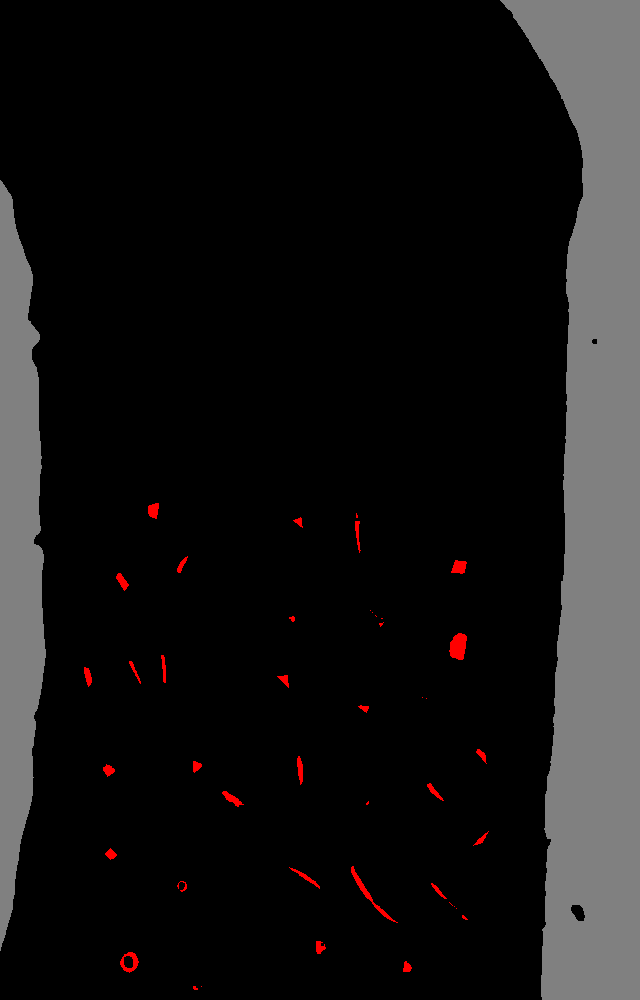}
 \\
  \textbf{(a)} &
 \textbf{(b)} &
 \textbf{(c)} &
 \textbf{(d)} &
 \textbf{(e)} 
 \\
\end{tabular}
\caption{Prediction examples on meat. (a) Spectral wavelength \SI{1169.5}{\nano\meter}; (b) Prediction of the normalizing model; (c) Prediction of the normalizing model followed by Post-Processing and Contaminant Erosion; (d) Prediction of the model without normalizing; (e) Prediction of the model without normalizing followed by Post-Processing and Contaminant Erosion.}
\label{fig:normalizing_model_results_meat}
\end{figure}

\figref{fig:normalizing_model_results_fat} and \figref{fig:normalizing_model_results_meat} introduce several prediction examples from the test set with contaminants, showing a successful true positive prediction for both fat and meat. In both figures, the prediction images in (b) and (d) have the same material colors as for the annotation in \figref{fig:annotations}, while the prediction masks in (c) and (e) use a simplified color notation: red for samples of contaminant, black for pork and gray for conveyor belt. 

\begin{figure}
\centering
\begin{tabular}{cc}
 \includegraphics[width=1in]{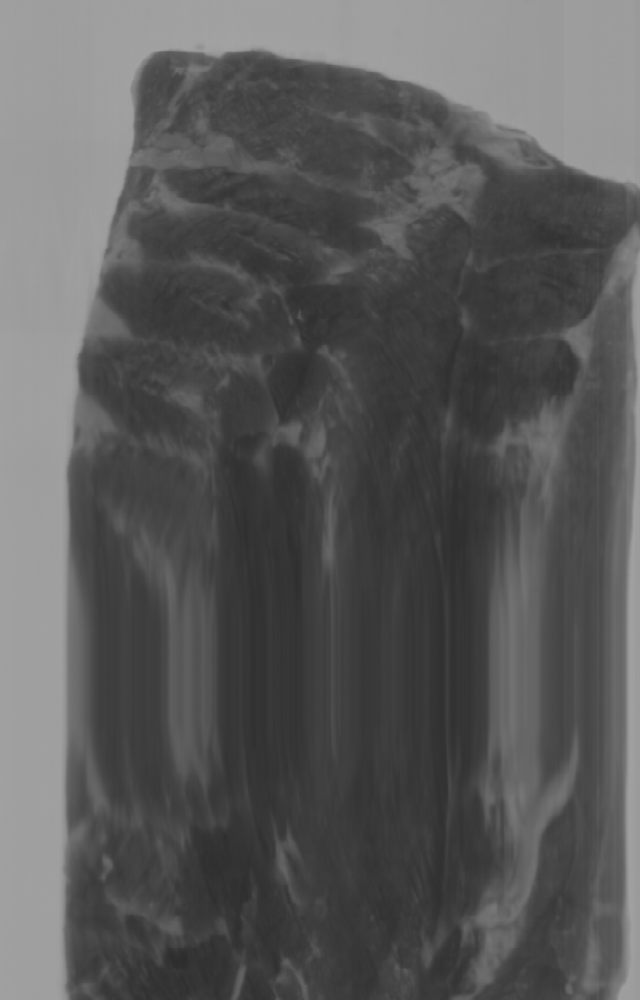} &
 \includegraphics[width=1in]{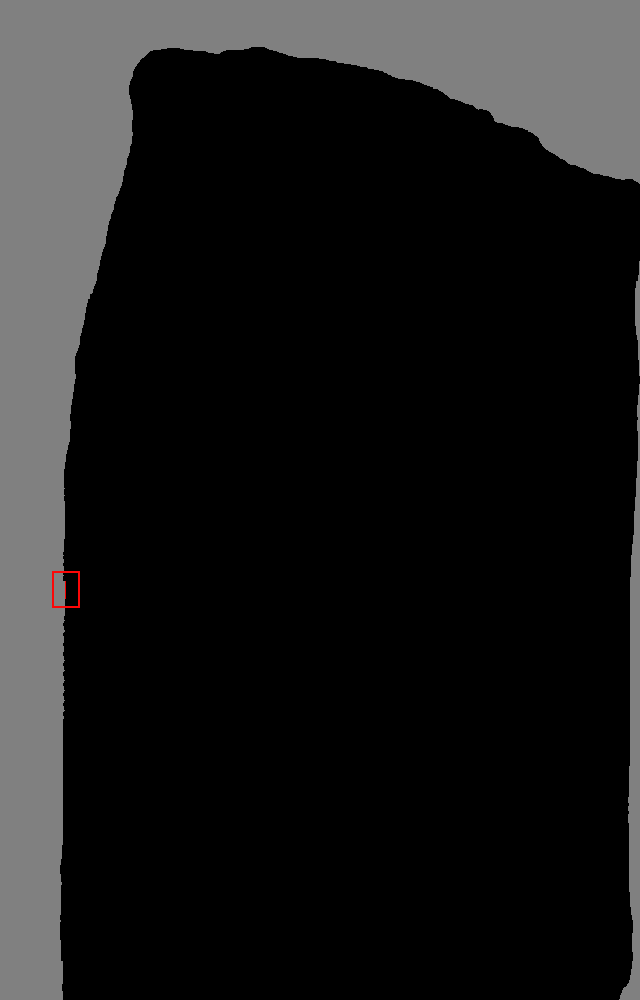} \\ 
\end{tabular}
\caption{Image with false positives marked in red.}
\label{fig:image_fp_temperature}
\end{figure}

To further assess robustness, we tested the model with spectral normalization on an additional set of 208 hyperspectral images acquired under the same illumination and camera parameters but across a wide range of sensor temperatures (\SI{55}[10-]{\degreeCelsius}). This dataset presented typical variations in temperature encountered on an industrial production day. The model produced false positives in only one image, which included a 7-pixel blob labeled as Nitrile and an 11-pixel blob labeled as White Conveyor Belt when the sensor temperature was \SI{48}{\degreeCelsius}. This image, shown in Figure \ref{fig:image_fp_temperature}, was malformed, explaining the misclassification. 

Of the ten investigated contaminants, the fine pieces of PEHD contaminant are the most challenging to segment. The plot examples in \figref{fig:plot_fp_samples} illustrate the most difficult scenario. During our numerous experiments testing different architectures, we found that all models pose difficulties when hyperspectral samples have similar curves to these examples. Finally, we evaluated the model on a challenging dataset of 95 hyperspectral images of pork belly on the conveyor belt containing blobs with spectral curves similar to the blue curves in \figref{fig:plot_fp_samples} around the inflection point at frequency \SI{1222}{\nano\meter}. The model with spectral normalization, post-processing, and erosion steps produced no false positives on this dataset, further demonstrating its robustness in separating challenging contaminants like PEHD from other materials.
 

\section{Conclusion and future work}
\label{sec:conclusion}
In this study, we implemented a segmentation-based approach for detecting foreign objects on pork belly meat using hyperspectral imaging and a lightweight Vision Transformer. Unlike traditional patch-based methods that classify only the center pixel of overlapping patches \textendash resulting in significant computational overhead \textendash we classified all pixels within a patch while leveraging the spatial context of neighboring pixels. This approach not only reduced training time but also addressed challenges associated with spectral similarity between contaminants and pork belly, improving efficiency and accuracy.

To enhance hyperspectral image quality, we also introduced two preprocessing steps. First, flat field correction was applied to remove fixed pattern noise using dark frame and flat field images. However, we observed that the temperature of the camera fluctuated significantly in real industrial scenarios, affecting these corrections and therefore the quality of predictions. The dark frame and flat field should therefore be computed for each sensor temperature value or range. Temperature fluctuations also influenced the average gray-level intensity and noise level. Consequently, normalization was employed as the second preprocessing step to mitigate variations in average gray-level intensities. Additionally, a bright material like Teflon was chosen as the homogeneous surface for flat field correction. Future work could incorporate materials with different gray-level intensities as uniform surfaces, and construct one model that fuzzes information from all of them into the flat field correction process.

After being tested on 55 images containing various samples of small foreign bodies, the 13-class model identified contaminants with a high degree of accuracy, demonstrating its effectiveness. However, it should be noted that when dealing with very small samples, the spectral signature on the boundaries of these foreign bodies merges with that of the surrounding material. This is a common diffraction effect in all image formation processes that remains a challenge. Post-processing steps were proposed to address this, effectively reducing false positives with minimal impact on true positive detection. Overall, the proposed approach offers a scalable solution for real-time foreign object detection.




\printcredits

\section*{Declaration of competing interest}
The authors declare that they have no known competing financial interests or personal relationships that could have appeared to influence the work reported in this paper.

\section*{Acknowledgements}
This study was supported by ACCIO, Cupó Indústria 4.0 with number ACE91/23/000435.

\bibliographystyle{cas-model2-names}

\bibliography{bibliography}

\begin{thebibliography}{30}
\expandafter\ifx\csname natexlab\endcsname\relax\def\natexlab#1{#1}\fi
\providecommand{\url}[1]{\texttt{#1}}
\providecommand{\href}[2]{#2}
\providecommand{\path}[1]{#1}
\providecommand{\DOIprefix}{doi:}
\providecommand{\ArXivprefix}{arXiv:}
\providecommand{\URLprefix}{URL: }
\providecommand{\Pubmedprefix}{pmid:}
\providecommand{\doi}[1]{\href{http://dx.doi.org/#1}{\path{#1}}}
\providecommand{\Pubmed}[1]{\href{pmid:#1}{\path{#1}}}
\providecommand{\bibinfo}[2]{#2}
\ifx\xfnm\relax \def\xfnm[#1]{\unskip,\space#1}\fi
\bibitem[{Buyukcan and Kavdir(2017)}]{Buyukcan2017}
\bibinfo{author}{Buyukcan, M.B.}, \bibinfo{author}{Kavdir, I.}, \bibinfo{year}{2017}.
\newblock \bibinfo{title}{Prediction of some internal quality parameters of apricot using ft-nir spectroscopy}.
\newblock \bibinfo{journal}{Journal of Food Measurement and Characterization} \bibinfo{volume}{11}, \bibinfo{pages}{651–659}.
\newblock \DOIprefix\doi{https://doi.org/10.1007/s11694-016-9434-9}.
\bibitem[{Chakraborty and Trehan(2021)}]{Chakraborty2021}
\bibinfo{author}{Chakraborty, T.}, \bibinfo{author}{Trehan, U.}, \bibinfo{year}{2021}.
\newblock \bibinfo{title}{Spectralnet: Exploring spatial-spectral waveletcnn for hyperspectral image classification} \href{http://arxiv.org/abs/2104.00341}{\tt arXiv:2104.00341}.
\bibitem[{Cheng et~al.(2017)Cheng, Nicolai and Sun}]{Cheng2017}
\bibinfo{author}{Cheng, J.H.}, \bibinfo{author}{Nicolai, B.}, \bibinfo{author}{Sun, D.W.}, \bibinfo{year}{2017}.
\newblock \bibinfo{title}{Hyperspectral imaging with multivariate analysis for technological parameters prediction and classification of muscle foods: A review}.
\newblock \bibinfo{journal}{Meat Science} \bibinfo{volume}{123}, \bibinfo{pages}{182--191}.
\newblock \DOIprefix\doi{https://doi.org/10.1016/j.meatsci.2016.09.017}.
\bibitem[{Dang et~al.(2023)Dang, Weng, Hou, Zuo and Liu}]{Dang2023}
\bibinfo{author}{Dang, L.}, \bibinfo{author}{Weng, L.}, \bibinfo{author}{Hou, Y.}, \bibinfo{author}{Zuo, X.}, \bibinfo{author}{Liu, Y.}, \bibinfo{year}{2023}.
\newblock \bibinfo{title}{Double-branch feature fusion transformer for hyperspectral image classification}.
\newblock \bibinfo{journal}{Scientific Reports} \bibinfo{volume}{13}.
\newblock \DOIprefix\doi{10.1038/s41598-023-27472-z}.
\bibitem[{Dosovitskiy et~al.(2020)Dosovitskiy, Beyer, Kolesnikov, Weissenborn, Zhai, Unterthiner, Dehghani, Minderer, Heigold, Gelly, Uszkoreit and Houlsby}]{Dosovitskiy2020}
\bibinfo{author}{Dosovitskiy, A.}, \bibinfo{author}{Beyer, L.}, \bibinfo{author}{Kolesnikov, A.}, \bibinfo{author}{Weissenborn, D.}, \bibinfo{author}{Zhai, X.}, \bibinfo{author}{Unterthiner, T.}, \bibinfo{author}{Dehghani, M.}, \bibinfo{author}{Minderer, M.}, \bibinfo{author}{Heigold, G.}, \bibinfo{author}{Gelly, S.}, \bibinfo{author}{Uszkoreit, J.}, \bibinfo{author}{Houlsby, N.}, \bibinfo{year}{2020}.
\newblock \bibinfo{title}{An image is worth 16x16 words: Transformers for image recognition at scale}.
\newblock \bibinfo{journal}{ArXiv} \bibinfo{volume}{abs/2010.11929}.
\bibitem[{Gualtieri and Cromp(1999)}]{Gualtieri1999}
\bibinfo{author}{Gualtieri, J.A.}, \bibinfo{author}{Cromp, R.F.}, \bibinfo{year}{1999}.
\newblock \bibinfo{title}{Support vector machines for hyperspectral remote sensing classification}.
\newblock \bibinfo{journal}{27th AIPR workshop: Advances in computer-assisted recognition} \bibinfo{volume}{3584}, \bibinfo{pages}{221--232}.
\bibitem[{Hang et~al.(2017)Hang, Liu, Hong and Ghamisi}]{Hang2017}
\bibinfo{author}{Hang, R.}, \bibinfo{author}{Liu, Q.}, \bibinfo{author}{Hong, D.}, \bibinfo{author}{Ghamisi, P.}, \bibinfo{year}{2017}.
\newblock \bibinfo{title}{Cascaded recurrent neural networks for hyperspectral image classification}.
\newblock \bibinfo{journal}{IEEE Transactions on Geoscience and Remote Sensing} \bibinfo{volume}{57}, \bibinfo{pages}{5384--5394}.
\bibitem[{Hu et~al.(2017)Hu, Pan, Sun, Tu, Sun, Wei and Tu}]{Hu2017}
\bibinfo{author}{Hu, H.}, \bibinfo{author}{Pan, L.}, \bibinfo{author}{Sun, K.}, \bibinfo{author}{Tu, S.}, \bibinfo{author}{Sun, Y.}, \bibinfo{author}{Wei, Y.}, \bibinfo{author}{Tu, K.}, \bibinfo{year}{2017}.
\newblock \bibinfo{title}{Differentiation of deciduous-calyx and persistent-calyx pears using hyperspectral reflectance imaging and multivariate analysis}.
\newblock \bibinfo{journal}{Computers and Electronics in Agriculture} \bibinfo{volume}{137}, \bibinfo{pages}{150--156}.
\newblock \DOIprefix\doi{https://doi.org/10.1016/j.compag.2017.04.002}.
\bibitem[{Hu et~al.(2016)Hu, Dong and Liu}]{Hu2016}
\bibinfo{author}{Hu, M.H.}, \bibinfo{author}{Dong, Q.L.}, \bibinfo{author}{Liu, B.L.}, \bibinfo{year}{2016}.
\newblock \bibinfo{title}{Classification and characterization of blueberry mechanical damage with time evolution using reflectance, transmittance and interactance imaging spectroscopy}.
\newblock \bibinfo{journal}{Computers and Electronics in Agriculture} \bibinfo{volume}{122}, \bibinfo{pages}{19--28}.
\newblock \DOIprefix\doi{https://doi.org/10.1016/j.compag.2016.01.015}.
\bibitem[{Hu et~al.(2015)Hu, Huang, Li, Zhang and Li}]{Hu2015}
\bibinfo{author}{Hu, W.}, \bibinfo{author}{Huang, Y.}, \bibinfo{author}{Li, W.}, \bibinfo{author}{Zhang, F.}, \bibinfo{author}{Li, H.}, \bibinfo{year}{2015}.
\newblock \bibinfo{title}{Deep convolutional neural networks for hyperspectral image classification}.
\newblock \bibinfo{journal}{J. Sensors} \bibinfo{volume}{2015}, \bibinfo{pages}{258619:1--258619:12}.
\bibitem[{Jelali and Papadopoulos(2024)}]{Jelali2024}
\bibinfo{author}{Jelali, M.}, \bibinfo{author}{Papadopoulos, K.}, \bibinfo{year}{2024}.
\newblock \bibinfo{title}{Inline inspection of packaged food using microwave/terahertz sensing—an overview with focus on confectionery products}.
\newblock \bibinfo{journal}{Processes} .
\bibitem[{Kalaiarasi(2012)}]{Kalaiarasi2012}
\bibinfo{author}{Kalaiarasi, G.;~Maheswari, S.}, \bibinfo{year}{2012}.
\newblock \bibinfo{title}{{Frost filtered scale-invariant feature extraction and multilayer perceptron for hyperspectral image classification}}.
\newblock \bibinfo{journal}{arXiv 2020} \URLprefix \url{https://arxiv.org/ftp/arxiv/papers/2006/2006.12556.pdf}.
\bibitem[{Khodabakhshian et~al.(2017)Khodabakhshian, Emadi, Khojastehpour, Golzarian and Sazgarnia}]{Khodabakhshian2017}
\bibinfo{author}{Khodabakhshian, R.}, \bibinfo{author}{Emadi, B.}, \bibinfo{author}{Khojastehpour, M.}, \bibinfo{author}{Golzarian, M.R.}, \bibinfo{author}{Sazgarnia, A.}, \bibinfo{year}{2017}.
\newblock \bibinfo{title}{Development of a multispectral imaging system for online quality assessment of pomegranate fruit}.
\newblock \bibinfo{journal}{International Journal of Food Properties} \bibinfo{volume}{20}, \bibinfo{pages}{107--118}.
\newblock \DOIprefix\doi{10.1080/10942912.2016.1144200}.
\bibitem[{Liu et~al.(2014)Liu, Liu, Lu, Ma, Chen, Yang and Zheng}]{Liu2014}
\bibinfo{author}{Liu, C.}, \bibinfo{author}{Liu, W.}, \bibinfo{author}{Lu, X.}, \bibinfo{author}{Ma, F.}, \bibinfo{author}{Chen, W.}, \bibinfo{author}{Yang, J.}, \bibinfo{author}{Zheng, L.}, \bibinfo{year}{2014}.
\newblock \bibinfo{title}{Application of multispectral imaging to determine quality attributes and ripeness stage in strawberry fruit}.
\newblock \bibinfo{journal}{PLoS ONE} \bibinfo{volume}{9}.
\bibitem[{Liu et~al.(2023)Liu, Wang, Zhang, Feng, Liu and Zhu}]{Liu2023}
\bibinfo{author}{Liu, Z.}, \bibinfo{author}{Wang, S.}, \bibinfo{author}{Zhang, Y.}, \bibinfo{author}{Feng, Y.}, \bibinfo{author}{Liu, J.}, \bibinfo{author}{Zhu, H.}, \bibinfo{year}{2023}.
\newblock \bibinfo{title}{Artificial intelligence in food safety: A decade review and bibliometric analysis}.
\newblock \bibinfo{journal}{Foods} \bibinfo{volume}{12}.
\newblock \DOIprefix\doi{10.3390/foods12061242}.
\bibitem[{Ma et~al.(2010)Ma, Crawford and Tian}]{Ma2010}
\bibinfo{author}{Ma, L.}, \bibinfo{author}{Crawford, M.}, \bibinfo{author}{Tian, J.}, \bibinfo{year}{2010}.
\newblock \bibinfo{title}{Local manifold learning-based -nearest-neighbor for hyperspectral image classification}.
\newblock \bibinfo{journal}{Geoscience and Remote Sensing, IEEE Transactions on} \bibinfo{volume}{48}, \bibinfo{pages}{4099 -- 4109}.
\newblock \DOIprefix\doi{10.1109/TGRS.2010.2055876}.
\bibitem[{Paoletti et~al.(2019a)Paoletti, Haut, Plaza and Plaza}]{Paoletti2019}
\bibinfo{author}{Paoletti, M.}, \bibinfo{author}{Haut, J.}, \bibinfo{author}{Plaza, J.}, \bibinfo{author}{Plaza, A.}, \bibinfo{year}{2019}a.
\newblock \bibinfo{title}{Deep learning classifiers for hyperspectral imaging: A review}.
\newblock \bibinfo{journal}{ISPRS Journal of Photogrammetry and Remote Sensing} \bibinfo{volume}{158}, \bibinfo{pages}{279--317}.
\newblock \DOIprefix\doi{https://doi.org/10.1016/j.isprsjprs.2019.09.006}.
\bibitem[{Paoletti et~al.(2019b)Paoletti, Haut, Fernandez-Beltran, Plaza, Plaza and Pla}]{Paoletti}
\bibinfo{author}{Paoletti, M.E.}, \bibinfo{author}{Haut, J.M.}, \bibinfo{author}{Fernandez-Beltran, R.}, \bibinfo{author}{Plaza, J.}, \bibinfo{author}{Plaza, A.J.}, \bibinfo{author}{Pla, F.}, \bibinfo{year}{2019}b.
\newblock \bibinfo{title}{Deep pyramidal residual networks for spectral–spatial hyperspectral image classification}.
\newblock \bibinfo{journal}{IEEE Transactions on Geoscience and Remote Sensing} \bibinfo{volume}{57}, \bibinfo{pages}{740--754}.
\newblock \DOIprefix\doi{10.1109/TGRS.2018.2860125}.
\bibitem[{Ropodi et~al.(2017)Ropodi, Panagou and Nychas}]{Ropodi2017}
\bibinfo{author}{Ropodi, A.I.}, \bibinfo{author}{Panagou, E.Z.}, \bibinfo{author}{Nychas, G.J.E.}, \bibinfo{year}{2017}.
\newblock \bibinfo{title}{Multispectral imaging (msi): A promising method for the detection of minced beef adulteration with horsemeat}.
\newblock \bibinfo{journal}{Food Control} \bibinfo{volume}{73}, \bibinfo{pages}{57--63}.
\newblock \DOIprefix\doi{https://doi.org/10.1016/j.foodcont.2016.05.048}. \bibinfo{note}{special issues from the ‘29th EFFoST International Conference’}.
\bibitem[{Roy et~al.(2019)Roy, Krishna, Dubey and Chaudhuri}]{Roy2019}
\bibinfo{author}{Roy, S.K.}, \bibinfo{author}{Krishna, G.}, \bibinfo{author}{Dubey, S.R.}, \bibinfo{author}{Chaudhuri, B.B.}, \bibinfo{year}{2019}.
\newblock \bibinfo{title}{Hybridsn: Exploring 3-d–2-d cnn feature hierarchy for hyperspectral image classification}.
\newblock \bibinfo{journal}{IEEE Geoscience and Remote Sensing Letters} \bibinfo{volume}{17}, \bibinfo{pages}{277--281}.
\bibitem[{Roy et~al.(2021)Roy, Manna, Song and Bruzzone}]{Roy2021}
\bibinfo{author}{Roy, S.K.}, \bibinfo{author}{Manna, S.}, \bibinfo{author}{Song, T.}, \bibinfo{author}{Bruzzone, L.}, \bibinfo{year}{2021}.
\newblock \bibinfo{title}{Attention-based adaptive spectral–spatial kernel resnet for hyperspectral image classification}.
\newblock \bibinfo{journal}{IEEE Transactions on Geoscience and Remote Sensing} \bibinfo{volume}{59}, \bibinfo{pages}{7831--7843}.
\newblock \DOIprefix\doi{10.1109/TGRS.2020.3043267}.
\bibitem[{Seibert et~al.(1998)Seibert, Boone and Lindfors}]{Seibert1998}
\bibinfo{author}{Seibert, J.A.}, \bibinfo{author}{Boone, J.M.}, \bibinfo{author}{Lindfors, K.K.}, \bibinfo{year}{1998}.
\newblock \bibinfo{title}{Flat-field correction technique for digital detectors}, in: \bibinfo{booktitle}{Medical Imaging}.
\bibitem[{Shi et~al.(2021)Shi, Wang, Mohammad, Young, Newman, Berg and Sun}]{Shi2021}
\bibinfo{author}{Shi, Y.}, \bibinfo{author}{Wang, X.}, \bibinfo{author}{Mohammad, B.}, \bibinfo{author}{Young, J.}, \bibinfo{author}{Newman, D.}, \bibinfo{author}{Berg, E.}, \bibinfo{author}{Sun, X.}, \bibinfo{year}{2021}.
\newblock \bibinfo{title}{A review on meat quality evaluation methods based on non-destructive computer vision and artificial intelligence technologies}.
\newblock \bibinfo{journal}{Food Science of Animal Resources} \bibinfo{volume}{41}.
\newblock \DOIprefix\doi{10.5851/kosfa.2021.e25}.
\bibitem[{Sun et~al.(2018)Sun, Xiao, Tu, Sun, Pan and Tu}]{Sun2018}
\bibinfo{author}{Sun, Y.}, \bibinfo{author}{Xiao, H.}, \bibinfo{author}{Tu, S.}, \bibinfo{author}{Sun, K.}, \bibinfo{author}{Pan, L.}, \bibinfo{author}{Tu, K.}, \bibinfo{year}{2018}.
\newblock \bibinfo{title}{Detecting decayed peach using a rotating hyperspectral imaging testbed}.
\newblock \bibinfo{journal}{LWT} \bibinfo{volume}{87}, \bibinfo{pages}{326--332}.
\newblock \DOIprefix\doi{https://doi.org/10.1016/j.lwt.2017.08.086}.
\bibitem[{Tian et~al.(2018)Tian, Li, Wang, Fan and Huang}]{Tian2018}
\bibinfo{author}{Tian, X.}, \bibinfo{author}{Li, J.}, \bibinfo{author}{Wang, Q.}, \bibinfo{author}{Fan, S.}, \bibinfo{author}{Huang, W.}, \bibinfo{year}{2018}.
\newblock \bibinfo{title}{A bi-layer model for nondestructive prediction of soluble solids content in apple based on reflectance spectra and peel pigments}.
\newblock \bibinfo{journal}{Food Chemistry} \bibinfo{volume}{239}, \bibinfo{pages}{1055--1063}.
\newblock \DOIprefix\doi{https://doi.org/10.1016/j.foodchem.2017.07.045}.
\bibitem[{Zappia et~al.(2021)Zappia, Crocco and Catapano}]{Zappia2021}
\bibinfo{author}{Zappia, S.}, \bibinfo{author}{Crocco, L.}, \bibinfo{author}{Catapano, I.}, \bibinfo{year}{2021}.
\newblock \bibinfo{title}{THz Imaging for Food Inspections: A Technology Review and Future Trends}.
\newblock \DOIprefix\doi{10.5772/intechopen.97615}.
\bibitem[{Zhang et~al.(2015)Zhang, Pan, Tu, Zhan and Tu}]{Zhang2015}
\bibinfo{author}{Zhang, W.}, \bibinfo{author}{Pan, L.}, \bibinfo{author}{Tu, S.}, \bibinfo{author}{Zhan, G.}, \bibinfo{author}{Tu, K.}, \bibinfo{year}{2015}.
\newblock \bibinfo{title}{Non-destructive internal quality assessment of eggs using a synthesis of hyperspectral imaging and multivariate analysis}.
\newblock \bibinfo{journal}{Journal of Food Engineering} \bibinfo{volume}{157}, \bibinfo{pages}{41--48}.
\newblock \DOIprefix\doi{https://doi.org/10.1016/j.jfoodeng.2015.02.013}.
\bibitem[{Zhong et~al.(2018)Zhong, Li, Luo and Chapman}]{Zhong2018}
\bibinfo{author}{Zhong, Z.}, \bibinfo{author}{Li, J.}, \bibinfo{author}{Luo, Z.}, \bibinfo{author}{Chapman, M.}, \bibinfo{year}{2018}.
\newblock \bibinfo{title}{Spectral–spatial residual network for hyperspectral image classification: A 3-d deep learning framework}.
\newblock \bibinfo{journal}{IEEE Transactions on Geoscience and Remote Sensing} \bibinfo{volume}{56}, \bibinfo{pages}{847--858}.
\newblock \DOIprefix\doi{10.1109/TGRS.2017.2755542}.
\bibitem[{Zhu et~al.(2021a)Zhu, Spachos, Pensini and Plataniotis}]{Zhu2021}
\bibinfo{author}{Zhu, L.}, \bibinfo{author}{Spachos, P.}, \bibinfo{author}{Pensini, E.}, \bibinfo{author}{Plataniotis, K.N.}, \bibinfo{year}{2021}a.
\newblock \bibinfo{title}{Deep learning and machine vision for food processing: A survey}.
\newblock \bibinfo{journal}{Current Research in Food Science} \bibinfo{volume}{4}, \bibinfo{pages}{233--249}.
\newblock \DOIprefix\doi{https://doi.org/10.1016/j.crfs.2021.03.009}.
\bibitem[{Zhu et~al.(2021b)Zhu, Deng, Zheng, Zhong, Guan, Lin, Zhang and Li}]{ZhuQ2021}
\bibinfo{author}{Zhu, Q.}, \bibinfo{author}{Deng, W.}, \bibinfo{author}{Zheng, Z.}, \bibinfo{author}{Zhong, Y.}, \bibinfo{author}{Guan, Q.}, \bibinfo{author}{Lin, W.}, \bibinfo{author}{Zhang, L.}, \bibinfo{author}{Li, D.}, \bibinfo{year}{2021}b.
\newblock \bibinfo{title}{A spectral-spatial-dependent global learning framework for insufficient and imbalanced hyperspectral image classification}.
\newblock \bibinfo{journal}{IEEE Transactions on Cybernetics} \bibinfo{volume}{52}, \bibinfo{pages}{11709--11723}.

\end{thebibliography}

\end{document}